%% file: neurips_2025.tex
\documentclass{article}

\usepackage[dblblindworkshop, final]{neurips_2025}

\usepackage[utf8]{inputenc} 
\usepackage[T1]{fontenc}    
\usepackage{hyperref}       
\usepackage{url}            
\usepackage{booktabs}       
\usepackage{amsfonts}       
\usepackage{nicefrac}       
\usepackage{microtype}      
\usepackage{xcolor}         

\usepackage{graphicx}
\usepackage{amsmath}
\usepackage{multirow}
\usepackage{graphicx}
\usepackage{algorithm}
\usepackage{algpseudocode}
\usepackage{multirow}

\usepackage{subcaption}  
\usepackage[font=small,labelfont=bf]{caption}  
\usepackage{array}
\usepackage{subcaption}
\usepackage{xfrac}
\usepackage{float}

\usepackage{amsmath}
\usepackage{amssymb}
\usepackage{mathtools}
\usepackage{amsthm}
\usepackage{subcaption}
\usepackage{amsmath, amsfonts, amssymb}
\usepackage{array}
\usepackage{subcaption}
\usepackage{xfrac}
\usepackage{float}

\usepackage{colortbl}

\title{Deep Modularity Networks with Diversity-Preserving Regularization}

\author{%
  Yasmin Salehi, Dennis Giannacopoulos\\
  Department of Electrical and Computering Engineering \\ 
  McGill University\\ 
  \texttt{yasmin.salehi@mail.mcgill.ca, dennis.giannacopoulos@mcgill.ca}\\ 
}

\begin{document}

\maketitle

\input{paper/abstract}

\input{paper/introduction}

\input{paper/related_work}

\input{paper/methodology}

\input{paper/experiments}

\input{paper/results}

\input{paper/conclusion}



\bibliographystyle{acl_natbib}
\bibliography{neurips_2025}


\newpage

\input{paper/checklist}

\newpage

\appendix

\input{paper/appendix}

\end{document}

%% file: paper/abstract.tex
\begin{abstract}
Graph clustering plays a crucial role in graph representation learning but often faces challenges in achieving feature-space diversity. While Deep Modularity Networks (DMoN) leverage modularity maximization and collapse regularization to ensure structural separation, they lack explicit mechanisms for feature-space separation, assignment dispersion, and assignment-confidence control. We address this limitation by proposing Deep Modularity Networks with Diversity-Preserving Regularization (DMoN–DPR), which introduces three novel regularization terms: distance-based for inter-cluster separation, variance-based for per-cluster assignment dispersion, and an assignment-entropy penalty with a small positive weight, encouraging more confident assignments gradually. Our method significantly enhances label-based clustering metrics on feature-rich benchmark datasets (paired two-tailed t-test, 
$p\leq0.05$), demonstrating the effectiveness of incorporating diversity-preserving regularizations in creating meaningful and interpretable clusters.
\end{abstract}

%% file: paper/introduction.tex
\section{Introduction}

\label{introduction}

Graph clustering is a crucial problem within the graph representation learning domain, and essential for various applications, including but not limited to community detection in social networks \citep{perozzi2014deepwalk, xiao2015detecting}, data exploration \citep{perozzi2018discovering}, and functional module identification in biological networks \citep{jin2021application}. Consequently, there has been a surge in methods aimed at enhancing graph clustering performance. Recent advancements have included techniques like Graph Neural Networks (GNNs) \citep{scarselli2008graph}, which leverage node features and graph structure for representation learning, often through unsupervised training \citep{tsitsulin2023graph}. In this context, graph pooling methods have also become prominent, as they provide a way to coarsen graphs by aggregating nodes into clusters \citep{cangea2018towards}. Yet, some of these methods such as DiffPool \citep{ying2018hierarchical} and MinCutPool \citep{bianchi2019mincut} were found to be computationally costly and/or too rigid, leading to poor convergence \citep{tsitsulin2023graph}.

To address these limitations, Deep Modularity Networks (DMoN) \citep{tsitsulin2023graph} were introduced, combining spectral modularity maximization \citep{newman2006modularity} with  collapse regularization \citep{tsitsulin2023graph} to avoid trivial clustering solutions. Although DMoN captures structural communities, two practical gaps remain. First, while the encoder may use node attributes, the objective itself is agnostic to feature‑space geometry: it contains no term that directly rewards inter‑cluster feature separation or specialization of cluster assignments across nodes. This can yield clusters that are structurally distinct yet not well separated in feature space. Second, the objective lacks an explicit control for assignment confidence (e.g., entropy/temperature), so soft assignments can harden early.

\paragraph{Why diversity matters.}
Accounting for both inter\mbox{-}cluster separation and intra\mbox{-}cluster variety prevents information collapse and yields richer representations. In practice, diversity improves downstream utility—for example, diversified recommendation in e\mbox{-}commerce optimizes cluster\mbox{-}level variety to boost engagement \citep{kim2025sequentially}, and drawing items from distinct clusters raises serendipity without sacrificing accuracy \citep{aytekin2014clustering}. Beyond recommendation, biological networks exhibit the same principle: modules that are cohesive inside yet distinct outside reveal functional relationships that density\mbox{-}only objectives can miss \citep{spirin2003protein}. Recent graph\mbox{-}pooling work therefore introduces explicit mechanisms to spread centroids and diversify within\mbox{-}cluster features \citep{liu2023exploring}. These considerations motivate our design.

\paragraph{Present Work}
Motivated by the above, we propose Deep Modularity Networks with Diversity\mbox{-}Preserving Regularization (DMoN\mbox{-}DPR), an extension of DMoN that augments its objective with three regularizers explicitly promoting diversity among clusters. These include a distance-based regularization term, which penalizes clusters with centroids too close in feature space to encourage distinct separation; a variance-based regularization term that increases the dispersion of assignment probabilities across nodes for each cluster, and an assignment‑entropy penalty added with a small positive weight, to avoid premature hard assignments and to let distance/variance terms drive diversity. By performing extensive evaluations, we demonstrate that our method improves clustering performance on the benchmark datasets, noticeably on Coauthor CS and Coauthor Physics datasets, which benefit from this enriched representation due to their rich feature spaces.

%% file: paper/related_work.tex
\section{Related Work}

Early graph clustering often decoupled features from structure. k-means on raw features \citep{lloyd1982least} ignores connectivity; pairing k-means with DeepWalk or DGI embeddings \citep{perozzi2014deepwalk, velivckovic2018deep, tsitsulin2023graph} injects structural signals but remains a two-stage pipeline, typically trailing end-to-end models that jointly learn representations and cluster assignments. Meanwhile, Chebyshev-based spectral convolutions \citep{defferrard2016convolutional} efficiently approximate graph filters, laying a scalable foundation for modern GNNs. 

On the other hand, pooling methods capture hierarchy. NOCD \citep{shchur2019overlapping} directly optimizes graph likelihood but can struggle with scale and feature use. DiffPool \citep{ying2018hierarchical} learns soft cluster assignments end-to-end yet incurs quadratic cost \citep{tsitsulin2023graph}. MinCut pooling \citep{bianchi2020spectral} adds normalized cut and orthogonality terms that may hinder convergence; the Ortho variant keeps only the latter and loses structural cues. SAGPool \citep{lee2019self} ranks node importance via self-attention, improving selectivity at added training cost on large graphs.

For unsupervised representation learning, DGI \citep{velickovic2019deep} maximizes mutual information between local and global summaries; InfoGraph extends this idea to graph-level representations \citep{sun2019infograph}.

More recently, DMoN \citep{tsitsulin2023graph} integrates modularity maximization with collapse regularization in an end-to-end framework, unifying community detection and neural feature learning and achieving strong NMI and modularity across benchmarks.

%% file: paper/methodology.tex
\section{Deep Modularity Networks with Diversity-Preserving Regularization}

DMoN tackles the issues of previous techniques by leveraging an optimization objective that combines insights from spectral modularity maximization \citep{newman2006modularity} with a unique regularization term called collapse regularization \citep{tsitsulin2023graph}. Specifically, DMoN encodes the cluster assignments, represented as a soft assignment matrix \( C \) by using a softmax function over the output of a GNN, allowing differentiation during optimization. For each node, a soft cluster assignment \( C \) is computed as follows:
\begin{equation}
C = \text{softmax}(\text{GCN}(\tilde{A}, X))
\end{equation}
where \( \text{GCN} \) is a multi-layer graph convolutional network, \( \tilde{A} \) is the normalized adjacency matrix \( \tilde{A} = D^{-\frac{1}{2}} A D^{-\frac{1}{2}} \), and \( X \) represents the node features. The objective function of DMoN, \( L_{\text{DMoN}} \), combines a modularity term with a collapse regularization term to optimize the clustering. The modularity term measures the quality of the cluster assignments by maximizing the density of intra-cluster edges relative to a null model, while the collapse regularization prevents trivial solutions where all nodes are assigned to the same cluster. The objective function is formulated as:

\begin{equation}
L_{\text{DMoN}}(C; A) = 
\underbrace{- \frac{1}{2m} \text{Tr}(C^\top B C)}_{\text{Modularity term}} 
+ 
\underbrace{{\frac{\sqrt k}{n}} \left\| \sum_{i} C_i^\top \right\|_F - 1}_{\text{Collapse Regularization term}}
\end{equation}

where \( (\frac{1}{2m}) \text{Tr}(C^\top B C) \) is the modularity term (measuring how well clusters are internally connected compared to random chance), \( B = A - \frac{d d^\top}{2m} \) is the modularity matrix, \( d \) is the degree vector, \( m \) is the total number of edges,  \( {\frac{\sqrt{k}}{n}} \left\| \sum_{i} C_i^\top \right\|_F - 1 \) represents the collapse regularization term, which discourages the formation of trivial clusters by penalizing the Frobenius norm of the assignment matrix \( C \), \( \|\cdot\|_F \) denotes the Frobenius norm, and \( k \) is the number of clusters. This regularization term encourages balanced cluster assignments, thereby improving the quality of clustering and avoiding degenerate solutions.

Building upon the DMoN framework, we augment the objective with three regularizers:
\begin{equation}
L_{\text{DMoN-DPR}}(C;A)
= L_{\text{DMoN}}
+ W_{\text{dist}}\,L_{\text{DPR}}^{\text{distance}}
+ W_{\text{var}}\,L_{\text{DPR}}^{\text{variance}}
+ W_{\text{entropy}}\,L_{\text{DPR}}^{\text{entropy}} .
\end{equation}
Here, $W_{\text{dist}}$, $W_{\text{var}}$, and $W_{\text{entropy}}$ control the influence of each term.
These additions promote \emph{inter-cluster separation} (distance) and \emph{per-cluster assignment dispersion} (variance). We use a \emph{small positive} entropy weight so per-node entropy decreases \emph{slowly}, preserving higher uncertainty early and thus \emph{indirectly} encouraging more balanced cluster usage during exploration; the primary mechanism for cluster-size balance remains DMoN’s collapse regularizer.

\paragraph{Distance-Based Regularization}

Inspired by SimCLR \citep{chen2020simple}, which showed that contrastive loss promotes well-separated clusters, the distance-based regularization term encourages distinct cluster centroids in feature space. It is defined as:
\begin{equation}
L_{\text{DPR}}^{\text{Distance}} = \frac{1}{k(k-1)} \sum_{i=1}^{k} \sum_{\substack{j=1 \\ j \neq i}}^{k} \text{ReLU}\left( \epsilon - \| \mu_i - \mu_j \|_2^2 \right)
\end{equation}  
where \( k \) is the number of clusters, \( \mu_i \) and \( \mu_j \) are the centroids of clusters \( i \) and \( j \), computed as \( \mu_i = \frac{\sum_{v=1}^{n} C_{vi} X_v}{\sum_{v=1}^{n} C_{vi}} \), where \( X_v \) is the feature vector of node \( v \),  \( \| \cdot \|_2^2 \) denotes the squared Euclidean norm, \( \epsilon \) is a predefined threshold that sets the minimum acceptable squared distance between cluster centroids, and \( \text{ReLU} \) is the Rectified Linear Unit function, ensuring that only distances below \( \epsilon \) contribute to the loss. By penalizing pairs of clusters whose centroids are closer than \( \epsilon \), clusters can be effectively pushed apart in the feature space. By encouraging greater separation between clusters, it enhances inter-cluster diversity and reduces overlap, leading to more distinguishable and meaningful clusters.

\paragraph{Variance-Based Regularization}

The variance-based regularization term encourages dispersion in the assignment matrix by maximizing, for each cluster, the variance of its assignment probabilities across nodes:
\begin{equation}
L_{\text{DPR}}^{\text{variance}} = -\frac{1}{k} \sum_{i=1}^{k} \text{Var}(C_{:i})
\end{equation}
where \( \text{Var}(C_{:i}) \) denotes the variance of the assignment probabilities of all nodes to cluster \( i \). Maximizing this variance prevents uniform cluster columns (e.g., every node assigned with equal probability), ensuring that clusters specialize over different subsets of nodes and thereby enhancing assignment diversity.

\paragraph{Entropy-Based Regularization}

We define:
\begin{equation}
L_{\text{DPR}}^{\text{entropy}} = -\frac{1}{n} \sum_{v=1}^{n} \sum_{i=1}^{k} C_{vi} \log (C_{vi} + \delta)
\end{equation} 
which is the average per‑node Shannon entropy of the soft assignments, where \( n \) is the number of nodes, \( C_{vi} \) is the soft assignment probability of node \( v \) to cluster \( i \), \( \delta \) is a small constant added to prevent logarithm of zero, ensuring numerical stability. We minimize $L_{\text{DPR}}$ to gently encourage more confident assignments. Because minimizing entropy can lead to premature hardening, we apply a small weight $W_{\text{entropy}}$ (typically 
0.001--0.1) so that entropy decreases \emph{slowly} during training.

%% file: paper/experiments.tex
\section{Experiments}

\paragraph{Evaluation Protocol} To evaluate the effectiveness of our proposed regularization objective, we conducted experiments on Cora, CiteSeer, and PubMed datasets \citep{yang2016revisiting}, as well as the Coauthor CS and Coauthor Physics datasets \citep{shchur2018pitfalls}. Information regarding these datasets is presented in Appendix \ref{sec:datasets}. Following the evaluation protocol outlined in \citep{tsitsulin2023graph}, we employed the following metrics to assess performance: graph conductance (C), modularity (Q), Normalized Mutual Information (NMI) with ground-truth labels, and the pairwise F1 measure. Additionally, we use the following quantitative measures to analyze feature‐space diversity: average inter‐centroid distance (mean pairwise Euclidean distance between cluster centroids), minimum inter‐centroid distance (the smallest pairwise distance among centroids), average intra‐cluster variance (average within‐cluster variance of node embeddings), and Silhouette score (a standard measure of per‐point cohesion vs. separation). The quantification of diversity results are presented in Appendix \ref{sec:quantification_of_diversity}.

\paragraph{Baseline \& Implementation Details} 

We select DiffPool \citep{ying2018hierarchical}, MinCut pooling \citep{bianchi2019mincut} and DMoN \citep{tsitsulin2023graph} as our baselines. Additionally, we focus on ablation studies and significance tests against vanilla DMoN to isolate the impact of the new diversity terms. Furthermore, similar to \citep{tsitsulin2023graph}, one layer of GCN \citep{kipf2016semi} with 512 neurons was used to create the graph embedding, followed by a pooling layer. Likewise, the models were trained for 1000 epochs using Adam optimizer with a learning rate of 0.001. The code--implemented by extending the DMoN implementation in PyTorch Geometric \citep{Fey/Lenssen/2019}-- is available at \url{www.github.com/YasminSalehi/DMoN-DPR}.

%% file: paper/results.tex
\section{Results}

\input{paper/gpu_results}

\subsection{Effectiveness of DMoN-DPR in Clustering}

To assess the effectiveness of DMoN-DPR, we compare the mean values of conductance ($C$), modularity ($Q$), NMI, and pairwise F1 measure across 10 randomly selected seeds achieved with DMoN-DPR (denoted as DPR in all the tables) to those obtained with our baselines. Tables \ref{tab:comparison_1} (Cora, CiteSeer, PubMed) and \ref{tab:comparison_2} (Coauthor CS, Coauthor Physics) summarize the results, where graph metrics (C, Q) do not rely on ground-truth labels, whereas ‘Labels’ metrics (NMI, F1) compare to known class labels.

For completeness, we defer supporting analyses to the appendix: (i) quantitative diagnostics of feature-space diversity (average/min inter-centroid distance, average intra-cluster variance, Silhouette score) (Appendix \ref{sec:quantification_of_diversity}), (ii) t\textendash SNE visualizations of the learned clusters versus ground-truth labels (Appendix \ref{sec:visualization}), (iii) 
ablation and hyperparameter-selection studies that isolate the effect of each DPR term (Appendix \ref{sec:ablation}), and (iv) per-seed scores and full tables with error bars (Appendix \ref{sec:full_results}).

\paragraph{Cora, CiteSeer, and PubMed.}
Overall, the results confirm that adding diversity-preserving regularizers to DMoN almost never harms the purely topological scores—conductance ($C$ ↓) and modularity ($Q$ ↑)—while consistently lifting label-aware metrics such as NMI and F1. On Cora, DPR(DV) achieves the best NMI (44.40\%) and only narrowly trails DPR(DE) on F1. On CiteSeer, although DiffPool attains the highest NMI (33.40\%) and F1 (47.83\%), DPR variants remain competitive without the instability that causes DiffPool to collapse on PubMed \footnote{With a single DiffPool layer and our fixed cluster budget ($C=3$),
the link-prediction loss on the 19 k-node PubMed graph stagnated and
the assignment matrix collapsed to one cluster. Increasing the budget restores numerical stability, but yields only marginal label quality and would also violate our fixed-$C$
protocol, so we omit DiffPool scores for PubMed.}. Among the DMoN-family methods, DPR(DV) is the strongest on CiteSeer. On PubMed, all DPR variants remain stable, in contrast to DiffPool, while MinCut yields the highest NMI (22.48\%) and DPR(DE) edges out others in F1. The fact that DPR(D) and DPR(DV) match the best conductance (8.60\%) confirms that the regularizers maintain cut quality.

\paragraph{Coauthor CS and Coauthor Physics.}
On the more feature-rich\footnote{We define feature-rich graphs as those that exhibit (i) high feature dimensionality and (ii) high average Shannon entropy per node. Please see Appendix \ref{sec:datasets} for more detail.} Coauthor graphs, diversity-preserving regularization brings substantial gains. On Coauthor CS, DPR(DVE) achieves a new best in F1 (62.67\%) while DPR(E) sets the highest NMI (71.58\%), both outperforming the already strong DMoN baseline without sacrificing conductance or modularity. On Coauthor Physics, DPR(DVE) obtains the lowest conductance (12.84\%) and, along with DPR(DV), surpasses 57\% F1—demonstrating that the full three-term objective generalizes well to larger, noisier graphs.

\paragraph{Key Insights.} Therefore, our results show that the effects of distance-based (D), variance-based (V), and entropy-based (E) regularizations depend on dataset feature richness. On feature-rich datasets like Coauthor CS and Physics, adding these terms—especially D—significantly boosts NMI and F1, with DMoN-DPR(DV) raising F1 by over 10 percentage points. The V term increases assignment dispersion across nodes and performs best when combined with D. The E term, used with a small positive weight, \emph{gradually} sharpens per-node assignments, which can \emph{indirectly} improve cluster balance during early exploration on feature-rich datasets, but offers little benefit on simpler datasets like PubMed; the collapse regularizer remains the primary mechanism for cluster-size balance. Overall, combining D and V often yields the highest performance, underscoring the need to tailor regularization to dataset characteristics for optimal clustering.

\paragraph{Statistical Significance.} 

To strengthen our claims, we conduct a paired t‐test on Conductance, Modularity, NMI, and F1, comparing DMoN vs. the best‐performing DMoN–DPR variants. We ensure both methods use exactly the same seeds in each dataset. The results are summarized in Table \ref{tab:sig_test_combined}. On the citation benchmarks we adopt the strongest variant for each graph (DV on Cora and CiteSeer, DVE on PubMed). Although the average NMI and F1 of DMoN–DPR surpass or match the vanilla DMoN baseline, the paired two-tailed $t$-tests in Table \ref{tab:sig_test_combined} show p-values above 0.10. This indicates that, given the modest sample of random seeds, the improvements are encouraging but not yet conclusive for these sparsely featured graphs. The picture is markedly different on the feature-rich coauthor networks.  The gains on NMI and F1 metrics are highly significant, indicated by $p \leq 0.05$, as seen in Table \ref{tab:sig_test_combined}. These results validate our hypothesis that encouraging latent cluster dispersion is most beneficial when node attributes are abundant and heterogeneous, and they confirm that DMoN–DPR maintains the stability advantage of the original objective while translating it into measurably better community recovery.

\paragraph{Trade-Offs Between Structural and Label-Based Metrics.} Our results reveal a clear trade-off between structural metrics (e.g., Modularity, Conductance) and label-based metrics (e.g., NMI, F1). Diversity-preserving regularization improves label alignment, especially on feature-rich datasets like Coauthor CS and Physics, by forming clusters that better match ground-truth labels—though sometimes at the cost of structural cohesion (e.g., lower modularity). Conversely, on datasets with less diverse or lower-dimensional features (e.g., PubMed), the structural metrics are less affected by the introduction of diversity-preserving regularization terms, and the benefits to label-based metrics are more modest. As a result, DMoN-DPR emerges as a more effective choice for supervised tasks prioritizing NMI and F1, whereas vanilla DMoN remains competitive for unsupervised scenarios where conductance and modularity are paramount. Tailoring the clustering strategy to the dataset and task requirements is therefore essential.

\subsection{Execution Times and Runtime Analysis}
\label{sec:runtime}

Table\ref{tab:runtime} reports the execution times (in seconds) of DMoN and DMoN--DPR on five commonly used graph datasets. All experiments were conducted on a CPU to ensure consistent runtime measurement without GPU scheduling effects.

\begin{table}[!htpb]
    \centering
    \small
    \caption{Execution times (in seconds) on CPU for DMoN and DMoN--DPR.}
    \label{tab:runtime}
    \begin{tabular}{lccccc}
        \hline
        \textbf{Method} & \textbf{Cora} & \textbf{CiteSeer} & \textbf{PubMed} & \textbf{Coauthor CS} & \textbf{Coauthor Physics} \\
        \hline
        DMoN         & 32   & 76   & 310  & 722  & 2107 \\
        DMoN--DPR    & 35   & 71   & 300  & 751  & 1946 \\
        \hline
    \end{tabular}
\end{table}

Although DMoN--DPR includes additional regularization terms for Distance-based, Entropy-based, and Variance-based constraints, its runtime is sometimes comparable to or even marginally lower than plain DMoN. The reason is that the most time-consuming part of both methods is dominated by adjacency-based operations (e.g., matrix multiplications with the graph Laplacian or adjacency matrix), which scale on the order of $\mathcal{O}(|E|\,k)$ or $\mathcal{O}(N^2 k)$ for large, dense graphs (where $N$ is the number of nodes, $E$ is the set of edges, and $k$ is the number of clusters). In contrast, the extra DPR calculations—computing entropy across nodes, the variance of cluster assignments, or centroid distances—represent only $\mathcal{O}(N\,k + k^2 F)$ overhead (with $F$ being the feature dimension) and are typically negligible next to the larger matrix multiplications.

\subsection{Limitations and Future Directions}

While DMoN-DPR introduces clear benefits, it also presents a few limitations. First, tuning the diversity-preserving weights for distance, variance, and entropy remains manual. Though our empirical studies revealed consistent patterns: high-dimensional datasets like Coauthor CS and Physics benefited from larger distance weights (1 or 10), variance weights of 1 or 0.1 worked well across most datasets, and entropy value of 0.1 served as a good upper bound. A promising direction for future work is to develop an adaptive scheme that learns these weights during training, reducing reliance on manual tuning. Second, the gains from diversity regularization are more pronounced on feature-rich datasets. On lower-dimensional datasets like PubMed, improvements in NMI and F1 are modest—though visualizations (Figure~\ref{fig:clusters}) suggest that DMoN–DPR still forms more coherent clusters than the baseline. Future research could explore alternative regularizers to better capture structure in such settings.

%% file: paper/gpu_results.tex
\begin{table*}[ht]
\centering
\small
\setlength{\tabcolsep}{4pt} 
\renewcommand{\arraystretch}{1.2} 
\caption{Comparison of clustering methods on three datasets (Cora, CiteSeer, PubMed). Values are in percentage.}
\begin{tabular}{@{}lcccccccccccc@{}}
\toprule
\multirow{2}{*}{\textbf{Method}} & \multicolumn{4}{c}{\textbf{Cora}} & \multicolumn{4}{c}{\textbf{CiteSeer}} & \multicolumn{4}{c}{\textbf{PubMed}} \\
\cmidrule(lr){2-5} \cmidrule(lr){6-9} \cmidrule(lr){10-13}
 & \multicolumn{2}{c}{\textbf{Graph}} & \multicolumn{2}{c}{\textbf{Labels}} & \multicolumn{2}{c}{\textbf{Graph}} & \multicolumn{2}{c}{\textbf{Labels}} & \multicolumn{2}{c}{\textbf{Graph}} & \multicolumn{2}{c}{\textbf{Labels}} \\
 & \( C \downarrow \) & \( Q \uparrow \) & NMI \( \uparrow \) & F1 \( \uparrow \) & \( C \downarrow \) & \( Q \uparrow \) & NMI \( \uparrow \) & F1 \( \uparrow \) & \( C \downarrow \) & \( Q \uparrow \) & NMI \( \uparrow \) & F1 \( \uparrow \) \\
\midrule
DiffPool & 15.99 & 62.78 & 40.13 & 46.55 & 7.92 & 66.69 & \textbf{33.40} & \textbf{47.83} & \multicolumn{4}{c}{--} \\
MinCut & 13.65 & 71.79 & 37.74 & 39.15 & 6.19 & 75.21 & 25.21 & 35.28 & 11.14 & 54.66 & \textbf{22.48} & 41.31 \\
DMoN &  \underline{10.53} & \textbf{72.77} & 43.92 & 46.93 & \textbf{4.86} & \underline{75.45} & 29.67 & 42.46 & 8.61 & 57.13 & \underline{22.39} & 43.23 \\
\midrule
DPR(D) & 10.95 & 72.20 & 43.98 & \underline{47.51} & 4.95 & \textbf{75.56} & 30.09 & 42.46 & \textbf{8.60} & \textbf{57.14} & 22.38 & \underline{43.24}  \\
DPR(V)  & 11.87 & 70.54 & 43.34 & 46.39 & 4.95 & 75.17 & 29.76 & 42.85 & 8.61 & 57.13 & \underline{22.39} & 43.23	 \\
DPR(E) &  \textbf{10.49} & \textbf{72.77} & 44.00 & 47.02 & \underline{4.93} & 75.25 & 29.95 & 42.96 & 8.62 & 57.11 & 22.36 & 43.22	 \\
DPR(DV) & 10.89 & 72.07 & \textbf{44.40} & 47.36 & 5.09 & 75.18 & \underline{30.50} & 43.25 & \textbf{8.60} & \textbf{57.14} & 22.38 & \underline{43.24}  \\
DPR(DE) & 10.88 & 72.30 & 44.30 & \textbf{47.53} & 5.07 & 75.18 & 30.47 & 43.13 & \textbf{8.60} & 57.12 & 22.38 & \textbf{43.25}	\\
DPR(VE) & 11.86 & 70.51 & 43.24 & 46.36 & 5.05 & 75.01 & 30.17 & \underline{43.27} & \textbf{8.60} & 57.13 & 22.37 & 43.21	 \\
DPR(DVE) & 10.93 & 72.05 & \underline{44.37} & 47.32 & 5.16 & 75.00 & 30.25 & 43.00 & 8.61 & 57.11 & 22.36 & \underline{43.24}	\\
\bottomrule
\end{tabular}
\label{tab:comparison_1}
\end{table*}

\begin{table*}[ht]
\centering
\small
\setlength{\tabcolsep}{4pt} 
\renewcommand{\arraystretch}{1.2} 
\caption{Comparison of clustering methods on two datasets (Coauthor CS, Coauthor Physics). Values are in percentage.}
\begin{tabular}{@{}lcccccccc@{}}
\toprule
\multirow{2}{*}{\textbf{Method}} & \multicolumn{4}{c}{\textbf{Coauthor CS}} & \multicolumn{4}{c}{\textbf{Coauthor Physics}} \\
\cmidrule(lr){2-5} \cmidrule(lr){6-9}
 & \multicolumn{2}{c}{\textbf{Graph}} & \multicolumn{2}{c}{\textbf{Labels}} & \multicolumn{2}{c}{\textbf{Graph}} & \multicolumn{2}{c}{\textbf{Labels}} \\
 & \( C \downarrow \) & \( Q \uparrow \) & NMI \( \uparrow \) & F1 \( \uparrow \) & \( C \downarrow \) & \( Q \uparrow \) & NMI \( \uparrow \) & F1 \( \uparrow \) \\
\midrule
DiffPool & \textbf{18.19} & 62.24 & 53.92 & 53.47 & 13.91 & 57.44 & \textbf{56.22} & 51.71 \\
MinCut & 21.36 &	71.58 &	64.33 &	49.00 & 13.93 &	\underline{61.75} &	51.39 &	47.79 \\
DMoN & \underline{18.63} & \textbf{72.60} & 69.26 & 59.26 & 13.70 & \textbf{63.45} & 53.50 & 47.51	 \\
\midrule
DPR(D) & 18.89 & 72.30 & 71.17 & 61.82	& 16.44 & 56.89 & 53.49 & 50.99  \\
DPR(V) & 18.75 & \underline{72.50} & 69.56 & 59.94 & 13.18 & 58.59 & 53.93 & 52.78  \\
DPR(E) & 19.85 & 70.92 & \textbf{71.58} & 61.33 & 13.49 & 59.33 & 52.83 & 51.09  \\
DPR(DV) & 19.14 & 71.96 & 70.72 & 61.35	& 13.43 & 56.48 & \underline{55.84} & \textbf{57.99} \\
DPR(DE) & 19.81 & 70.57 & 70.96 & \underline{62.36} & 14.17 & 56.89 & 54.02 & 55.02	 \\
DPR(VE) & 19.89 & 70.81 & \underline{71.47} & 61.33 & \underline{13.11} & 55.88 & 49.95 & 53.58	 \\
DPR(DVE) & 19.69 & 70.71 & 71.28 & \textbf{62.67} & \textbf{12.84} & 55.47 & 53.50 & \underline{57.96}	 \\
\bottomrule
\end{tabular}
\label{tab:comparison_2}
\end{table*}

\begin{table}[ht]
\centering
\small
\caption{t-statistics and p-values for all datasets (rounded to two decimal places).}
\label{tab:sig_test_combined}
\begin{tabular}{lccccccccccc}
\toprule
\textbf{Metric} 
& \multicolumn{2}{c}{\textbf{Cora}} 
& \multicolumn{2}{c}{\textbf{CiteSeer}} 
& \multicolumn{2}{c}{\textbf{PubMed}} 
& \multicolumn{2}{c}{\textbf{Coauthor CS}} 
& \multicolumn{2}{c}{\textbf{Coauthor Physics}} \\
\cmidrule(lr){2-3} \cmidrule(lr){4-5} \cmidrule(lr){6-7} \cmidrule(lr){8-9} \cmidrule(lr){10-11}
 & t-stat & p-val & t-stat & p-val & t-stat & p-val & t-stat & p-val & t-stat & p-val \\ 
\midrule
Conductance & -0.88 & 0.40 & -1.62 & 0.14 & 0.37 & 0.72 & -5.25 & $1e^{-6}$ & 0.51 & 0.62 \\ 
Modularity  & 1.81  & 0.10 & 1.17  & 0.27 & 0.21 & 0.84 & 5.88  & $1e^{-6}$ & 17.56 & $1e^{-6}$ \\ 
NMI         & -0.63 & 0.54 & -1.56 & 0.15 & 0.23 & 0.83 & -4.64 & $1e^{-6}$ & -2.49 & 0.03 \\ 
F1          & -0.40 & 0.70 & -1.28 & 0.23 & -0.95 & 0.37 & -2.51 & 0.03 & -14.65 & $1e^{-6}$ \\
\bottomrule
\end{tabular}
\end{table}

%% file: paper/conclusion.tex
\section{Conclusion}

In this work, we presented DMoN–DPR, an enhanced version of Deep Modularity Networks (DMoN) that incorporates diversity-preserving regularizations to enrich the clustering objective. By introducing distance-, entropy-, and variance-based penalties alongside the original modularity and collapse regularization terms, our approach promotes diversity by increasing inter-cluster feature separation (distance) and within-cluster assignment dispersion (variance), while a small-weight entropy term gradually sharpens assignments without sacrificing exploration. Our empirical results on the Coauthor CS and Coauthor Physics benchmark datasets indicate that DMoN–DPR significantly improves alignment with ground-truth labels in feature-rich datasets. While structural metrics such as modularity and conductance remained competitive, the additional diversity constraints contributed to more interpretable and semantically meaningful cluster formations, marked by significant improvements in NMI and F1 scores, producing higher alignment with ground‐truth labels and generally reflecting clearer semantic distinctions. Overall, DMoN-DPR achieves a balance between interpretability and structural integrity, making it a powerful solution for graph clustering tasks that demand both topological cohesion and semantic differentiation, marking a significant advancement in feature-aware graph clustering.

%% file: paper/checklist.tex
\section*{NeurIPS Paper Checklist}

\begin{enumerate}

\item {\bf Claims}
    \item[] Question: Do the main claims made in the abstract and introduction accurately reflect the paper's contributions and scope?
    \item[] Answer: \answerYes{} 
    \item[] Justification: We identified a key limitation of DMoN—namely, that it often yields clusters with insufficient feature‐space diversity—and addressed it by introducing three novel regularization terms that are distance, variance, and entropy based. In our paper, we have shown that the impact on label‐based metrics can be substantial on feature‐rich datasets such as Coauthor CS and Coauthor Physics, where DMoN--DPR significantly improves NMI and F1 compared to baseline DMoN. This demonstrates that our approach works especially well in scenarios where node attributes carry high signal, thus expanding DMoN’s applicability to more attribute‐driven clustering tasks.

\item {\bf Limitations}
    \item[] Question: Does the paper discuss the limitations of the work performed by the authors?
    \item[] Answer:\answerYes{} 
    \item[] Justification: In Section 5.1, we clearly highlight the trade-off between DMoN and DMoN--DPR, outlining the scenarios in which each is most suitable. Furthermore, Section 5.2 provides a detailed discussion of the limitations of our approach and potential directions for future work. We have also included a runtime analysis in Appendix D. 

\item {\bf Theory assumptions and proofs}
    \item[] Question: For each theoretical result, does the paper provide the full set of assumptions and a complete (and correct) proof?
    \item[] Answer: \answerNA{} 
    \item[] Justification: Our main contribution is an empirical extension of DMoN that aims to preserve feature diversity, rather than a new theoretical framework. \citet{tsitsulin2023graph} showed that DMoN’s collapse regularization avoids trivial solutions and is asymptotically consistent under certain generative assumptions. Our distance-, variance-, and entropy-based regularizers further promote diverse, well-separated clusters without disrupting DMoN’s convergence. Each term is differentiable (or piecewise differentiable), ensuring compatibility with gradient-based optimization. Specifically:

\begin{itemize}
    \item $R_D$ (distance) repels centroids in feature space, reducing cluster overlap.
    \item $R_V$ (variance) prevents overly tight clusters, mitigating premature ``hard'' assignments.
    \item $R_E$ (entropy) avoids early dominance by any single cluster, helping to avert mode collapse.
\end{itemize}

These terms add complementary constraints but do not create new problematic local minima or break DMoN’s consistency. Rather, they refine the solution space by adding complementary constraints. While we do not offer formal proofs here, we consistently observe stable training across all studied datasets.


    \item {\bf Experimental result reproducibility}
    \item[] Question: Does the paper fully disclose all the information needed to reproduce the main experimental results of the paper to the extent that it affects the main claims and/or conclusions of the paper (regardless of whether the code and data are provided or not)?
    \item[] Answer: \answerYes{} 
    \item[] Justification: All ablation studies, seed values, and hyperparameter settings are thoroughly documented in the appendix.

\item {\bf Open access to data and code}
    \item[] Question: Does the paper provide open access to the data and code, with sufficient instructions to faithfully reproduce the main experimental results, as described in supplemental material?
    \item[] Answer: \answerYes{} 
    \item[] Justification: The datasets used in this paper are standard benchmarks, all publicly available through the PyTorch Geometric library. The code will also be released publicly.

\item {\bf Experimental setting/details}
    \item[] Question: Does the paper specify all the training and test details (e.g., data splits, hyperparameters, how they were chosen, type of optimizer, etc.) necessary to understand the results?
    \item[] Answer: \answerYes{} 
    \item[] Justification: All the details have been discussed in the appendix. 

\item {\bf Experiment statistical significance}
    \item[] Question: Does the paper report error bars suitably and correctly defined or other appropriate information about the statistical significance of the experiments?
    \item[] Answer: \answerYes{} 
    \item[] Justification: The result for all trial runs as well as their mean and standard deviation has been reported in the appendix. Additionally, statistical significance testing has also been reported in the Results (Section 5.1). 

\item {\bf Experiments compute resources}
    \item[] Question: For each experiment, does the paper provide sufficient information on the computer resources (type of compute workers, memory, time of execution) needed to reproduce the experiments?
    \item[] Answer: \answerYes{} 
    \item[] Justification: The hardware and implementation settings have been mentioned in the appendix as well as in Section 4.
    
\item {\bf Code of ethics}
    \item[] Question: Does the research conducted in the paper conform, in every respect, with the NeurIPS Code of Ethics \url{https://neurips.cc/public/EthicsGuidelines}?
    \item[] Answer: \answerYes{} 
    \item[] Justification: The work uses public, consented academic-citation and co-authorship data and involves no sensitive attributes or human subjects.

\item {\bf Broader impacts}
    \item[] Question: Does the paper discuss both potential positive societal impacts and negative societal impacts of the work performed?
    \item[] Answer: \answerYes{} 
    \item[] Justification: Improved graph clustering can aid knowledge discovery in scientific networks (positive), but could also facilitate surveillance or profiling if applied to social graphs (negative). 
    
\item {\bf Safeguards}
    \item[] Question: Does the paper describe safeguards that have been put in place for responsible release of data or models that have a high risk for misuse (e.g., pretrained language models, image generators, or scraped datasets)?
    \item[] Answer: \answerNA{}   
    \item[] Justification: The released code trains only on publicly available graphs and does not include any pre-trained model with dual-use risk.

\item {\bf Licenses for existing assets}
    \item[] Question: Are the creators or original owners of assets (e.g., code, data, models), used in the paper, properly credited and are the license and terms of use explicitly mentioned and properly respected?
    \item[] Answer: \answerYes{} 
    \item[] Justification: The paper cites PyTorch Geometric and references each dataset with its original publication and Creative Commons/licensing where applicable.

\item {\bf New assets}
    \item[] Question: Are new assets introduced in the paper well documented and is the documentation provided alongside the assets?
    \item[] Answer: \answerNA{} 
    \item[] Justification: No new datasets or pre-trained models are released; only source code is provided.

\item {\bf Crowdsourcing and research with human subjects}
    \item[] Question: For crowdsourcing experiments and research with human subjects, does the paper include the full text of instructions given to participants and screenshots, if applicable, as well as details about compensation (if any)? 
    \item[] Answer: \answerNA{} 
    \item[] Justification: The study uses only machine-generated data; no human participants were involved.

\item {\bf Institutional review board (IRB) approvals or equivalent for research with human subjects}
    \item[] Question: Does the paper describe potential risks incurred by study participants, whether such risks were disclosed to the subjects, and whether Institutional Review Board (IRB) approvals (or an equivalent approval/review based on the requirements of your country or institution) were obtained?
    \item[] Answer: \answerNA{} 
    \item[] Justification: Not applicable—no human-subject research.

\item {\bf Declaration of LLM usage}
    \item[] Question: Does the paper describe the usage of LLMs if it is an important, original, or non-standard component of the core methods in this research? Note that if the LLM is used only for writing, editing, or formatting purposes and does not impact the core methodology, scientific rigorousness, or originality of the research, declaration is not required.
    \item[] Answer: \answerNA{} 
    \item[] Justification: No large-language models are part of the methodology; any text editing assistance was purely editorial.

\end{enumerate}


%% file: paper/appendix.tex
\section{Datasets}
\label{sec:datasets}

The Cora, CiteSeer, and PubMed datasets are citation networks where nodes represent papers and edges denote citation relationships, with node labels corresponding to the topic of each paper. The Coauthor CS and Coauthor Physics datasets are co-authorship graphs, where nodes represent authors and edges indicate collaborations, with labels reflecting the field of research. The following table provides information about the datasets used for running experiments. 

\paragraph{Feature-rich Datasets} A graph is considered to be feature-rich when its nodes are described by high-dimensional feature vectors that carry substantial information content. Formally, for each node $v$, the feature vector $x_v$ is normalized into a probability distribution $p_v = \frac{x_v}{\mathbf{1}^\top x_v},$ and its information content is quantified by the Shannon entropy $H(p_v) = -\sum_{i=1}^F p_{v,i} \log_2 p_{v,i}.$ A dataset of such nodes is said to be feature-rich when it combines two key properties: (1) high dimensionality, meaning that the number of features $|X| = F$ is large, and (2) high average entropy, defined as: 

$$\frac{1}{N}\sum_v H(p_v),$$

indicating that on average, the features across nodes are diverse and well-distributed rather than concentrated in a few dimensions. Together, these conditions ensure that the feature space provides both breadth and depth, enabling richer representations for learning tasks. For example, Coauthor‑CS is a feature-rich dataset for having $|X|=6805$ and entropy = 5.51 bits, versus PubMed with $|X|=500$ and entropy = 5.17 bits.

\begin{table}[!htbp]
    \centering
    \small
    \caption{Dataset characteristics, where ($|V|$) is the number of vertices , ($|E|$) is the number of edges, $|X|$ is the number of features, and $|Y|$ is the number of cluster labels.}
    \begin{tabular}{cccccc}
    \toprule
        Dataset & $|V|$ & $|E|$ & $|X|$ & $|Y|$ & Mean Entropy (bits) \\
        \toprule
        Cora & 2708 & 5278 & 1433 & 7 & 4.05 \\
        CiteSeer & 3327 & 4614 & 3703 & 6 & 4.94 \\
        PubMed & 19717 & 44325 & 500 & 3 & 5.17 \\
        Coauthor CS & 18333 & 81894 & 6805 & 15 & 5.51 \\
        Coauthor Physics & 34493 & 247962 & 8415 & 5 & 4.76 \\
        \toprule
    \end{tabular}
    \label{tab:datasets}
\end{table}

\input{paper/results_appendix}

\section{Ablation Study}
\label{sec:ablation}

\subsection{Hyperparameter Tuning} 

\paragraph{DiffPool and MinCut} For DiffPool, the entropy weight was set to \(1 \times 10^{-5}\) on the Coauthor Physics dataset to achieve optimal performance, and to \(1 \times 10^{-4}\) for all other datasets. For MinCut pooling, both the mincut loss and orthogonality loss weights were set to 1, which consistently yielded the best results.

\paragraph{DMoN--DPR} The weighting coefficients \( W_{\text{dist}} \), \( W_{\text{entropy}} \), and \( W_{\text{var}} \), as well as $\epsilon$ are hyperparameters that need to be tuned based on the specific dataset and desired clustering behavior. They control the trade-off between structural modularity and diversity preservation. To optimize these hyperparameters, we conducted evaluations on the Cora, CiteSeer, PubMed, Coauthor CS and Coauthor Physics datasets using 10 random seeds, similar to how it was done in \citep{tsitsulin2023graph}, selected using a random number generator. The seeds that resulted in the best performance when using the DMoN pooling layer were selected. For each dataset, we tuned \( \epsilon \) (for the distance weight \( W_{\text{dist}} \)), \( W_{\text{var}} \) (variance weight), and \( W_{\text{entropy}} \) (entropy weight) independently. To find the best \( \epsilon \), the variance and entropy weights were set to zero, and \( \epsilon \) was varied from 10 to \( 10^{-5} \). Similarly, \( W_{\text{var}} \) and \( W_{\text{entropy}} \) were tuned by fixing the other weights to zero and varying them across \{1, 0.1, 0.01, 0.001\}. The best weights identified were used to construct various DMoN-DPR models, including DMoN (baseline), DMoN--DPR(D) (distance), DMoN--DPR(V) (variance), DMoN--DPR(E) (entropy), and combinations: DMoN--DPR(DV) (distance and variance), DMoN--DPR(DE) (distance and entropy), DMoN--DPR(VE) (variance and entropy), and DMoN--DPR(DVE) (distance, variance, and entropy). This systematic approach allowed us to assess the impact of each regularization term and their combinations on model performance. The ablation study is presented in the following subsections.

\subsubsection{Varying the Epsilon}

\begin{figure}[!h]
\centering
\small
\begin{subfigure}{0.30\textwidth}
  \includegraphics[width=\linewidth]{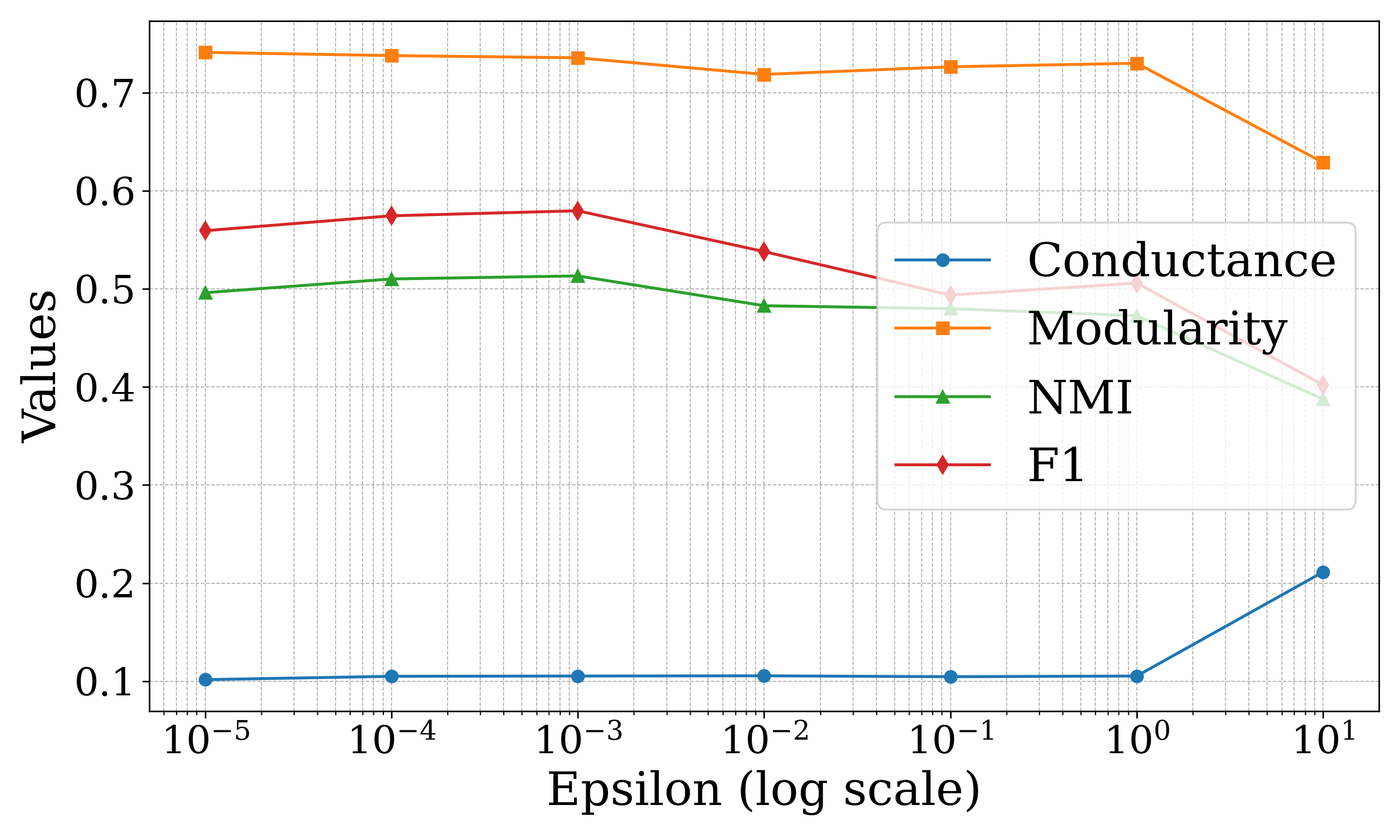}
  \captionsetup{justification=centering}
  \caption{}
  \label{fig:cora_eps}
\end{subfigure} 
\begin{subfigure}{0.30\textwidth}
  \includegraphics[width=\linewidth]{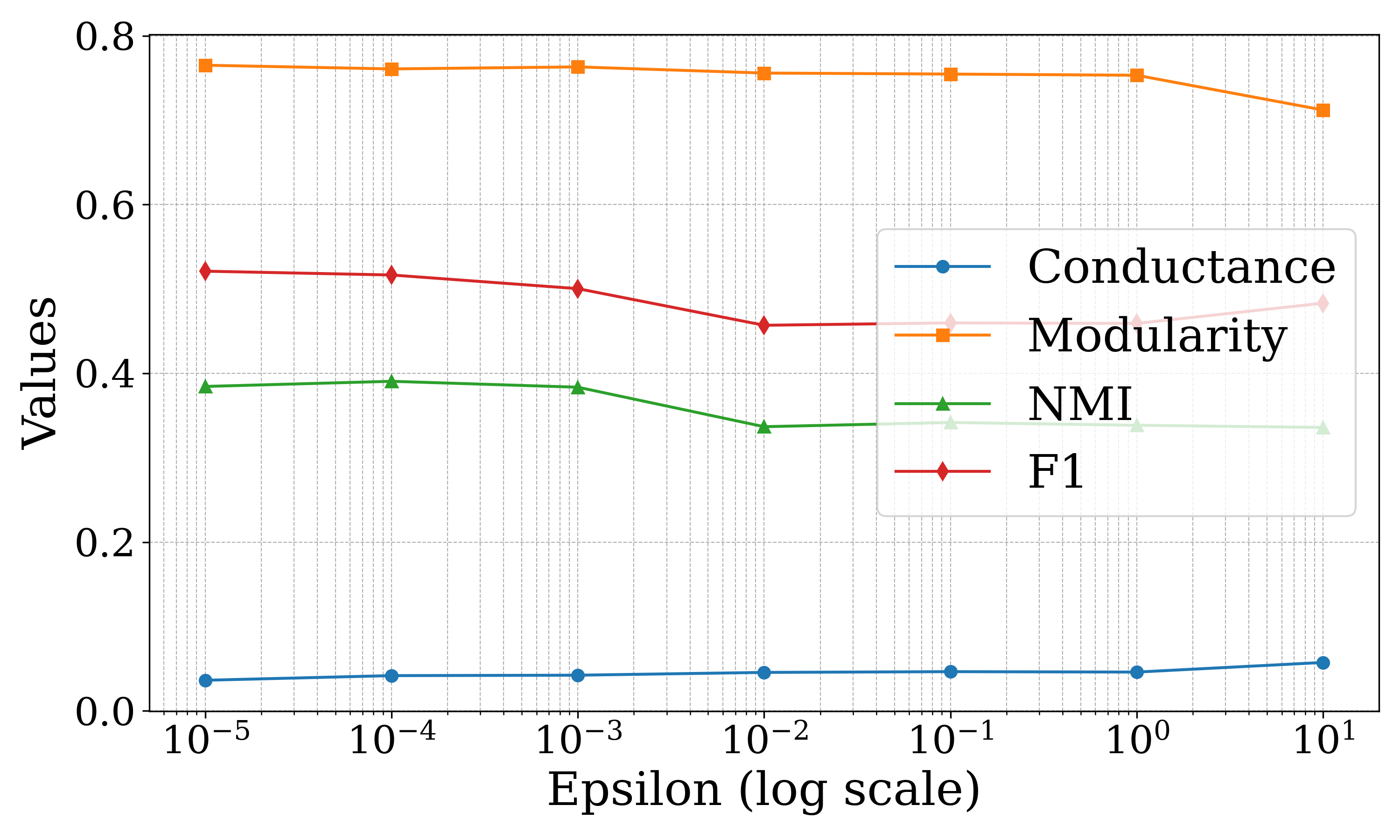}
  \captionsetup{justification=centering}
  \caption{}
  \label{fig:citeseet_eps}
\end{subfigure}
\begin{subfigure}{0.30\textwidth}
  \includegraphics[width=\linewidth]{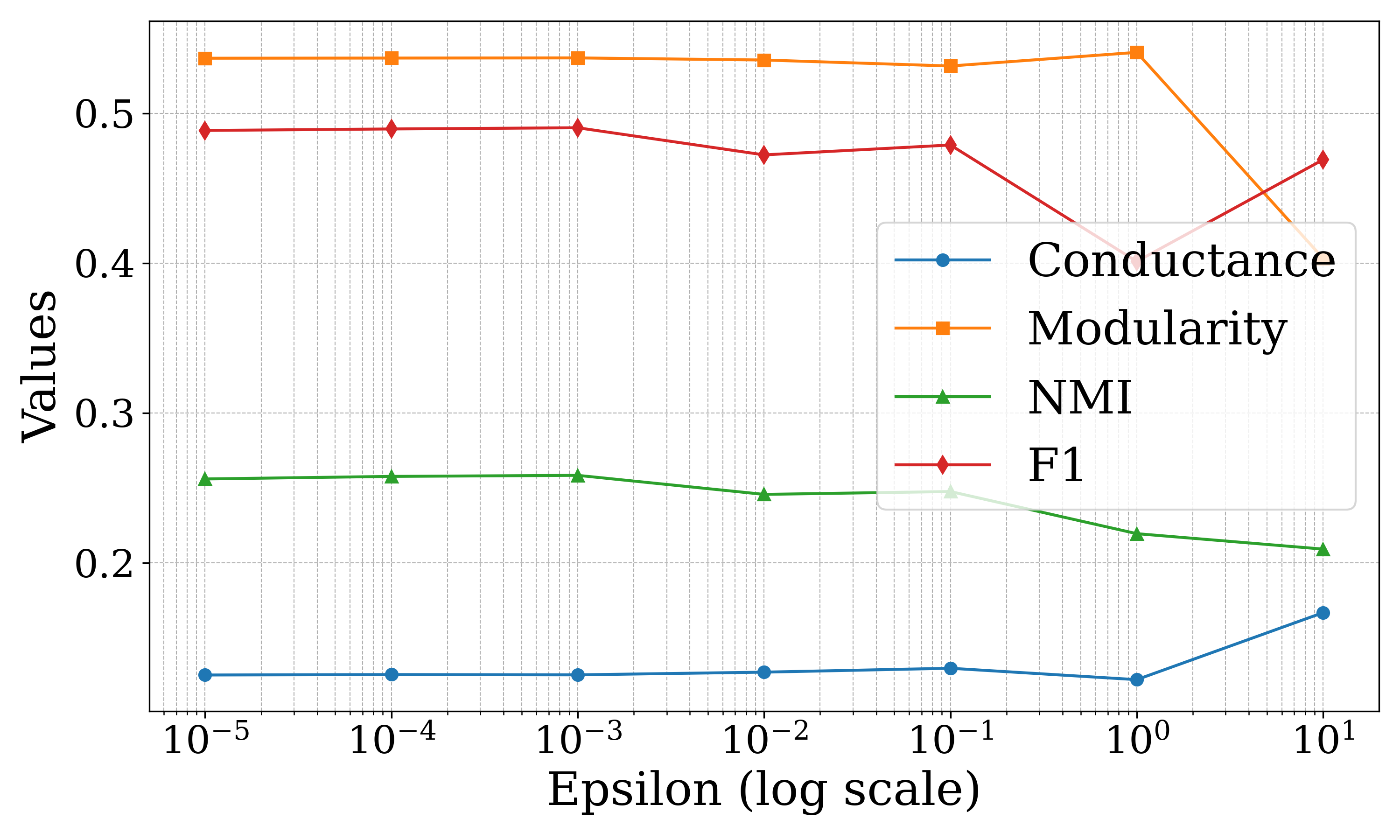}
  \captionsetup{justification=centering}
  \caption{}
  \label{fig:pubmed_eps}
\end{subfigure}
\begin{subfigure}{0.30\textwidth}
  \includegraphics[width=\linewidth]{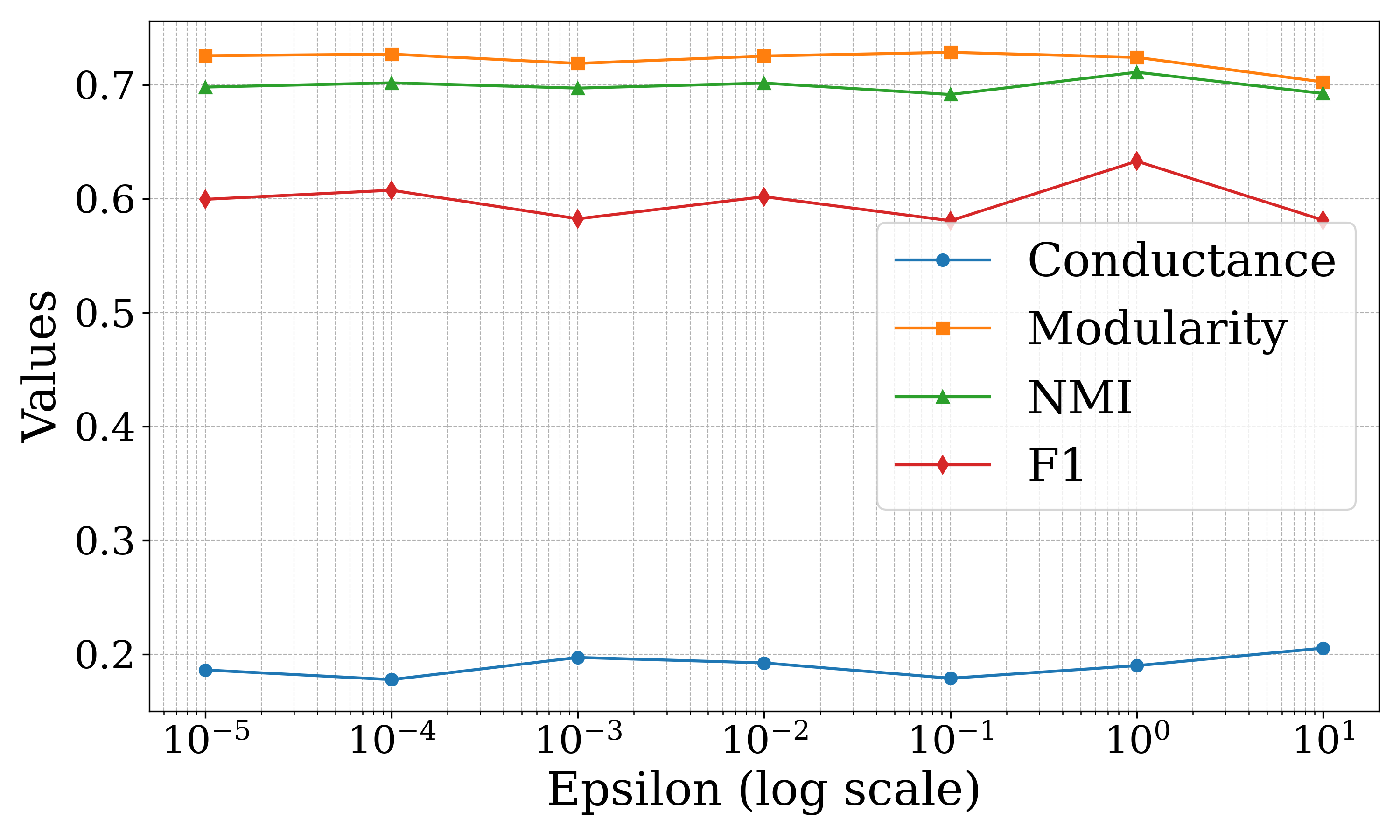}
  \captionsetup{justification=centering}
  \caption{}
  \label{fig:cs_eps}
\end{subfigure}
\begin{subfigure}{0.30\textwidth}
  \includegraphics[width=\linewidth]{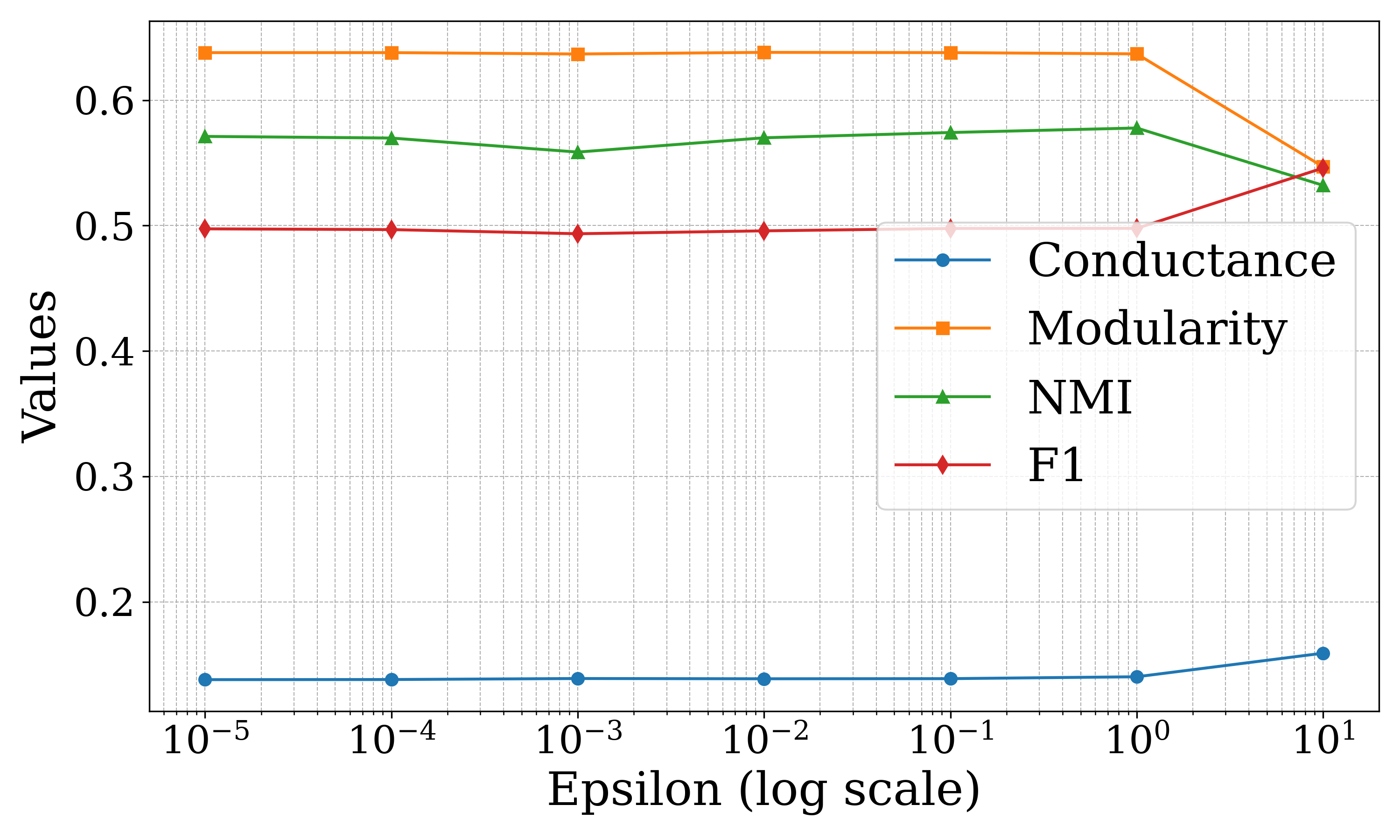}
  \captionsetup{justification=centering}
  \caption{}
  \label{fig:phys_eps}
\end{subfigure}
\caption{Finding the best value of epsilon ($\epsilon$) by setting \( W_{\text{dist}} \) to 1 and \( W_{\text{var}} \) and \( W_{\text{entropy}} \) to zero on the (a) Cora, (b) CiteSeer, (c) PubMed, (d) Coauthor CS, and (e) Coauthor Physics datasets.}
\label{fig:ablation_epsilon}
\end{figure}

Figure \ref{fig:ablation_epsilon} depicts the ablation study done to find the best value of $\epsilon$ associated with $W_{\text{dist}}$ by setting $W_{\text{var}}$ and $W_{\text{entropy}}$ weights to 0. For the Cora, CiteSeer, PubMed, Coauthor CS, and Coauthor Physics datasets, the best $\epsilon$ values were found to be $0.0001$, $0.0001$, $0.001$, $1.0$, and $10.0$ respectively.

\subsubsection{Varying the Variance Weight}

\begin{figure}[!h]
\centering
\small
\begin{subfigure}{0.30\textwidth}
  \includegraphics[width=\linewidth]{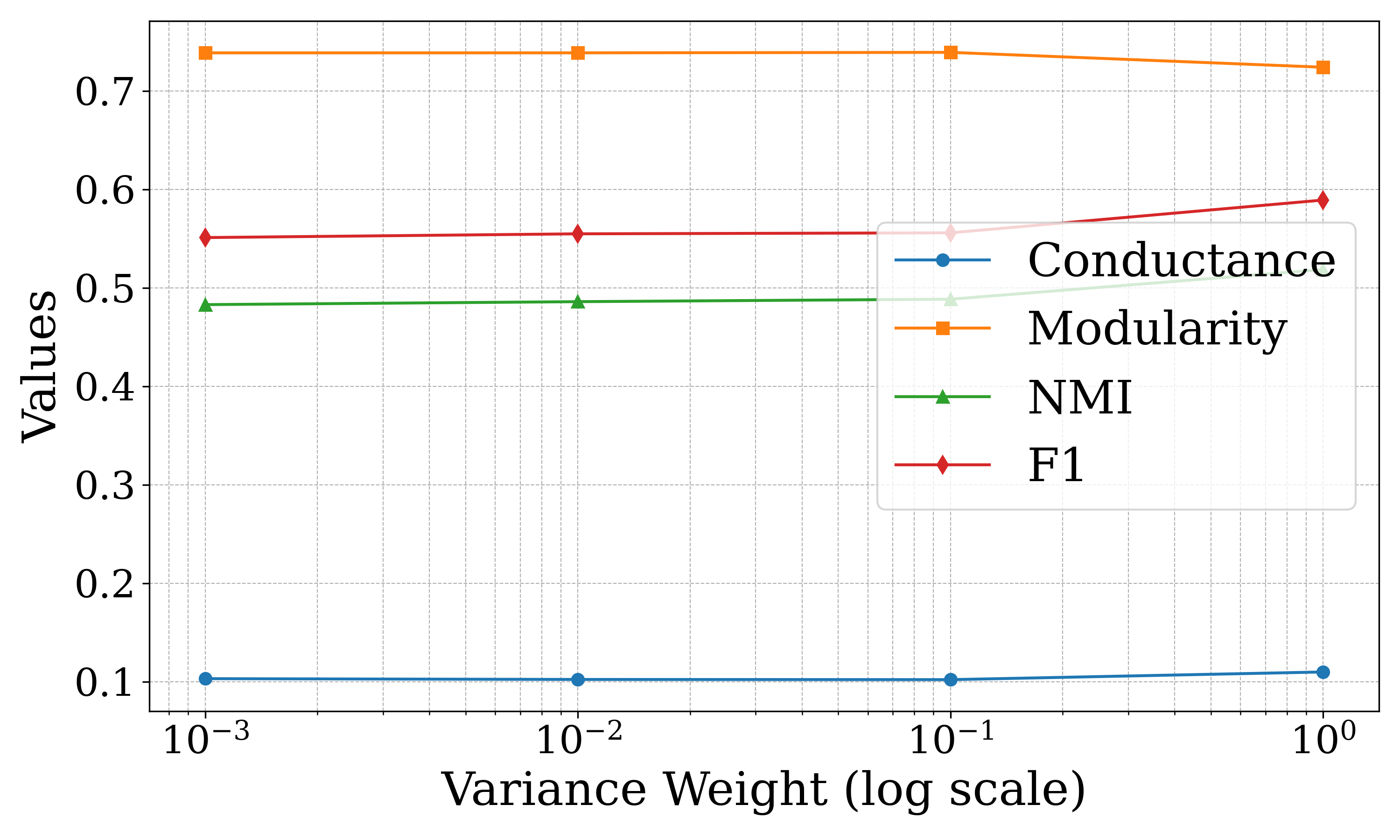}
  \captionsetup{justification=centering}
  \caption{}
  \label{fig:cora_var}
\end{subfigure} 
\begin{subfigure}{0.30\textwidth}
  \includegraphics[width=\linewidth]{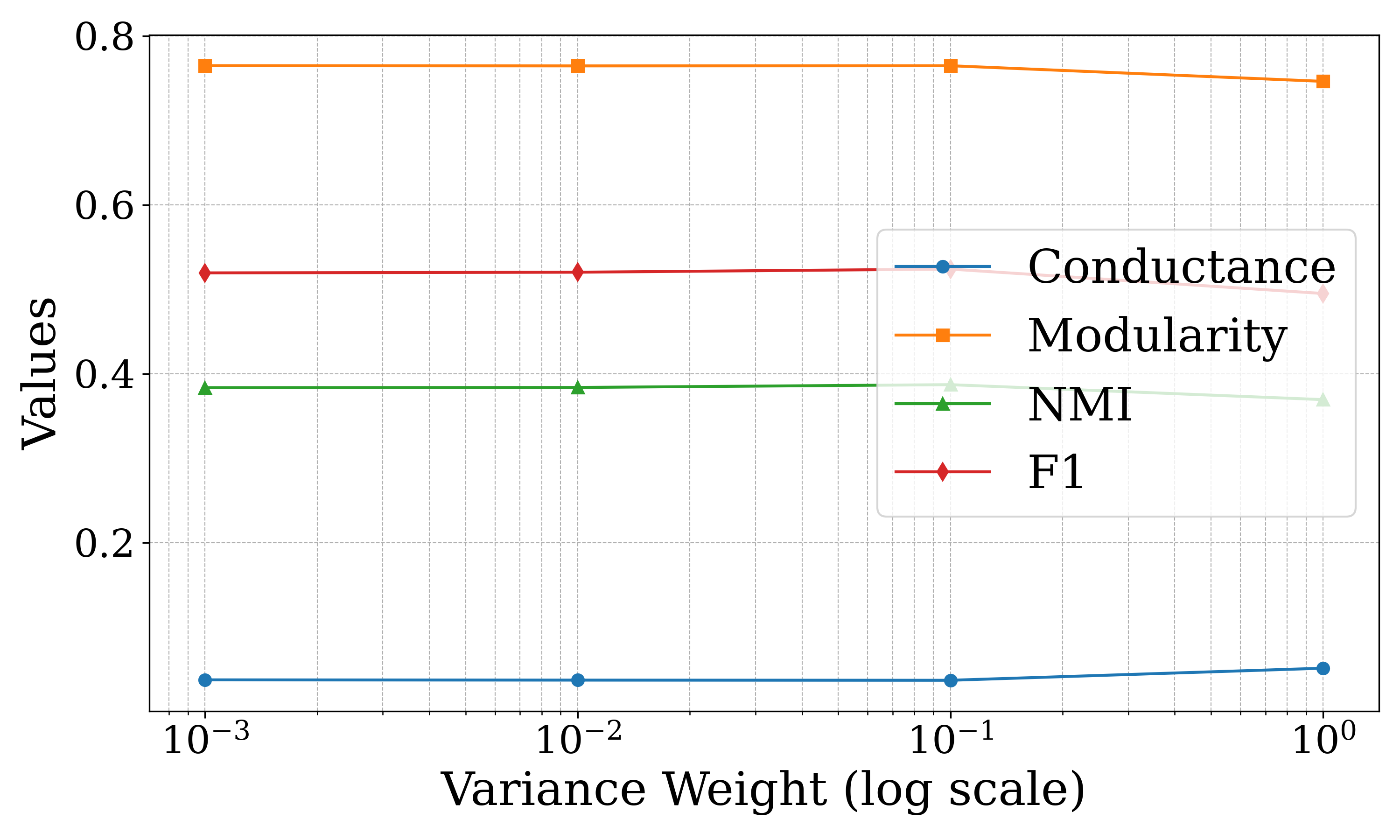}
  \captionsetup{justification=centering}
  \caption{}
  \label{fig:citeseet_var}
\end{subfigure}
\begin{subfigure}{0.30\textwidth}
  \includegraphics[width=\linewidth]{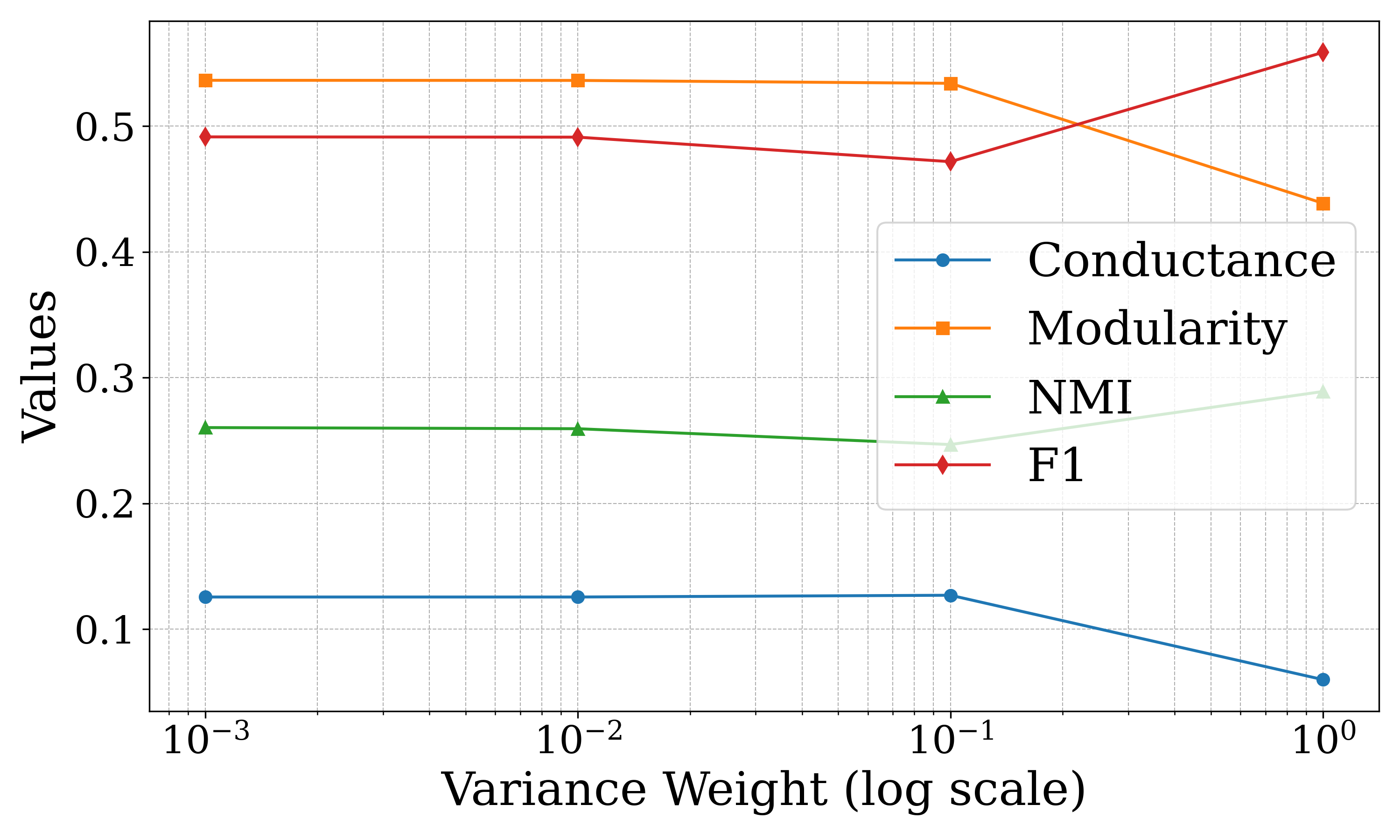}
  \captionsetup{justification=centering}
  \caption{}
  \label{fig:pubmed_var}
\end{subfigure}
\begin{subfigure}{0.30\textwidth}
  \includegraphics[width=\linewidth]{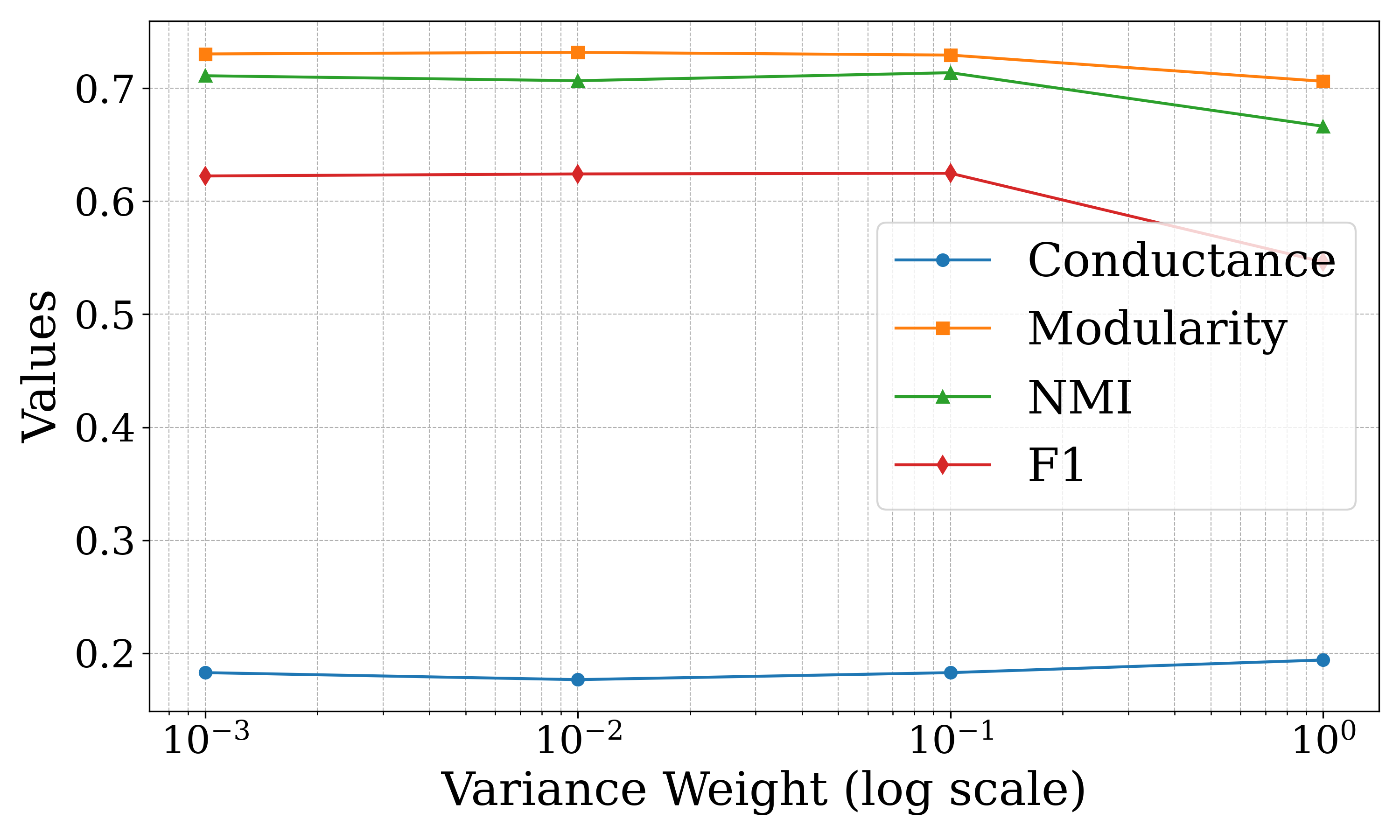}
  \captionsetup{justification=centering}
  \caption{}
  \label{fig:cs_var}
\end{subfigure}
\begin{subfigure}{0.30\textwidth}
  \includegraphics[width=\linewidth]{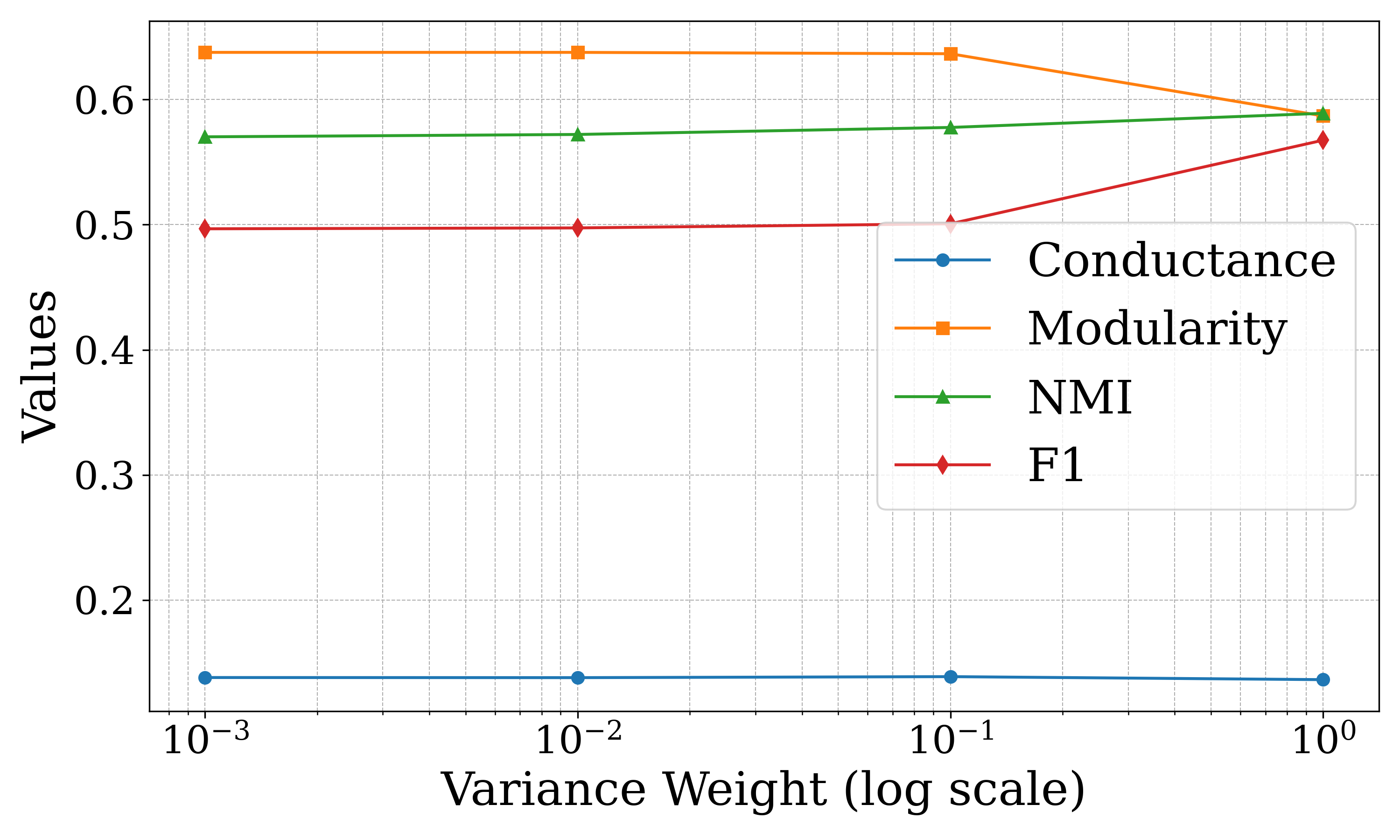}
  \captionsetup{justification=centering}
  \caption{}
  \label{fig:phys_var}
\end{subfigure}
\caption{Finding the best value of \( W_{\text{var}} \) by setting \( W_{\text{dist}} \) and \( W_{\text{entropy}} \) to zero on the (a) Cora, (b) CiteSeer, (c) PubMed, (d) Coauthor CS, and (e) Coauthor Physics datasets.}
\label{fig:ablation_variance}
\end{figure}

Figure \ref{fig:ablation_variance} depicts the ablation study done to find the best value of \( W_{\text{var}} \) by setting $W_{\text{dist}}$ and $W_{\text{entropy}}$ weights to 0. For the Cora, CiteSeer, PubMed, Coauthor CS, and Coauthor Physics datasets, the best \( W_{\text{var}} \) values were found to be $1$/$0.1$, $0.1$, $0.001$, $0.1$, and $1.0$ respectively.

\subsubsection{Varying the Entropy Weight}

\begin{figure}[!htbp]
\centering
\small
\begin{subfigure}{0.30\textwidth}
  \includegraphics[width=\linewidth]{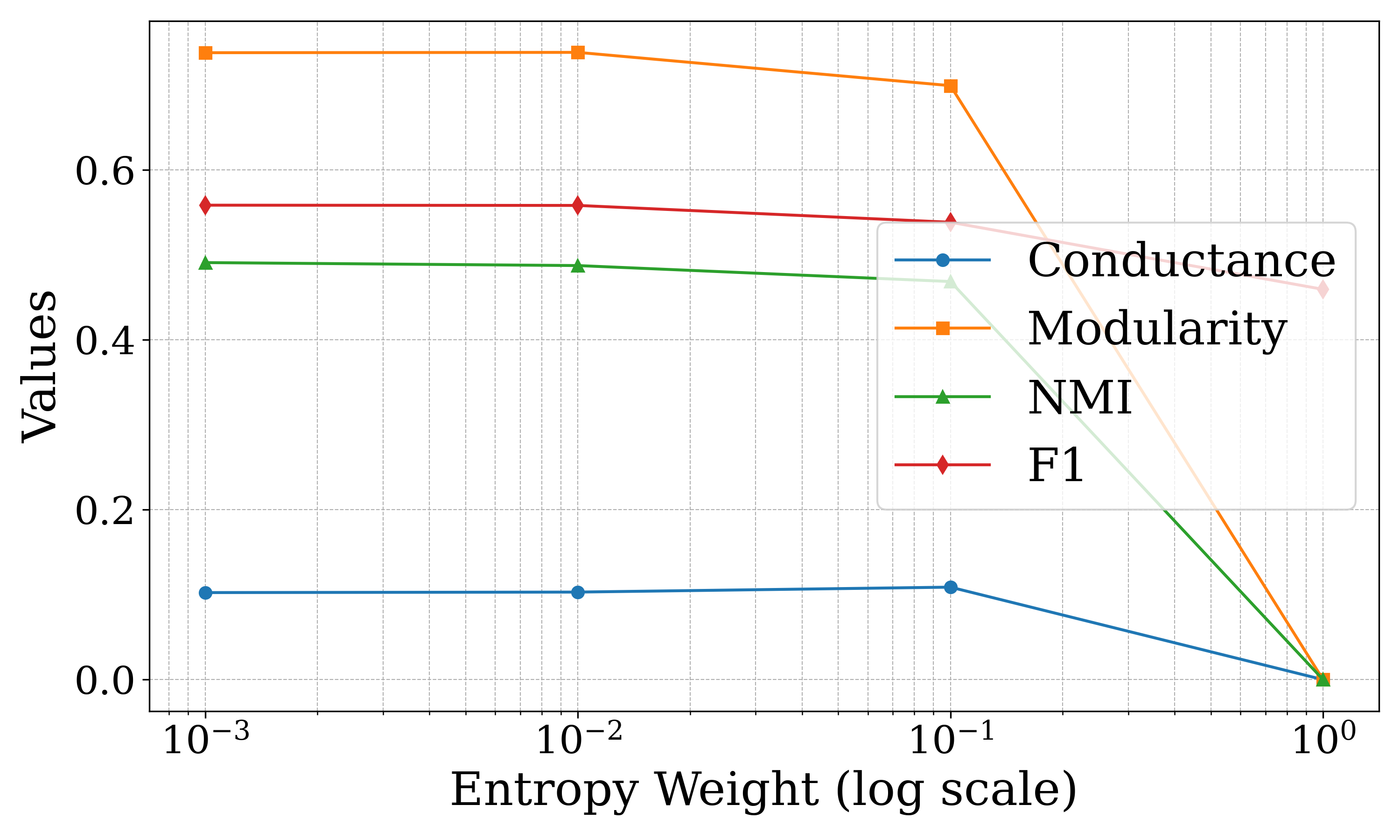}
  \captionsetup{justification=centering}
  \caption{}
  \label{fig:cora_ent}
\end{subfigure} 
\begin{subfigure}{0.30\textwidth}
  \includegraphics[width=\linewidth]{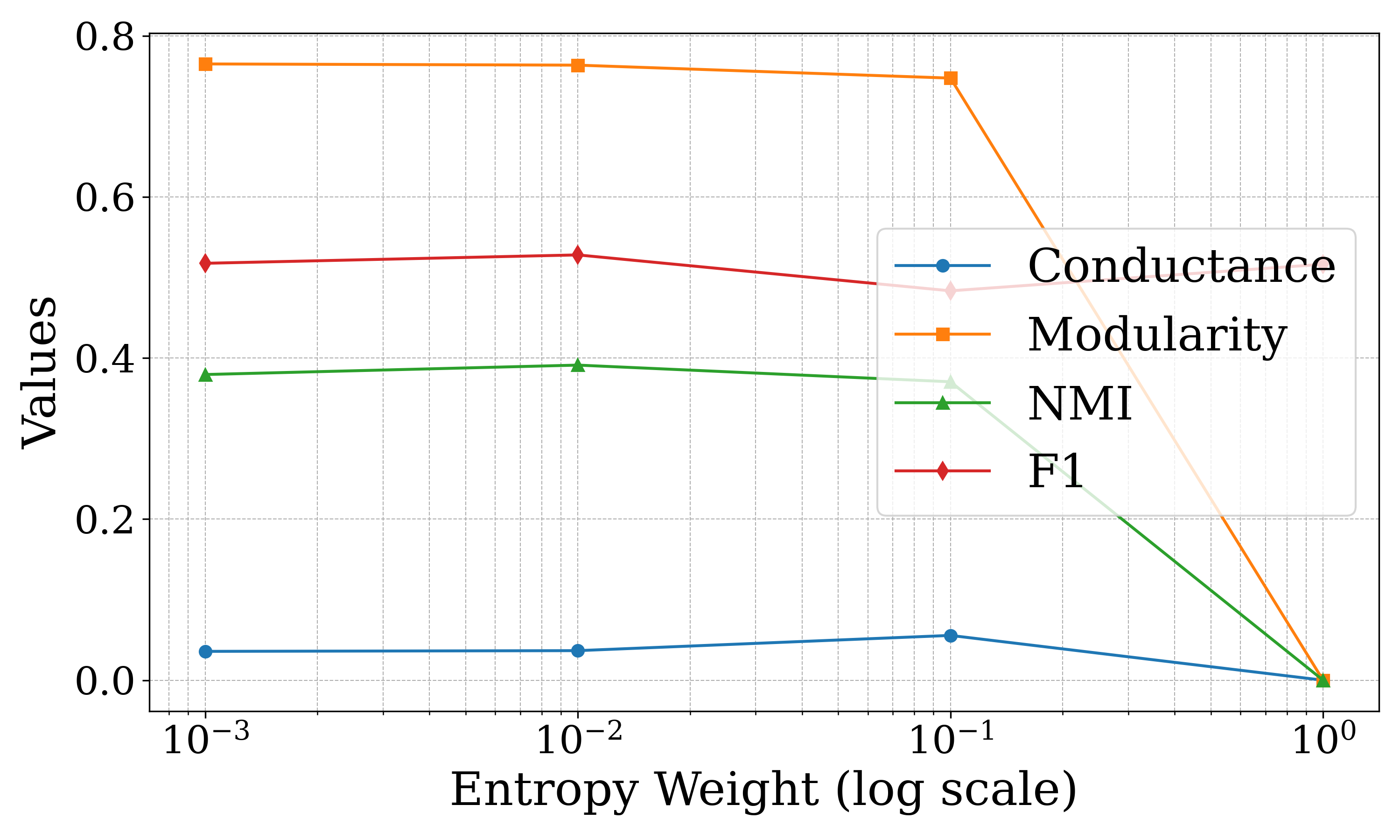}
  \captionsetup{justification=centering}
  \caption{}
  \label{fig:citeseet_ent}
\end{subfigure}
\begin{subfigure}{0.30\textwidth}
  \includegraphics[width=\linewidth]{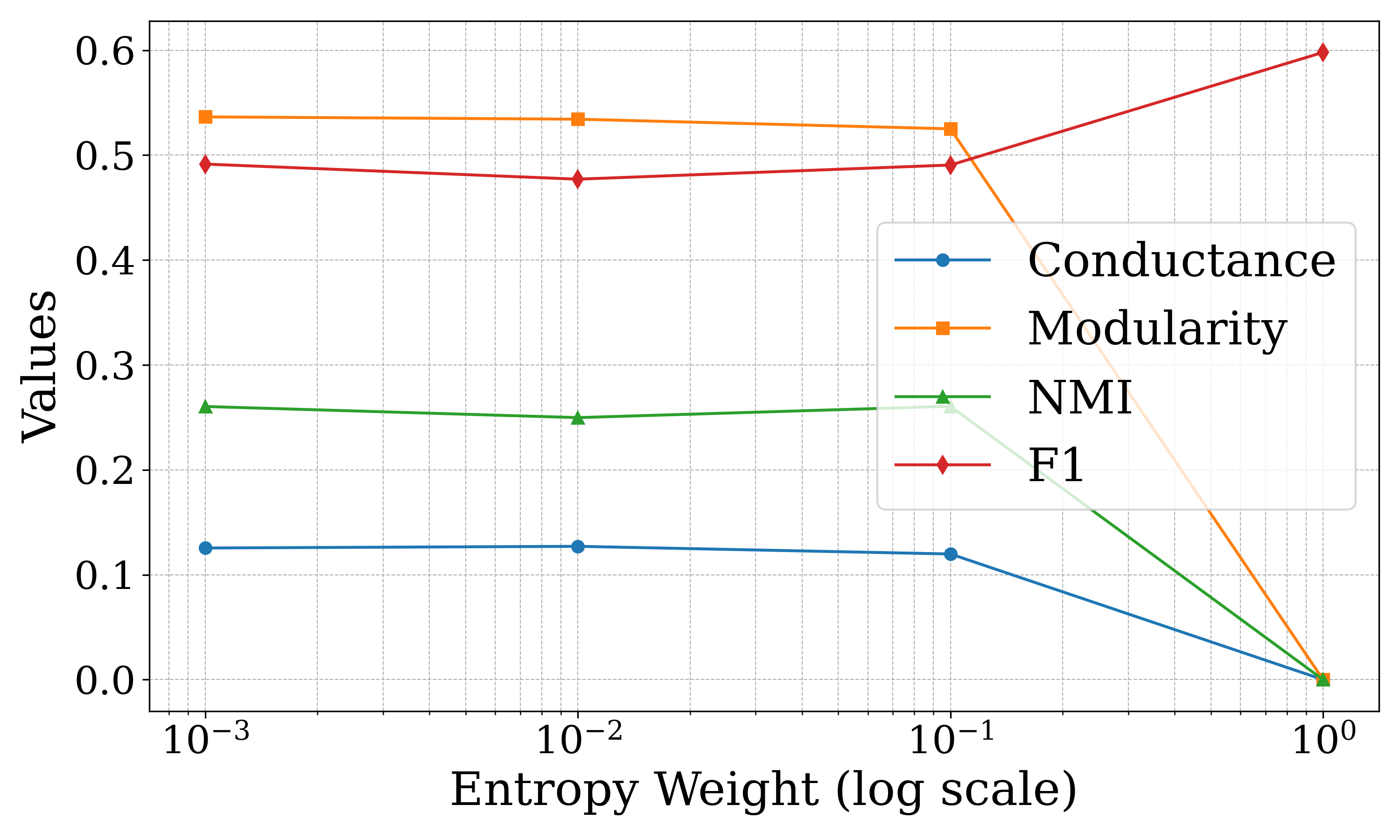}
  \captionsetup{justification=centering}
  \caption{}
  \label{fig:pubmed_ent}
\end{subfigure}
\begin{subfigure}{0.30\textwidth}
  \includegraphics[width=\linewidth]{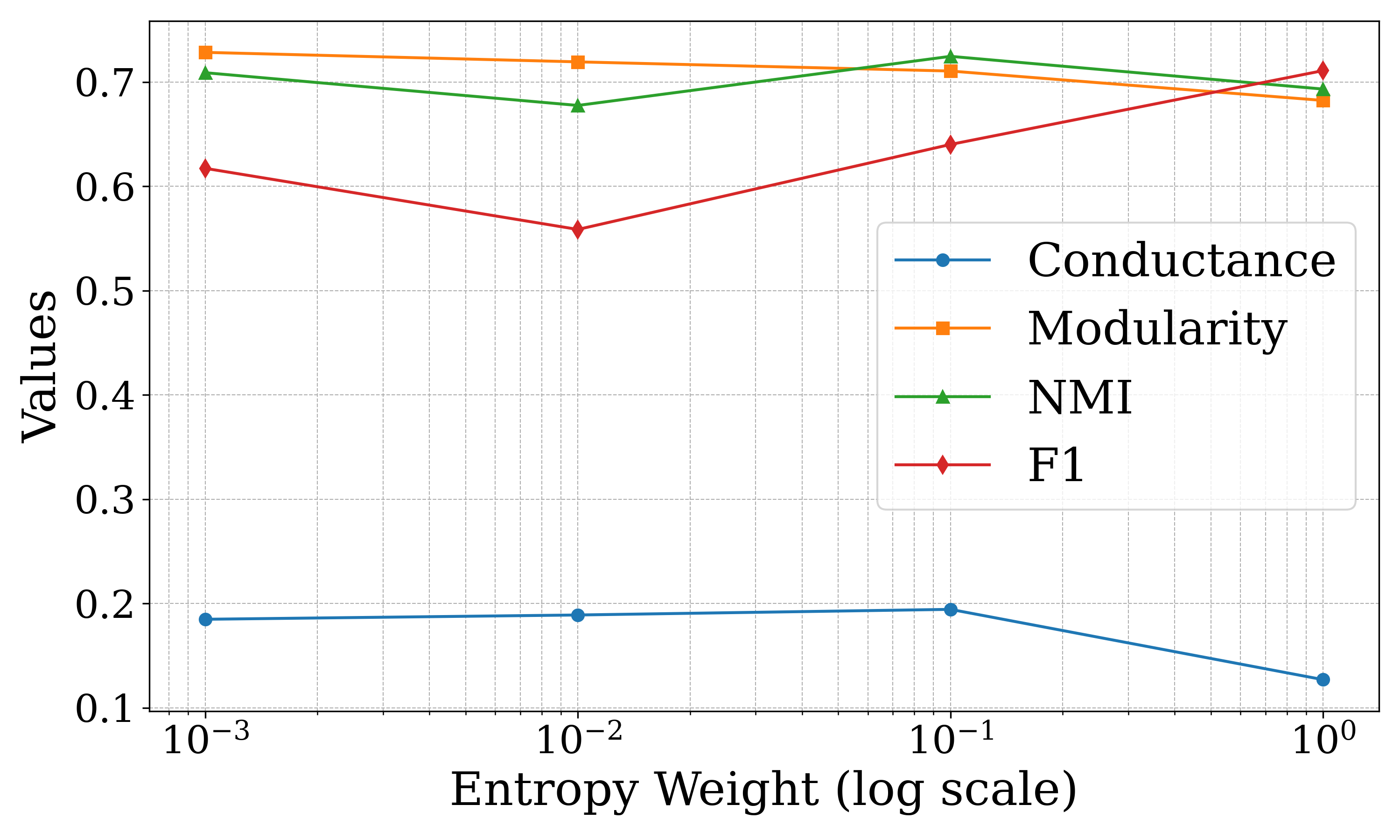}
  \captionsetup{justification=centering}
  \caption{}
  \label{fig:cs_ent}
\end{subfigure}
\begin{subfigure}{0.30\textwidth}
  \includegraphics[width=\linewidth]{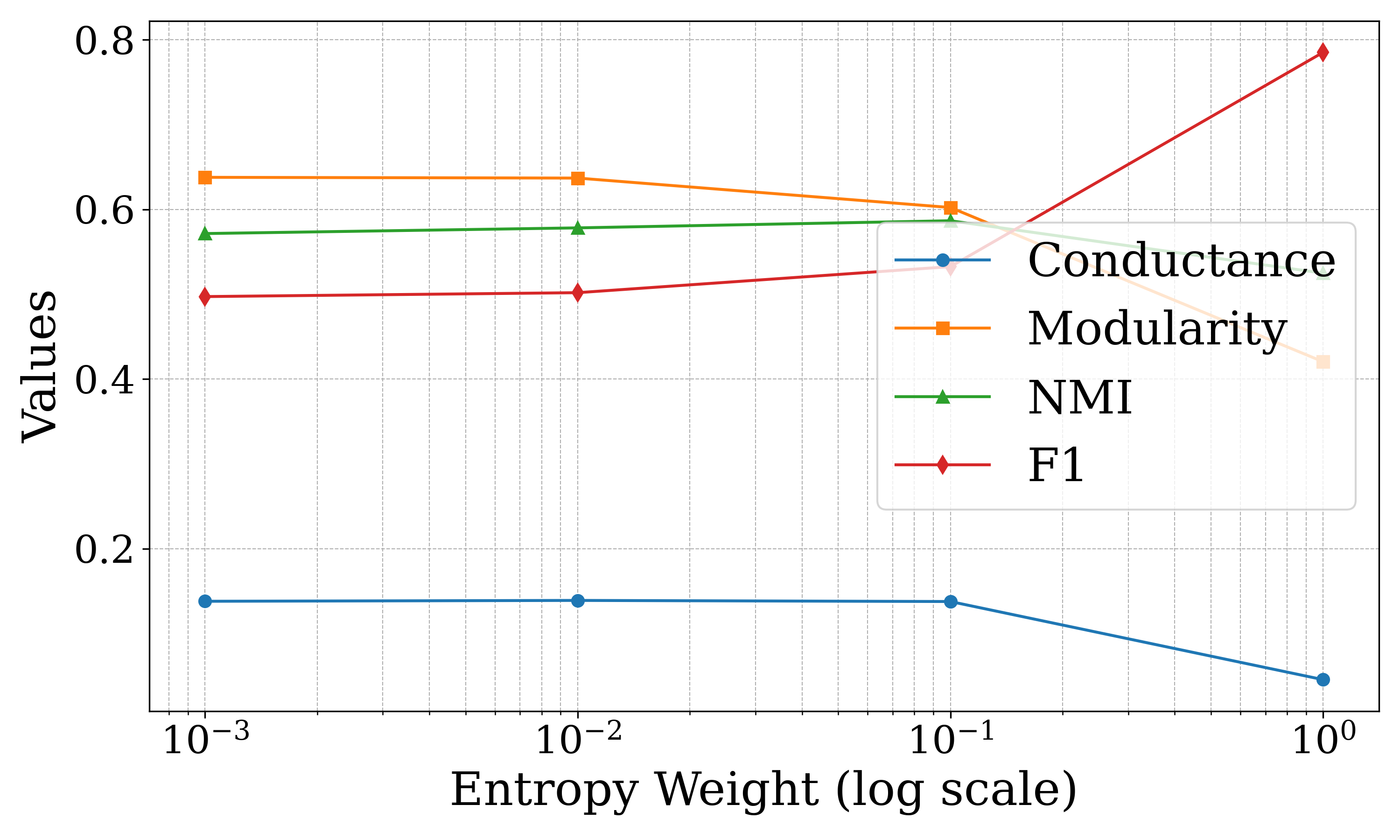}
  \captionsetup{justification=centering}
  \caption{}
  \label{fig:physics_ent}
\end{subfigure}
\caption{Finding the best value of \( W_{\text{entropy}} \) by setting \( W_{\text{dist}} \) and \( W_{\text{var}} \) to zero on the (a) Cora, (b) CiteSeer, (c) PubMed, (d) Coauthor CS, and (e) Coauthor Physics datasets.}
\label{fig:ablation_entropy}
\end{figure}

Figure \ref{fig:ablation_entropy} depicts the ablation study done to find the best value of \( W_{\text{entropy}} \) by setting $W_{\text{dist}}$ and $W_{\text{var}}$ weights to 0. For the Cora, CiteSeer, PubMed, Coauthor CS, and Coauthor Physics datasets, the best \( W_{\text{entropy}} \) values were found to be $0.001$, $0.01$, $0.001$, $0.1$ and $0.1$ respectively.

\subsubsection{Practical Challenges Regarding Manual Tuning}

We acknowledge that adding diversity‐preserving regularizers introduces more hyperparameters, which can complicate practical adoption. However, in our experiments:

\begin{itemize}
\item We found that larger coefficients for distance (R\_D = 1 or 10) were typically beneficial on highly feature‐rich datasets (Coauthor CS, Coauthor Physics).
\item Coefficients of 1 to 0.1 for variance (R\_V) often worked well across most datasets.
\item For the entropy term, 0.1 was a good upper bound on large feature spaces, whereas a smaller coefficient was better on less feature‐rich datasets.
\end{itemize}

From these observations, we hypothesize that stronger diversity coefficients are advantageous for richer feature spaces, whereas smaller coefficients are sufficient (and sometimes necessary) for lower‐dimensional or simpler datasets. As a direction for future work, we suggest exploring automated or adaptive tuning schemes. Such an approach could dynamically adjust the weight of each regularizer during training, further reducing manual overhead and improving reproducibility.

\section{Full Results}
\label{sec:full_results}

The tables below are indicative of the value of each evaluation metric obtained using DMoN/DMoN--DPR pooling for each individual seed for different datasets. The seeds were selected at random using a random number generator. The numbers are all in percentage (\%).

\subsection{Cora}

The following tables list the performance of DiffPool, MinCut Pool, DMoN and different DMoN--DPR configurations across different seeds on the Cora dataset.

\begin{table}[H]
\centering
\small
\caption{Results obtained by using DiffPool on the Cora dataset.}
\begin{tabular}{cccccc}
\toprule
\textbf{Seed} & \textbf{Method} & \textbf{Conductance} & \textbf{Modularity} & \textbf{NMI} & \textbf{F1} \\ \midrule
993 & DiffPool & 15.90 & 63.81 & 42.06 & 51.72 \\
550 & DiffPool & 18.74 & 61.16 & 35.22 & 36.23 \\
243 & DiffPool & 15.84 & 64.02 & 40.57 & 49.64 \\
16 & DiffPool & 17.07 & 57.90 & 37.43 & 46.73 \\
716 & DiffPool & 15.33 & 64.34 & 46.69 & 53.05 \\
383 & DiffPool & 15.54 & 63.93 & 39.65 & 44.05 \\
277 & DiffPool & 12.83 & 67.00 & 47.31 & 55.25 \\
274 & DiffPool & 16.31 & 63.31 & 36.01 & 39.97 \\
188 & DiffPool & 18.26 & 56.66 & 31.67 & 35.33 \\
796 & DiffPool & 14.08 & 65.66 & 44.65 & 53.50 \\
\midrule
\multicolumn{2}{c}{\textbf{Mean $\pm$ Std}} & 15.99 $\pm$ 1.77 & 62.78 $\pm$ 3.28 & 40.13 $\pm$ 5.15 & 46.55 $\pm$ 7.34 \\
\bottomrule
\end{tabular}
\label{tab:diffpool_cora}
\end{table}

\begin{table}[H]
\centering
\small
\caption{Results obtained by using MinCut pooling on the Cora dataset.}
\begin{tabular}{cccccc}
\toprule
\textbf{Seed} & \textbf{Method} & \textbf{Conductance} & \textbf{Modularity} & \textbf{NMI} & \textbf{F1} \\ \midrule
993 & MinCut & 12.73 & 72.59 & 41.66 & 43.90 \\
550 & MinCut & 11.61 & 73.70 & 45.63 & 45.03 \\
243 & MinCut & 13.89 & 71.56 & 31.48 & 33.71 \\
16 & MinCut & 14.08 & 71.37 & 38.06 & 37.21 \\
716 & MinCut & 14.38 & 71.14 & 33.82 & 40.37 \\
383 & MinCut & 15.18 & 70.36 & 29.04 & 28.48 \\
277 & MinCut & 13.09 & 72.42 & 42.81 & 46.15 \\
274 & MinCut & 13.32 & 72.16 & 39.60 & 39.77 \\
188 & MinCut & 13.81 & 71.65 & 40.52 & 44.13 \\
796 & MinCut & 14.44 & 70.91 & 34.75 & 32.71 \\
\midrule
\multicolumn{2}{c}{\textbf{Mean $\pm$ Std}} & 13.65 $\pm$ 1.01 & 71.79 $\pm$ 0.96 & 37.74 $\pm$ 5.31 & 39.15 $\pm$ 5.98 \\
\bottomrule
\end{tabular}
\label{tab:mincut_cora}
\end{table}

\begin{table}[H]
\centering
\small
\caption{Results obtained by using DMoN pooling on the Cora dataset.}
\begin{tabular}{cccccc}
\toprule
\textbf{Seed} & \textbf{Method} & \textbf{Conductance} & \textbf{Modularity} & \textbf{NMI} & \textbf{F1} \\ \midrule
993 & DMoN--DPR (DVE) & 11.35 & 71.32 & 48.90 & 52.56 \\
550 & DMoN--DPR (DVE) & 9.81 & 72.13 & 38.15 & 42.84 \\
243 & DMoN--DPR (DVE) & 10.72 & 73.08 & 40.71 & 40.03 \\
16 & DMoN--DPR (DVE) & 11.06 & 71.48 & 42.84 & 46.41 \\
716 & DMoN--DPR (DVE) & 9.93 & 72.55 & 42.77 & 44.43 \\
383 & DMoN--DPR (DVE) & 11.54 & 73.41 & 44.12 & 51.88 \\
277 & DMoN--DPR (DVE) & 9.66 & 73.52 & 45.61 & 49.43 \\
274 & DMoN--DPR (DVE) & 11.03 & 74.44 & 41.98 & 41.83 \\
188 & DMoN--DPR (DVE) & 9.51 & 74.37 & 51.31 & 56.52 \\
796 & DMoN--DPR (DVE) & 10.65 & 71.36 & 42.78 & 43.37 \\
\midrule
\multicolumn{2}{c}{\textbf{Mean $\pm$ Std}} & 10.53 $\pm$ 0.74 & 72.77 $\pm$ 1.18 & 43.92 $\pm$ 3.85 & 46.93 $\pm$ 5.42 \\
\bottomrule

\end{tabular}
\label{tab:dmon_cora}
\end{table}

\begin{table}[H]
\centering
\small
\caption{Results obtained by using DMoN--DPR pooling, using the best epsilon value of 0.0001, and distance weight of 1 on the Cora dataset.}
\begin{tabular}{cccccc}
\toprule
\textbf{Seed} & \textbf{Method} & \textbf{Conductance} & \textbf{Modularity} & \textbf{NMI} & \textbf{F1} \\ \midrule
993 & DMoN--DPR (D) & 11.31 & 71.37 & 49.62 & 53.10 \\
550 & DMoN--DPR (D) & 9.76 & 72.28 & 40.96 & 45.72 \\
243 & DMoN--DPR (D) & 10.40 & 74.23 & 43.14 & 44.89 \\
16 & DMoN--DPR (D) & 11.90 & 71.00 & 42.22 & 44.14 \\
716 & DMoN--DPR (D) & 9.62 & 71.51 & 43.06 & 46.72 \\
383 & DMoN--DPR (D) & 11.24 & 72.86 & 43.27 & 51.09 \\
277 & DMoN--DPR (D) & 13.64 & 69.81 & 40.51 & 44.53 \\
274 & DMoN--DPR (D) & 11.12 & 73.24 & 43.58 & 45.78 \\
188 & DMoN--DPR (D) & 9.80 & 74.15 & 50.89 & 56.19 \\
796 & DMoN--DPR (D) & 10.70 & 71.52 & 42.57 & 42.95 \\
\midrule
\multicolumn{2}{c}{\textbf{Mean $\pm$ Std}} & 10.95 $\pm$ 1.21 & 72.20 $\pm$ 1.42 & 43.98 $\pm$ 3.46 & 47.51 $\pm$ 4.40 \\
\bottomrule
\end{tabular}
\label{tab:dpr_cora_d}
\end{table}

\begin{table}[H]
\centering
\small
\caption{Results obtained by using DMoN--DPR pooling, using the best variance weight of 1 on the Cora dataset.}
\begin{tabular}{cccccc}
\toprule
\textbf{Seed} & \textbf{Method} & \textbf{Conductance} & \textbf{Modularity} & \textbf{NMI} & \textbf{F1} \\ \midrule
993 & DMoN--DPR (V) & 11.25 & 70.12 & 48.19 & 52.60 \\
550 & DMoN--DPR (V) & 10.95 & 70.66 & 39.21 & 43.64 \\
243 & DMoN--DPR (V) & 11.63 & 71.41 & 40.33 & 39.36 \\
16 & DMoN--DPR (V) & 11.25 & 70.02 & 43.27 & 45.92 \\
716 & DMoN--DPR (V) & 11.22 & 70.58 & 43.18 & 45.43 \\
383 & DMoN--DPR (V) & 13.40 & 70.06 & 42.14 & 51.60 \\
277 & DMoN--DPR (V) & 11.33 & 69.87 & 41.46 & 45.44 \\
274 & DMoN--DPR (V) & 12.43 & 72.57 & 45.58 & 44.85 \\
188 & DMoN--DPR (V) & 12.83 & 70.46 & 46.59 & 51.89 \\
796 & DMoN--DPR (V) & 12.37 & 69.61 & 43.49 & 43.14 \\
\midrule
\multicolumn{2}{c}{\textbf{Mean $\pm$ Std}} & 11.87 $\pm$ 0.83 & 70.54 $\pm$ 0.88 & 43.34 $\pm$ 2.79 & 46.39 $\pm$ 4.32 \\
\bottomrule
\end{tabular}
\label{tab:dpr_cora_v}
\end{table}

\begin{table}[H]
\centering
\small
\caption{Results obtained by using DMoN--DPR pooling, using the best entropy weight of 0.001 on the Cora dataset.}
\begin{tabular}{cccccc}
\toprule
\textbf{Seed} & \textbf{Method} & \textbf{Conductance} & \textbf{Modularity} & \textbf{NMI} & \textbf{F1} \\ \midrule
993 & DMoN--DPR (E) & 11.20 & 71.38 & 48.47 & 52.36 \\
550 & DMoN--DPR (E) & 9.91 & 72.05 & 38.08 & 42.93 \\
243 & DMoN--DPR (E) & 10.57 & 73.23 & 40.85 & 39.97 \\
16 & DMoN--DPR (E) & 11.06 & 71.47 & 42.93 & 46.56 \\
716 & DMoN--DPR (E) & 9.89 & 72.54 & 42.62 & 44.72 \\
383 & DMoN--DPR (E) & 11.42 & 73.51 & 44.30 & 51.74 \\
277 & DMoN--DPR (E) & 9.55 & 73.52 & 45.49 & 49.53 \\
274 & DMoN--DPR (E) & 11.08 & 74.33 & 43.10 & 42.42 \\
188 & DMoN--DPR (E) & 9.61 & 74.26 & 51.24 & 56.67 \\
796 & DMoN--DPR (E) & 10.63 & 71.45 & 42.93 & 43.32 \\
\midrule
\multicolumn{2}{c}{\textbf{Mean $\pm$ Std}} & 10.49 $\pm$ 0.70 & 72.77 $\pm$ 1.15 & 44.00 $\pm$ 3.72 & 47.02 $\pm$ 5.34 \\
\bottomrule
\end{tabular}
\label{tab:dpr_cora_e}
\end{table}

\begin{table}[H]
\centering
\small
\caption{Results obtained by using DMoN--DPR pooling, using the best epsilon value of 0.0001, distance weight of 1, and variance weight of 0.1 on the Cora dataset.}
\begin{tabular}{cccccc}
\toprule
\textbf{Seed} & \textbf{Method} & \textbf{Conductance} & \textbf{Modularity} & \textbf{NMI} & \textbf{F1} \\ \midrule
993 & DMoN--DPR (DV) & 10.93 & 71.05 & 49.87 & 53.66 \\
550 & DMoN--DPR (DV) & 9.51 & 72.33 & 41.04 & 45.65 \\
243 & DMoN--DPR (DV) & 10.84 & 73.28 & 42.18 & 42.58 \\
16 & DMoN--DPR (DV) & 10.86 & 71.23 & 43.08 & 44.77 \\
716 & DMoN--DPR (DV) & 9.61 & 71.39 & 44.18 & 47.59 \\
383 & DMoN--DPR (DV) & 11.80 & 73.16 & 43.66 & 46.75 \\
277 & DMoN--DPR (DV) & 13.74 & 69.61 & 40.23 & 44.66 \\
274 & DMoN--DPR (DV) & 11.33 & 73.29 & 45.41 & 47.47 \\
188 & DMoN--DPR (DV) & 9.61 & 74.03 & 51.32 & 56.96 \\
796 & DMoN--DPR (DV) & 10.72 & 71.30 & 43.02 & 43.54 \\
\midrule
\multicolumn{2}{c}{\textbf{Mean $\pm$ Std}} & 10.89 $\pm$ 1.26 & 72.07 $\pm$ 1.37 & 44.40 $\pm$ 3.60 & 47.36 $\pm$ 4.55 \\
\bottomrule
\end{tabular}
\label{tab:dpr_cora_dv}
\end{table}

\begin{table}[H]
\centering
\small
\caption{Results obtained by using DMoN--DPR pooling, using the best epsilon value of 0.0001, distance weight of 1, and entropy weight of 0.001 on the Cora dataset.}
\begin{tabular}{cccccc}
\toprule
\textbf{Seed} & \textbf{Method} & \textbf{Conductance} & \textbf{Modularity} & \textbf{NMI} & \textbf{F1} \\ \midrule
993 & DMoN--DPR (DE) & 11.16 & 71.38 & 49.75 & 53.29 \\
550 & DMoN--DPR (DE) & 9.68 & 72.33 & 40.85 & 45.67 \\
243 & DMoN--DPR (DE) & 10.08 & 74.45 & 43.94 & 46.11 \\
16 & DMoN--DPR (DE) & 11.29 & 71.57 & 43.55 & 43.97 \\
716 & DMoN--DPR (DE) & 9.61 & 71.58 & 44.19 & 47.71 \\
383 & DMoN--DPR (DE) & 11.73 & 73.06 & 42.66 & 48.52 \\
277 & DMoN--DPR (DE) & 13.58 & 69.91 & 40.43 & 44.48 \\
274 & DMoN--DPR (DE) & 11.01 & 73.29 & 43.61 & 45.77 \\
188 & DMoN--DPR (DE) & 9.81 & 74.04 & 51.34 & 56.67 \\
796 & DMoN--DPR (DE) & 10.82 & 71.36 & 42.67 & 43.10 \\
\midrule
\multicolumn{2}{c}{\textbf{Mean $\pm$ Std}} & 10.88 $\pm$ 1.21 & 72.30 $\pm$ 1.40 & 44.30 $\pm$ 3.54 & 47.53 $\pm$ 4.32 \\
\bottomrule
\end{tabular}
\label{tab:dpr_cora_de}
\end{table}

\begin{table}[H]
\centering
\small
\caption{Results obtained by using DMoN--DPR pooling, using the best variance weight of 1, and entropy weight of 0.001 on the Cora dataset.}
\begin{tabular}{cccccc}
\toprule
\textbf{Seed} & \textbf{Method} & \textbf{Conductance} & \textbf{Modularity} & \textbf{NMI} & \textbf{F1} \\ \midrule
993 & DMoN--DPR (VE) & 11.16 & 70.08 & 48.03 & 52.85 \\
550 & DMoN--DPR (VE) & 10.86 & 70.77 & 39.39 & 43.76 \\
243 & DMoN--DPR (VE) & 11.78 & 71.27 & 40.17 & 39.23 \\
16 & DMoN--DPR (VE) & 11.29 & 69.93 & 43.33 & 45.86 \\
716 & DMoN--DPR (VE) & 11.10 & 70.70 & 43.42 & 45.66 \\
383 & DMoN--DPR (VE) & 13.17 & 70.14 & 41.54 & 51.22 \\
277 & DMoN--DPR (VE) & 11.46 & 69.60 & 41.28 & 45.35 \\
274 & DMoN--DPR (VE) & 12.43 & 72.56 & 45.66 & 44.94 \\
188 & DMoN--DPR (VE) & 12.90 & 70.39 & 46.46 & 51.76 \\
796 & DMoN--DPR (VE) & 12.41 & 69.61 & 43.10 & 42.98 \\
\midrule
\multicolumn{2}{c}{\textbf{Mean $\pm$ Std}} & 11.86 $\pm$ 0.82 & 70.50 $\pm$ 0.89 & 43.24 $\pm$ 2.79 & 46.36 $\pm$ 4.32 \\
\bottomrule
\end{tabular}
\label{tab:dpr_cora_ve}
\end{table}

\begin{table}[H]
\centering
\small
\caption{Results obtained by using DMoN--DPR pooling, using the best epsilon value of 0.0001, variance weight of 0.1, and entropy weight of 0.001 on the Cora dataset.}
\begin{tabular}{cccccc}
\toprule
\textbf{Seed} & \textbf{Method} & \textbf{Conductance} & \textbf{Modularity} & \textbf{NMI} & \textbf{F1} \\ \midrule
993 & DMoN--DPR (DVE) & 10.65 & 71.20 & 49.95 & 53.90 \\
550 & DMoN--DPR (DVE) & 9.51 & 72.33 & 41.04 & 45.64 \\
243 & DMoN--DPR (DVE) & 10.86 & 73.34 & 41.92 & 42.43 \\
16 & DMoN--DPR (DVE) & 10.78 & 71.29 & 42.83 & 44.73 \\
716 & DMoN--DPR (DVE) & 9.74 & 71.30 & 43.98 & 47.36 \\
383 & DMoN--DPR (DVE) & 12.58 & 72.50 & 42.83 & 45.78 \\
277 & DMoN--DPR (DVE) & 13.81 & 69.51 & 40.61 & 45.21 \\
274 & DMoN--DPR (DVE) & 10.97 & 73.71 & 46.42 & 47.92 \\
188 & DMoN--DPR (DVE) & 9.64 & 73.99 & 51.33 & 56.97 \\
796 & DMoN--DPR (DVE) & 10.80 & 71.35 & 42.74 & 43.29 \\
\midrule
\multicolumn{2}{c}{\textbf{Mean $\pm$ Std}} & 10.93 $\pm$ 1.34 & 72.05 $\pm$ 1.39 & 44.37 $\pm$ 3.69 & 47.32 $\pm$ 4.63 \\
\bottomrule
\end{tabular}
\label{tab:dpr_cora_dve}
\end{table}

\subsection{CiteSeer}

The following tables list the performance of DiffPool, MinCut pooling, DMoN and different DMoN--DPR configurations across different seeds on the CiteSeer dataset.

\begin{table}[H]
\centering
\small
\caption{Results obtained by using DiffPool pooling on the CiteSeer dataset.}
\begin{tabular}{cccccc}
\toprule
\textbf{Seed} & \textbf{Method} & \textbf{Conductance} & \textbf{Modularity} & \textbf{NMI} & \textbf{F1} \\ \midrule
993 & DiffPool & 8.24 & 64.76 & 33.99 & 48.98 \\
550 & DiffPool & 7.27 & 70.84 & 38.27 & 51.01 \\
243 & DiffPool & 6.59 & 66.78 & 34.97 & 51.19 \\
16 & DiffPool & 10.43 & 68.61 & 34.15 & 47.16 \\
716 & DiffPool & 7.01 & 70.87 & 33.12 & 46.30 \\
383 & DiffPool & 9.49 & 56.09 & 28.78 & 45.91 \\
277 & DiffPool & 9.58 & 64.40 & 29.61 & 41.80 \\
274 & DiffPool & 6.50 & 71.68 & 32.23 & 44.80 \\
188 & DiffPool & 7.18 & 66.15 & 32.83 & 48.86 \\
796 & DiffPool & 6.94 & 66.72 & 36.02 & 52.33 \\
\midrule
\multicolumn{2}{c}{\textbf{Mean $\pm$ Std}} & 7.92 $\pm$ 1.42 & 66.69 $\pm$ 4.53 & 33.40 $\pm$ 2.82 & 47.83 $\pm$ 3.27 \\
\bottomrule
\end{tabular}
\label{tab:diffpool_citeseer}
\end{table}

\begin{table}[H]
\centering
\small
\caption{Results obtained by using MinCut pooling on the CiteSeer dataset.}
\begin{tabular}{cccccc}
\toprule
\textbf{Seed} & \textbf{Method} & \textbf{Conductance} & \textbf{Modularity} & \textbf{NMI} & \textbf{F1} \\ \midrule
993 & MinCut & 5.45 & 76.59 & 33.17 & 45.03 \\
550 & MinCut & 5.34 & 75.36 & 24.92 & 35.92 \\
243 & MinCut & 5.01 & 76.67 & 30.26 & 42.82 \\
16 & MinCut & 7.21 & 73.20 & 18.09 & 26.08 \\
716 & MinCut & 7.01 & 74.82 & 22.11 & 30.09 \\
383 & MinCut & 6.04 & 75.25 & 23.28 & 34.72 \\
277 & MinCut & 7.05 & 73.86 & 17.40 & 25.14 \\
274 & MinCut & 5.32 & 76.16 & 25.47 & 38.90 \\
188 & MinCut & 6.44 & 75.23 & 32.94 & 42.54 \\
796 & MinCut & 7.05 & 74.93 & 24.43 & 31.51 \\
\midrule
\multicolumn{2}{c}{\textbf{Mean $\pm$ Std}} & 6.19 $\pm$ 0.86 & 75.21 $\pm$ 1.11 & 25.21 $\pm$ 5.52 & 35.28 $\pm$ 7.05 \\
\bottomrule
\end{tabular}
\label{tab:mincut_citeseer}
\end{table}

\begin{table}[H]
\centering
\small
\caption{Results obtained by using DMoN pooling on the CiteSeer dataset.}
\begin{tabular}{cccccc}
\toprule
\textbf{Seed} & \textbf{Method} & \textbf{Conductance} & \textbf{Modularity} & \textbf{NMI} & \textbf{F1} \\ \midrule
993 & DMoN & 6.81 & 74.92 & 20.78 & 26.09 \\
550 & DMoN & 4.11 & 75.53 & 26.91 & 42.64 \\
243 & DMoN & 4.28 & 75.57 & 32.12 & 45.74 \\
16 & DMoN & 4.64 & 77.22 & 33.11 & 44.56 \\
716 & DMoN & 4.53 & 76.24 & 36.95 & 50.12 \\
383 & DMoN & 4.24 & 75.55 & 33.25 & 47.00 \\
277 & DMoN & 4.26 & 75.63 & 29.25 & 42.33 \\
274 & DMoN & 4.48 & 75.77 & 30.83 & 45.65 \\
188 & DMoN & 6.17 & 72.74 & 24.52 & 36.01 \\
796 & DMoN & 5.12 & 75.34 & 28.98 & 44.42 \\
\midrule
\multicolumn{2}{c}{\textbf{Mean $\pm$ Std}} & 4.86 $\pm$ 0.91 & 75.45 $\pm$ 1.13 & 29.67 $\pm$ 4.70 & 42.46 $\pm$ 6.82 \\
\bottomrule
\end{tabular}
\label{tab:dmon_citeseer}
\end{table}

\begin{table}[H]
\centering
\small
\caption{Results obtained by using DMoN--DPR pooling using the best epsilon value of 0.0001, and distance weight of 1 on the CiteSeer dataset.}
\begin{tabular}{cccccc}
\toprule
\textbf{Seed} & \textbf{Method} & \textbf{Conductance} & \textbf{Modularity} & \textbf{NMI} & \textbf{F1} \\ \midrule
993 & DMoN--DPR (D) & 7.18 & 74.27 & 23.90 & 27.73 \\
550 & DMoN--DPR (D) & 4.22 & 75.80 & 26.23 & 41.98 \\
243 & DMoN--DPR (D) & 3.82 & 76.02 & 35.91 & 50.13 \\
16 & DMoN--DPR (D) & 4.83 & 76.31 & 32.78 & 44.47 \\
716 & DMoN--DPR (D) & 4.57 & 76.13 & 36.99 & 50.14 \\
383 & DMoN--DPR (D) & 4.81 & 75.97 & 33.02 & 45.96 \\
277 & DMoN--DPR (D) & 3.87 & 75.75 & 29.37 & 42.34 \\
274 & DMoN--DPR (D) & 4.15 & 76.06 & 30.88 & 45.48 \\
188 & DMoN--DPR (D) & 6.37 & 73.98 & 25.00 & 36.47 \\
796 & DMoN--DPR (D) & 5.71 & 75.26 & 26.83 & 39.93 \\
\midrule
\multicolumn{2}{c}{\textbf{Mean $\pm$ Std}} & 4.95 $\pm$ 1.12 & 75.55 $\pm$ 0.81 & 30.09 $\pm$ 4.57 & 42.46 $\pm$ 6.69 \\
\bottomrule
\end{tabular}
\label{tab:dpr_citeseer_d}
\end{table}

\begin{table}[H]
\centering
\small
\caption{Results obtained by using DMoN--DPR pooling using the best variance weight of 0.1 on the CiteSeer dataset.}
\label{tab:dpr_citeseer_v}
\begin{tabular}{cccccc}
\toprule
\textbf{Seed} & \textbf{Method} & \textbf{Conductance} & \textbf{Modularity} & \textbf{NMI} & \textbf{F1} \\ \midrule
993 & DMoN--DPR (V) & 6.83 & 74.31 & 21.24 & 28.03 \\
550 & DMoN--DPR (V) & 4.04 & 75.60 & 26.97 & 42.75 \\
243 & DMoN--DPR (V) & 4.28 & 75.41 & 32.23 & 45.77 \\
16 & DMoN--DPR (V) & 5.14 & 76.33 & 34.01 & 46.10 \\
716 & DMoN--DPR (V) & 4.57 & 76.13 & 36.89 & 50.11 \\
383 & DMoN--DPR (V) & 4.15 & 75.38 & 33.48 & 47.59 \\
277 & DMoN--DPR (V) & 4.35 & 75.42 & 28.95 & 42.12 \\
274 & DMoN--DPR (V) & 4.59 & 75.13 & 30.92 & 46.21 \\
188 & DMoN--DPR (V) & 6.39 & 73.03 & 24.67 & 35.88 \\
796 & DMoN--DPR (V) & 5.21 & 74.93 & 28.27 & 43.93 \\
\midrule
\multicolumn{2}{c}{\textbf{Mean $\pm$ Std}} & 4.96 $\pm$ 0.96 & 75.17 $\pm$ 0.94 & 29.76 $\pm$ 4.71 & 42.85 $\pm$ 6.46 \\
\bottomrule
\end{tabular}
\end{table}

\begin{table}[H]
\centering
\small
\caption{Results obtained by using DMoN--DPR pooling using the best entropy weight of 0.01 on the CiteSeer dataset.}
\begin{tabular}{cccccc}
\toprule
\textbf{Seed} & \textbf{Method} & \textbf{Conductance} & \textbf{Modularity} & \textbf{NMI} & \textbf{F1} \\ \midrule
993 & DMoN--DPR (E) & 7.27 & 73.95 & 21.36 & 28.28 \\
550 & DMoN--DPR (E) & 4.15 & 75.67 & 26.79 & 42.59 \\
243 & DMoN--DPR (E) & 4.31 & 75.39 & 32.27 & 45.75 \\
16 & DMoN--DPR (E) & 4.92 & 76.51 & 34.13 & 46.04 \\
716 & DMoN--DPR (E) & 4.48 & 76.00 & 37.03 & 50.46 \\
383 & DMoN--DPR (E) & 4.46 & 75.36 & 32.82 & 46.38 \\
277 & DMoN--DPR (E) & 4.20 & 75.58 & 29.78 & 43.08 \\
274 & DMoN--DPR (E) & 4.26 & 75.81 & 31.63 & 46.59 \\
188 & DMoN--DPR (E) & 6.22 & 73.30 & 25.10 & 36.09 \\
796 & DMoN--DPR (E) & 5.07 & 74.98 & 28.59 & 44.38 \\
\midrule
\multicolumn{2}{c}{\textbf{Mean $\pm$ Std}} & 4.93 $\pm$ 1.03 & 75.25 $\pm$ 0.96 & 29.95 $\pm$ 4.63 & 42.96 $\pm$ 6.35 \\
\bottomrule
\end{tabular}
\label{tab:dpr_citeseer_e}
\end{table}

\begin{table}[H]
\centering
\small
\caption{Results obtained by using DMoN--DPR pooling using the best epsilon value of 0.0001, distance weight of 1, and variance weight of 0.1 on the CiteSeer dataset.}
\begin{tabular}{cccccc}
\toprule
\textbf{Seed} & \textbf{Method} & \textbf{Conductance} & \textbf{Modularity} & \textbf{NMI} & \textbf{F1} \\ \midrule
993 & DMoN--DPR & 7.54 & 73.39 & 23.91 & 29.07 \\
550 & DMoN--DPR & 4.35 & 75.62 & 26.61 & 42.29 \\
243 & DMoN--DPR & 3.84 & 75.78 & 36.17 & 50.51 \\
16 & DMoN--DPR & 4.86 & 76.09 & 33.03 & 44.97 \\
716 & DMoN--DPR & 4.55 & 76.03 & 38.13 & 51.55 \\
383 & DMoN--DPR & 4.90 & 75.53 & 33.12 & 46.09 \\
277 & DMoN--DPR & 4.06 & 75.33 & 29.89 & 42.79 \\
274 & DMoN--DPR & 4.28 & 75.82 & 31.51 & 46.34 \\
188 & DMoN--DPR & 6.50 & 73.71 & 25.38 & 36.63 \\
796 & DMoN--DPR & 6.04 & 74.53 & 27.25 & 42.22 \\
\midrule
\multicolumn{2}{c}{\textbf{Mean $\pm$ Std}} & 5.09 $\pm$ 1.21 & 75.18 $\pm$ 0.97 & 30.50 $\pm$ 4.72 & 43.25 $\pm$ 6.58 \\
\bottomrule
\end{tabular}
\label{tab:dpr_citeseer_dv}
\end{table}

\begin{table}[H]
\centering
\small
\caption{Results obtained by using DMoN--DPR pooling using the best epsilon value of 0.0001, distance weight of 1, and entropy weight of 0.01 on the CiteSeer dataset.}
\begin{tabular}{cccccc}
\toprule
\textbf{Seed} & \textbf{Method} & \textbf{Conductance} & \textbf{Modularity} & \textbf{NMI} & \textbf{F1} \\ \midrule
993 & DMoN--DPR (DE) & 7.45 & 73.34 & 24.03 & 29.50 \\
550 & DMoN--DPR (DE) & 4.37 & 75.63 & 26.92 & 42.46 \\
243 & DMoN--DPR (DE) & 3.89 & 75.81 & 36.33 & 50.53 \\
16 & DMoN--DPR (DE) & 5.07 & 75.80 & 32.09 & 43.42 \\
716 & DMoN--DPR (DE) & 4.33 & 75.97 & 38.51 & 52.05 \\
383 & DMoN--DPR (DE) & 4.83 & 75.58 & 32.59 & 45.72 \\
277 & DMoN--DPR (DE) & 4.02 & 75.47 & 30.22 & 43.10 \\
274 & DMoN--DPR (DE) & 4.35 & 75.80 & 31.50 & 46.10 \\
188 & DMoN--DPR (DE) & 6.35 & 73.93 & 25.29 & 36.35 \\
796 & DMoN--DPR (DE) & 6.06 & 74.49 & 27.23 & 42.08 \\
\midrule
\multicolumn{2}{c}{\textbf{Mean $\pm$ Std}} & 5.07 $\pm$ 1.17 & 75.18 $\pm$ 0.92 & 30.47 $\pm$ 4.70 & 43.13 $\pm$ 6.53 \\
\bottomrule
\end{tabular}
\label{tab:dpr_citeseer_de}
\end{table}

\begin{table}[H]
\centering
\small
\caption{Results obtained by using DMoN--DPR pooling using the best variance weight of 0.1, and best entropy weight of 0.01 on the CiteSeer dataset.}
\begin{tabular}{cccccc}
\toprule
\textbf{Seed} & \textbf{Method} & \textbf{Conductance} & \textbf{Modularity} & \textbf{NMI} & \textbf{F1} \\ \midrule
993 & DMoN--DPR (VE) & 7.51 & 73.51 & 21.85 & 28.66 \\
550 & DMoN--DPR (VE) & 4.35 & 75.43 & 26.50 & 42.19 \\
243 & DMoN--DPR (VE) & 4.48 & 75.15 & 32.58 & 46.11 \\
16 & DMoN--DPR (VE) & 4.79 & 76.51 & 33.09 & 45.03 \\
716 & DMoN--DPR (VE) & 4.50 & 75.95 & 37.46 & 51.15 \\
383 & DMoN--DPR (VE) & 4.61 & 74.94 & 33.20 & 46.98 \\
277 & DMoN--DPR (VE) & 4.24 & 75.33 & 30.08 & 43.40 \\
274 & DMoN--DPR (VE) & 4.37 & 75.69 & 31.72 & 46.90 \\
188 & DMoN--DPR (VE) & 6.39 & 72.92 & 25.40 & 36.39 \\
796 & DMoN--DPR (VE) & 5.23 & 74.71 & 29.84 & 45.84 \\
\midrule
\multicolumn{2}{c}{\textbf{Mean $\pm$ Std}} & 5.05 $\pm$ 1.07 & 75.01 $\pm$ 1.09 & 30.17 $\pm$ 4.53 & 43.27 $\pm$ 6.40 \\
\bottomrule
\end{tabular}
\label{tab:dpr_citeseer_ve}
\end{table}

\begin{table}[H]
\centering
\small
\caption{Results obtained by using DMoN--DPR pooling using the best epsilon value of 0.0001, distance weight of 1, variance weight of 0.1, and entropy weight of 0.01 on the CiteSeer dataset.}
\begin{tabular}{cccccc}
\toprule
\textbf{Seed} & \textbf{Method} & \textbf{Conductance} & \textbf{Modularity} & \textbf{NMI} & \textbf{F1} \\ \midrule
993 & DMoN--DPR & 7.45 & 73.17 & 23.84 & 29.48 \\
550 & DMoN--DPR & 4.15 & 75.58 & 26.66 & 42.48 \\
243 & DMoN--DPR & 3.91 & 75.68 & 36.30 & 50.67 \\
16 & DMoN--DPR & 5.23 & 75.44 & 31.58 & 42.92 \\
716 & DMoN--DPR & 4.59 & 75.96 & 37.14 & 50.08 \\
383 & DMoN--DPR & 4.75 & 75.43 & 32.95 & 45.93 \\
277 & DMoN--DPR & 4.44 & 75.10 & 29.34 & 41.94 \\
274 & DMoN--DPR & 4.35 & 75.91 & 31.30 & 46.10 \\
188 & DMoN--DPR & 6.52 & 73.57 & 26.09 & 37.49 \\
796 & DMoN--DPR & 6.24 & 74.21 & 27.32 & 42.91 \\
\midrule
\multicolumn{2}{c}{\textbf{Mean $\pm$ Std}} & 5.16 $\pm$ 1.18 & 75.01 $\pm$ 1.00 & 30.25 $\pm$ 4.41 & 43.00 $\pm$ 6.16 \\
\bottomrule
\end{tabular}
\label{tab:dpr_citeseer_dve}
\end{table}

\subsection{PubMed}

The following tables list the performance of MinCut pooling, DMoN and different DMoN--DPR configurations across different seeds on the PubMed dataset. 

\begin{table}[H]
\centering
\small
\caption{Results obtained by using MinCut pooling on the PubMed dataset.}
\begin{tabular}{cccccc}
\toprule
\textbf{Seed} & \textbf{Method} & \textbf{Conductance} & \textbf{Modularity} & \textbf{NMI} & \textbf{F1} \\ \midrule
993 & MinCut & 12.82 & 53.38 & 23.17 & 41.29 \\
550 & MinCut & 8.41 & 56.48 & 21.15 & 44.79 \\
243 & MinCut & 12.52 & 53.77 & 23.64 & 40.72 \\
16 & MinCut & 12.60 & 53.65 & 23.65 & 39.98 \\
716 & MinCut & 10.43 & 55.36 & 21.44 & 37.44 \\
383 & MinCut & 12.62 & 53.67 & 23.85 & 41.18 \\
277 & MinCut & 8.32 & 56.56 & 20.47 & 44.08 \\
274 & MinCut & 8.33 & 56.46 & 20.25 & 44.46 \\
188 & MinCut & 12.52 & 53.76 & 23.34 & 38.89 \\
796 & MinCut & 12.79 & 53.53 & 23.84 & 40.30 \\
\midrule
\multicolumn{2}{c}{\textbf{Mean $\pm$ Std}} & 11.14 $\pm$ 2.04 & 54.66 $\pm$ 1.38 & 22.48 $\pm$ 1.47 & 41.31 $\pm$ 2.44 \\
\bottomrule
\end{tabular}
\label{tab:mincut_pubmed}
\end{table}

\begin{table}[H]
\centering
\small
\caption{Results obtained by using DMoN pooling on the PubMed dataset.}
\begin{tabular}{cccccc}
\toprule
\textbf{Seed} & \textbf{Method} & \textbf{Conductance} & \textbf{Modularity} & \textbf{NMI} & \textbf{F1} \\ \midrule
993 & DMoN & 10.94 & 55.54 & 22.49 & 40.20 \\
550 & DMoN & 7.64 & 58.82 & 26.92 & 49.53 \\
243 & DMoN & 7.93 & 57.62 & 21.77 & 44.56 \\
16 & DMoN & 7.96 & 57.63 & 21.74 & 44.37 \\
716 & DMoN & 8.05 & 57.34 & 21.70 & 44.97 \\
383 & DMoN & 9.79 & 56.12 & 22.79 & 37.01 \\
277 & DMoN & 7.80 & 57.63 & 21.69 & 44.89 \\
274 & DMoN & 7.90 & 57.45 & 21.60 & 44.80 \\
188 & DMoN & 10.09 & 55.60 & 21.39 & 37.09 \\
796 & DMoN & 7.96 & 57.52 & 21.76 & 44.85 \\
\midrule
\multicolumn{2}{c}{\textbf{Mean $\pm$ Std}} & 8.61 $\pm$ 1.19 & 57.13 $\pm$ 1.04 & 22.38 $\pm$ 1.65 & 43.23 $\pm$ 3.93 \\
\bottomrule
\end{tabular}
\label{tab:dmon_pubmed}
\end{table}

\begin{table}[H]
\centering
\small
\caption{Results obtained by using DMoN--DPR pooling using the best epsilon value of 0.001 and distance weight of 1 on the PubMed dataset.}
\begin{tabular}{cccccc}
\toprule
\textbf{Seed} & \textbf{Method} & \textbf{Conductance} & \textbf{Modularity} & \textbf{NMI} & \textbf{F1} \\ \midrule
993 & DMoN--DPR (D) & 10.82 & 55.67 & 22.35 & 40.01 \\
550 & DMoN--DPR (D) & 7.65 & 58.81 & 26.89 & 49.57 \\
243 & DMoN--DPR (D) & 7.95 & 57.60 & 21.79 & 44.56 \\
16 & DMoN--DPR (D) & 7.99 & 57.59 & 21.73 & 44.35 \\
716 & DMoN--DPR (D) & 8.05 & 57.36 & 21.71 & 44.95 \\
383 & DMoN--DPR (D) & 9.78 & 56.13 & 22.83 & 37.08 \\
277 & DMoN--DPR (D) & 7.81 & 57.62 & 21.67 & 44.88 \\
274 & DMoN--DPR (D) & 7.82 & 57.53 & 21.67 & 44.91 \\
188 & DMoN--DPR (D) & 10.14 & 55.54 & 21.34 & 37.09 \\
796 & DMoN--DPR (D) & 7.97 & 57.51 & 21.82 & 44.96 \\
\midrule
\multicolumn{2}{c}{\textbf{Mean $\pm$ Std}} & 8.60 $\pm$ 1.17 & 57.14 $\pm$ 1.03 & 22.38 $\pm$ 1.64 & 43.24 $\pm$ 3.95 \\
\bottomrule
\end{tabular}
\label{tab:dpr_pubmed_d}
\end{table}

\begin{table}[H]
\centering
\small
\caption{Results obtained by using DMoN--DPR pooling using the best variance weight of 0.001 on the PubMed dataset.}
\begin{tabular}{cccccc}
\toprule
\textbf{Seed} & \textbf{Method} & \textbf{Conductance} & \textbf{Modularity} & \textbf{NMI} & \textbf{F1} \\ \midrule
993 & DMoN--DPR (V) & 10.94 & 55.54 & 22.54 & 40.25 \\
550 & DMoN--DPR (V) & 7.64 & 58.82 & 26.91 & 49.52 \\
243 & DMoN--DPR (V) & 7.94 & 57.61 & 21.76 & 44.54 \\
16 & DMoN--DPR (V) & 7.96 & 57.63 & 21.73 & 44.37 \\
716 & DMoN--DPR (V) & 8.05 & 57.34 & 21.70 & 44.97 \\
383 & DMoN--DPR (V) & 9.78 & 56.13 & 22.85 & 37.03 \\
277 & DMoN--DPR (V) & 7.80 & 57.63 & 21.69 & 44.89 \\
274 & DMoN--DPR (V) & 7.91 & 57.45 & 21.60 & 44.79 \\
188 & DMoN--DPR (V) & 10.09 & 55.60 & 21.38 & 37.07 \\
796 & DMoN--DPR (V) & 7.98 & 57.50 & 21.76 & 44.86 \\
\midrule
\multicolumn{2}{c}{\textbf{Mean $\pm$ Std}} & 8.61 $\pm$ 1.19 & 57.12 $\pm$ 1.04 & 22.39 $\pm$ 1.65 & 43.23 $\pm$ 3.93 \\
\bottomrule
\end{tabular}
\label{tab:dpr_pubmed_v}
\end{table}

\begin{table}[H]
\centering
\small
\caption{Results obtained by using DMoN--DPR pooling using the best entropy weight of 0.001 on the PubMed dataset.}
\begin{tabular}{cccccc}
\toprule
\textbf{Seed} & \textbf{Method} & \textbf{Conductance} & \textbf{Modularity} & \textbf{NMI} & \textbf{F1} \\ \midrule
993 & DMoN--DPR (E) & 10.96 & 55.52 & 22.51 & 40.19 \\
550 & DMoN--DPR (E) & 7.65 & 58.81 & 26.90 & 49.50 \\
243 & DMoN--DPR (E) & 7.94 & 57.61 & 21.76 & 44.55 \\
16 & DMoN--DPR (E) & 7.97 & 57.62 & 21.74 & 44.37 \\
716 & DMoN--DPR (E) & 8.05 & 57.33 & 21.63 & 44.91 \\
383 & DMoN--DPR (E) & 9.82 & 56.08 & 22.79 & 36.95 \\
277 & DMoN--DPR (E) & 7.80 & 57.63 & 21.69 & 44.89 \\
274 & DMoN--DPR (E) & 7.91 & 57.45 & 21.60 & 44.80 \\
188 & DMoN--DPR (E) & 10.08 & 55.56 & 21.26 & 37.16 \\
796 & DMoN--DPR (E) & 7.98 & 57.51 & 21.75 & 44.85 \\
\midrule
\multicolumn{2}{c}{\textbf{Mean $\pm$ Std}} & 8.62 $\pm$ 1.19 & 57.11 $\pm$ 1.05 & 22.36 $\pm$ 1.66 & 43.22 $\pm$ 3.92 \\
\bottomrule
\end{tabular}
\label{tab:dpr_pubmed_e}
\end{table}

\begin{table}[H]
\centering
\small
\caption{Results obtained by using DMoN--DPR pooling using the best epsilon value of 0.001, distance weight of 1, and variance weight of 0.001 on the PubMed dataset.}
\begin{tabular}{cccccc}
\toprule
\textbf{Seed} & \textbf{Method} & \textbf{Conductance} & \textbf{Modularity} & \textbf{NMI} & \textbf{F1} \\ \midrule
993 & DMoN--DPR (DV) & 10.84 & 55.65 & 22.38 & 40.02 \\
550 & DMoN--DPR (DV) & 7.65 & 58.81 & 26.87 & 49.55 \\
243 & DMoN--DPR (DV) & 7.95 & 57.60 & 21.79 & 44.56 \\
16 & DMoN--DPR (DV) & 7.99 & 57.59 & 21.74 & 44.37 \\
716 & DMoN--DPR (DV) & 8.06 & 57.35 & 21.72 & 44.96 \\
383 & DMoN--DPR (DV) & 9.76 & 56.15 & 22.84 & 37.08 \\
277 & DMoN--DPR (DV) & 7.81 & 57.62 & 21.67 & 44.88 \\
274 & DMoN--DPR (DV) & 7.82 & 57.53 & 21.67 & 44.91 \\
188 & DMoN--DPR (DV) & 10.12 & 55.55 & 21.30 & 37.10 \\
796 & DMoN--DPR (DV) & 7.97 & 57.50 & 21.79 & 44.92 \\
\midrule
\multicolumn{2}{c}{\textbf{Mean $\pm$ Std}} & 8.60 $\pm$ 1.17 & 57.13 $\pm$ 1.03 & 22.38 $\pm$ 1.64 & 43.23 $\pm$ 3.95 \\
\bottomrule
\end{tabular}
\label{tab:dpr_pubmed_dv}
\end{table}

\begin{table}[H]
\centering
\small
\caption{Results obtained by using DMoN--DPR pooling using the best epsilon value of 0.001, distance weight of 1, and entropy weight of 0.001 on the PubMed dataset.}
\begin{tabular}{cccccc}
\toprule
\textbf{Seed} & \textbf{Method} & \textbf{Conductance} & \textbf{Modularity} & \textbf{NMI} & \textbf{F1} \\ \midrule
993 & DMoN--DPR (DE) & 10.83 & 55.65 & 22.42 & 40.03 \\
550 & DMoN--DPR (DE) & 7.65 & 58.80 & 26.87 & 49.55 \\
243 & DMoN--DPR (DE) & 7.95 & 57.60 & 21.79 & 44.57 \\
16 & DMoN--DPR (DE) & 8.02 & 57.57 & 21.69 & 44.34 \\
716 & DMoN--DPR (DE) & 8.03 & 57.37 & 21.77 & 45.03 \\
383 & DMoN--DPR (DE) & 9.81 & 56.10 & 22.81 & 37.07 \\
277 & DMoN--DPR (DE) & 7.81 & 57.62 & 21.67 & 44.87 \\
274 & DMoN--DPR (DE) & 7.82 & 57.52 & 21.68 & 44.92 \\
188 & DMoN--DPR (DE) & 10.10 & 55.50 & 21.32 & 37.24 \\
796 & DMoN--DPR (DE) & 7.98 & 57.50 & 21.79 & 44.92 \\
\midrule
\multicolumn{2}{c}{\textbf{Mean $\pm$ Std}} & 8.60 $\pm$ 1.17 & 57.12 $\pm$ 1.04 & 22.38 $\pm$ 1.63 & 43.25 $\pm$ 3.93 \\
\bottomrule
\end{tabular}
\label{tab:dpr_pubmed_de}
\end{table}

\begin{table}[H]
\centering
\small
\caption{Results obtained by using DMoN--DPR pooling using the best variance weight of 0.001, and entropy weight of 0.001 on the PubMed dataset.}
\begin{tabular}{cccccc}
\toprule
\textbf{Seed} & \textbf{Method} & \textbf{Conductance} & \textbf{Modularity} & \textbf{NMI} & \textbf{F1} \\ \midrule
993 & DMoN--DPR (VE) & 10.85 & 55.64 & 22.48 & 40.10 \\
550 & DMoN--DPR (VE) & 7.65 & 58.82 & 26.90 & 49.50 \\
243 & DMoN--DPR (VE) & 7.94 & 57.61 & 21.76 & 44.54 \\
16 & DMoN--DPR (VE) & 7.97 & 57.62 & 21.74 & 44.37 \\
716 & DMoN--DPR (VE) & 8.06 & 57.33 & 21.63 & 44.91 \\
383 & DMoN--DPR (VE) & 9.80 & 56.09 & 22.79 & 36.98 \\
277 & DMoN--DPR (VE) & 7.80 & 57.63 & 21.68 & 44.88 \\
274 & DMoN--DPR (VE) & 7.92 & 57.44 & 21.60 & 44.79 \\
188 & DMoN--DPR (VE) & 10.01 & 55.65 & 21.43 & 37.15 \\
796 & DMoN--DPR (VE) & 7.97 & 57.50 & 21.73 & 44.84 \\
\midrule
\multicolumn{2}{c}{\textbf{Mean $\pm$ Std}} & 8.60 $\pm$ 1.16 & 57.13 $\pm$ 1.02 & 22.37 $\pm$ 1.65 & 43.21 $\pm$ 3.93 \\
\bottomrule
\end{tabular}
\label{tab:dpr_pubmed_ve}
\end{table}

\begin{table}[H]
\centering
\small
\caption{Results obtained by using DMoN--DPR pooling using the best epsilon value of 0.001, distance weight of 1, variance weight of 0.001, and entropy weight of 0.001 on the PubMed dataset.}
\begin{tabular}{cccccc}
\toprule
\textbf{Seed} & \textbf{Method} & \textbf{Conductance} & \textbf{Modularity} & \textbf{NMI} & \textbf{F1} \\ \midrule
993 & DMoN--DPR (DVE) & 10.85 & 55.63 & 22.40 & 40.04 \\
550 & DMoN--DPR (DVE) & 7.66 & 58.80 & 26.86 & 49.54 \\
243 & DMoN--DPR (DVE) & 7.94 & 57.61 & 21.79 & 44.57 \\
16 & DMoN--DPR (DVE) & 8.02 & 57.57 & 21.69 & 44.34 \\
716 & DMoN--DPR (DVE) & 8.07 & 57.34 & 21.68 & 44.93 \\
383 & DMoN--DPR (DVE) & 9.81 & 56.09 & 22.81 & 37.08 \\
277 & DMoN--DPR (DVE) & 7.81 & 57.62 & 21.67 & 44.87 \\
274 & DMoN--DPR (DVE) & 7.83 & 57.51 & 21.69 & 44.93 \\
188 & DMoN--DPR (DVE) & 10.15 & 55.47 & 21.31 & 37.20 \\
796 & DMoN--DPR (DVE) & 7.97 & 57.50 & 21.75 & 44.87 \\
\midrule
\multicolumn{2}{c}{\textbf{Mean $\pm$ Std}} & 8.61 $\pm$ 1.18 & 57.11 $\pm$ 1.05 & 22.37 $\pm$ 1.64 & 43.24 $\pm$ 3.92 \\
\bottomrule
\end{tabular}
\label{tab:dpr_pubmed_dve}
\end{table}

\subsection{Coauthor CS}

The following tables list the performance of DiffPool, MinCut pooling, DMoN and different DMoN--DPR configurations across different seeds on the Coauthor CS dataset. 

\begin{table}[H]
\centering
\small
\caption{Results obtained by using DiffPool pooling on the Coauthor CS dataset.}
\begin{tabular}{cccccc}
\toprule
\textbf{Seed} & \textbf{Method} & \textbf{Conductance} & \textbf{Modularity} & \textbf{NMI} & \textbf{F1} \\ \midrule
993 & DiffPool & 16.20 & 61.90 & 53.06 & 53.73 \\
550 & DiffPool & 19.88 & 62.52 & 54.40 & 46.18 \\
243 & DiffPool & 20.41 & 59.39 & 49.01 & 49.43 \\
16 & DiffPool & 16.41 & 62.75 & 53.63 & 56.78 \\
716 & DiffPool & 16.64 & 61.46 & 54.03 & 57.47 \\
383 & DiffPool & 19.21 & 62.93 & 54.93 & 56.09 \\
277 & DiffPool & 20.95 & 63.84 & 56.25 & 47.35 \\
274 & DiffPool & 16.04 & 62.92 & 53.34 & 52.78 \\
188 & DiffPool & 20.58 & 61.21 & 56.02 & 56.85 \\
796 & DiffPool & 15.61 & 63.44 & 54.53 & 58.08 \\
\midrule
\multicolumn{2}{c}{\textbf{Mean $\pm$ Std}} & 18.19 $\pm$ 2.18 & 62.24 $\pm$ 1.30 & 53.92 $\pm$ 2.02 & 53.47 $\pm$ 4.40 \\
\bottomrule
\end{tabular}
\label{tab:diffpool_cs}
\end{table}

\begin{table}[H]
\centering
\small
\caption{Results obtained by using MinCut pooling on the Coauthor CS dataset.}
\begin{tabular}{cccccc}
\toprule
\textbf{Seed} & \textbf{Method} & \textbf{Conductance} & \textbf{Modularity} & \textbf{NMI} & \textbf{F1} \\ \midrule
993 & MinCut & 22.08 & 70.93 & 63.96 & 46.60 \\
550 & MinCut & 21.02 & 71.89 & 64.48 & 47.58 \\
243 & MinCut & 21.43 & 71.52 & 63.83 & 49.56 \\
16 & MinCut & 21.38 & 71.57 & 63.92 & 49.60 \\
716 & MinCut & 21.53 & 71.33 & 61.94 & 47.48 \\
383 & MinCut & 21.26 & 71.68 & 65.02 & 50.28 \\
277 & MinCut & 21.09 & 71.80 & 65.17 & 49.75 \\
274 & MinCut & 21.33 & 71.64 & 65.19 & 49.55 \\
188 & MinCut & 21.39 & 71.49 & 64.73 & 50.51 \\
796 & MinCut & 21.05 & 71.92 & 65.03 & 49.13 \\
\midrule
\multicolumn{2}{c}{\textbf{Mean $\pm$ Std}} & 21.36 $\pm$ 0.31 & 71.58 $\pm$ 0.29 & 64.33 $\pm$ 0.99 & 49.00 $\pm$ 1.31 \\
\bottomrule
\end{tabular}
\label{tab:mincut_cs}
\end{table}

\begin{table}[H]
\centering
\small
\caption{Results obtained by using DMoN pooling on the Coauthor CS dataset.}
\begin{tabular}{cccccc}
\toprule
\textbf{Seed} & \textbf{Method} & \textbf{Conductance} & \textbf{Modularity} & \textbf{NMI} & \textbf{F1} \\ \midrule
993 & DMoN & 18.28 & 72.47 & 68.53 & 58.97 \\
550 & DMoN & 18.01 & 72.56 & 67.90 & 57.26 \\
243 & DMoN & 18.50 & 72.80 & 70.05 & 59.31 \\
16 & DMoN & 18.04 & 72.93 & 70.70 & 62.98 \\
716 & DMoN & 18.23 & 72.24 & 68.54 & 62.21 \\
383 & DMoN & 17.72 & 73.06 & 70.11 & 60.39 \\
277 & DMoN & 20.05 & 72.04 & 66.73 & 54.74 \\
274 & DMoN & 18.13 & 73.24 & 71.11 & 62.35 \\
188 & DMoN & 18.49 & 73.22 & 71.11 & 60.66 \\
796 & DMoN & 20.81 & 71.42 & 67.82 & 53.72 \\
\midrule
\multicolumn{2}{c}{\textbf{Mean $\pm$ Std}} & 18.63 $\pm$ 0.99 & 72.60 $\pm$ 0.58 & 69.26 $\pm$ 1.55 & 59.26 $\pm$ 3.17 \\
\bottomrule
\end{tabular}
\label{tab:dmon_cs}
\end{table}

\begin{table}[H]
\centering
\small
\caption{Results obtained by using DMoN--DPR pooling using the best epsilon value of 1, and distance weight of 1 on the Coauthor CS dataset.}
\begin{tabular}{cccccc}
\toprule
\textbf{Seed} & \textbf{Method} & \textbf{Conductance} & \textbf{Modularity} & \textbf{NMI} & \textbf{F1} \\ \midrule
993 & DMoN--DPR (D) & 19.52 & 71.62 & 70.00 & 60.13 \\
550 & DMoN--DPR (D) & 18.63 & 72.19 & 70.76 & 61.55 \\
243 & DMoN--DPR (D) & 19.37 & 71.83 & 69.94 & 60.27 \\
16 & DMoN--DPR (D) & 19.14 & 72.53 & 72.67 & 64.33 \\
716 & DMoN--DPR (D) & 18.83 & 72.46 & 70.96 & 62.26 \\
383 & DMoN--DPR (D) & 19.05 & 72.16 & 71.26 & 62.23 \\
277 & DMoN--DPR (D) & 18.70 & 72.93 & 72.60 & 63.38 \\
274 & DMoN--DPR (D) & 18.46 & 72.51 & 71.16 & 61.65 \\
188 & DMoN--DPR (D) & 19.29 & 71.97 & 71.58 & 61.29 \\
796 & DMoN--DPR (D) & 17.94 & 72.77 & 70.78 & 61.08 \\
\midrule
\multicolumn{2}{c}{\textbf{Mean $\pm$ Std}} & 18.89 $\pm$ 0.48 & 72.30 $\pm$ 0.42 & 71.17 $\pm$ 0.93 & 61.82 $\pm$ 1.30 \\
\bottomrule
\end{tabular}
\label{tab:dpr_cs_d}
\end{table}

\begin{table}[H]
\centering
\small
\caption{Results obtained by using DMoN--DPR pooling using the best variance weight of 0.1 on the Coauthor CS dataset.}
\begin{tabular}{cccccc}
\toprule
\textbf{Seed} & \textbf{Method} & \textbf{Conductance} & \textbf{Modularity} & \textbf{NMI} & \textbf{F1} \\ \midrule
993 & DMoN--DPR (V) & 18.09 & 72.79 & 69.54 & 60.69 \\
550 & DMoN--DPR (V) & 19.20 & 72.05 & 69.23 & 60.66 \\
243 & DMoN--DPR (V) & 17.60 & 73.27 & 71.19 & 62.29 \\
16 & DMoN--DPR (V) & 18.45 & 72.43 & 68.76 & 60.54 \\
716 & DMoN--DPR (V) & 18.58 & 72.35 & 68.53 & 59.02 \\
383 & DMoN--DPR (V) & 19.05 & 72.25 & 69.61 & 59.12 \\
277 & DMoN--DPR (V) & 19.33 & 72.27 & 69.12 & 57.74 \\
274 & DMoN--DPR (V) & 18.62 & 72.97 & 70.38 & 62.66 \\
188 & DMoN--DPR (V) & 18.55 & 73.04 & 71.22 & 60.91 \\
796 & DMoN--DPR (V) & 20.01 & 71.56 & 67.98 & 55.77 \\
\midrule
\multicolumn{2}{c}{\textbf{Mean $\pm$ Std}} & 18.75 $\pm$ 0.68 & 72.50 $\pm$ 0.52 & 69.56 $\pm$ 1.08 & 59.94 $\pm$ 2.08 \\
\bottomrule
\end{tabular}
\label{tab:dpr_cs_v}
\end{table}

\begin{table}[H]
\centering
\small
\caption{Results obtained by using DMoN--DPR pooling using the best entropy weight of 0.1 on the Coauthor CS dataset.}
\begin{tabular}{cccccc}
\toprule
\textbf{Seed} & \textbf{Method} & \textbf{Conductance} & \textbf{Modularity} & \textbf{NMI} & \textbf{F1} \\ \midrule
993 & DMoN--DPR (E) & 20.73 & 69.18 & 70.29 & 57.08 \\
550 & DMoN--DPR (E) & 19.82 & 71.30 & 70.48 & 57.63 \\
243 & DMoN--DPR (E) & 19.35 & 71.39 & 72.46 & 64.00 \\
16 & DMoN--DPR (E) & 18.71 & 71.88 & 73.81 & 66.15 \\
716 & DMoN--DPR (E) & 18.77 & 72.20 & 74.36 & 66.49 \\
383 & DMoN--DPR (E) & 19.01 & 71.17 & 70.30 & 60.87 \\
277 & DMoN--DPR (E) & 21.91 & 69.36 & 70.12 & 59.57 \\
274 & DMoN--DPR (E) & 19.90 & 71.03 & 72.03 & 60.58 \\
188 & DMoN--DPR (E) & 19.49 & 71.84 & 73.02 & 62.90 \\
796 & DMoN--DPR (E) & 20.85 & 69.82 & 68.94 & 58.00 \\
\midrule
\multicolumn{2}{c}{\textbf{Mean $\pm$ Std}} & 19.85 $\pm$ 1.03 & 70.92 $\pm$ 1.08 & 71.58 $\pm$ 1.81 & 61.33 $\pm$ 3.43 \\
\bottomrule
\end{tabular}
\label{tab:dpr_cs_e}
\end{table}

\begin{table}[H]
\centering
\small
\caption{Results obtained by using DMoN--DPR pooling using the best epsilon value of 1, distance weight of 1, and variance weight of 0.1 on the Coauthor CS dataset.}
\begin{tabular}{cccccc}
\toprule
\textbf{Seed} & \textbf{Method} & \textbf{Conductance} & \textbf{Modularity} & \textbf{NMI} & \textbf{F1} \\ \midrule
993 & DMoN--DPR (DV) & 19.55 & 71.66 & 69.75 & 60.56 \\
550 & DMoN--DPR (DV) & 18.24 & 72.37 & 71.14 & 61.70 \\
243 & DMoN--DPR (DV) & 18.94 & 71.86 & 70.04 & 60.62 \\
16 & DMoN--DPR (DV) & 19.75 & 71.81 & 72.52 & 63.87 \\
716 & DMoN--DPR (DV) & 20.11 & 71.14 & 68.74 & 58.85 \\
383 & DMoN--DPR (DV) & 19.52 & 71.72 & 69.75 & 60.47 \\
277 & DMoN--DPR (DV) & 18.63 & 72.57 & 72.00 & 62.98 \\
274 & DMoN--DPR (DV) & 18.68 & 72.21 & 71.43 & 62.77 \\
188 & DMoN--DPR (DV) & 19.89 & 71.46 & 69.60 & 58.73 \\
796 & DMoN--DPR (DV) & 18.13 & 72.81 & 72.20 & 62.90 \\
\midrule
\multicolumn{2}{c}{\textbf{Mean $\pm$ Std}} & 19.14 $\pm$ 0.71 & 71.96 $\pm$ 0.52 & 70.72 $\pm$ 1.30 & 61.35 $\pm$ 1.78 \\
\bottomrule
\end{tabular}
\label{tab:dpr_cs_dv}
\end{table}

\begin{table}[H]
\centering
\small
\caption{Results obtained by using DMoN--DPR pooling using the best epsilon value of 1, distance weight of 1, and entropy weight of 0.1 on the Coauthor CS dataset.}
\begin{tabular}{cccccc}
\toprule
\textbf{Seed} & \textbf{Method} & \textbf{Conductance} & \textbf{Modularity} & \textbf{NMI} & \textbf{F1} \\ \midrule
993 & DMoN--DPR (DE) & 19.75 & 70.62 & 71.29 & 63.49 \\
550 & DMoN--DPR (DE) & 20.49 & 70.46 & 71.58 & 63.84 \\
243 & DMoN--DPR (DE) & 20.83 & 69.25 & 69.19 & 60.23 \\
16 & DMoN--DPR (DE) & 19.57 & 70.68 & 70.86 & 62.53 \\
716 & DMoN--DPR (DE) & 19.12 & 71.23 & 71.83 & 62.71 \\
383 & DMoN--DPR (DE) & 19.22 & 71.16 & 72.81 & 64.86 \\
277 & DMoN--DPR (DE) & 19.78 & 71.78 & 73.49 & 65.51 \\
274 & DMoN--DPR (DE) & 19.52 & 70.15 & 70.28 & 61.17 \\
188 & DMoN--DPR (DE) & 20.82 & 69.37 & 67.33 & 57.40 \\
796 & DMoN--DPR (DE) & 19.00 & 70.96 & 70.94 & 61.83 \\
\midrule
\multicolumn{2}{c}{\textbf{Mean $\pm$ Std}} & 19.81 $\pm$ 0.68 & 70.57 $\pm$ 0.80 & 70.96 $\pm$ 1.76 & 62.36 $\pm$ 2.37 \\
\bottomrule
\end{tabular}
\label{tab:dpr_cs_de}
\end{table}

\begin{table}[H]
\centering
\small
\caption{Results obtained by using DMoN--DPR pooling using the best variance weight of 0.1, and entropy weight of 0.1 on the Coauthor CS dataset.}
\begin{tabular}{cccccc}
\toprule
\textbf{Seed} & \textbf{Method} & \textbf{Conductance} & \textbf{Modularity} & \textbf{NMI} & \textbf{F1} \\ \midrule
993 & DMoN--DPR (VE) & 20.74 & 69.08 & 70.35 & 57.99 \\
550 & DMoN--DPR (VE) & 19.85 & 71.25 & 70.50 & 57.59 \\
243 & DMoN--DPR (VE) & 19.43 & 71.30 & 72.55 & 64.08 \\
16 & DMoN--DPR (VE) & 18.74 & 71.85 & 73.81 & 66.17 \\
716 & DMoN--DPR (VE) & 19.22 & 71.37 & 73.19 & 65.61 \\
383 & DMoN--DPR (VE) & 18.65 & 71.43 & 70.49 & 60.75 \\
277 & DMoN--DPR (VE) & 21.92 & 69.32 & 70.10 & 59.50 \\
274 & DMoN--DPR (VE) & 19.93 & 70.96 & 71.96 & 60.58 \\
188 & DMoN--DPR (VE) & 19.44 & 71.87 & 73.01 & 63.01 \\
796 & DMoN--DPR (VE) & 20.99 & 69.70 & 68.72 & 57.97 \\
\midrule
\multicolumn{2}{c}{\textbf{Mean $\pm$ Std}} & 19.89 $\pm$ 1.04 & 70.81 $\pm$ 1.04 & 71.47 $\pm$ 1.66 & 61.33 $\pm$ 3.21 \\
\bottomrule
\end{tabular}
\label{tab:dpr_cs_ve}
\end{table}

\begin{table}[H]
\centering
\small
\caption{Results obtained by using DMoN--DPR pooling using the best epsilon value of 1, distance weight of 1, variance weight of 0.1, and entropy weight of 0.1 on the Coauthor CS dataset.}
\begin{tabular}{cccccc}
\toprule
\textbf{Seed} & \textbf{Method} & \textbf{Conductance} & \textbf{Modularity} & \textbf{NMI} & \textbf{F1} \\ \midrule
993 & DMoN--DPR (DVE) & 20.34 & 70.31 & 70.86 & 63.39 \\
550 & DMoN--DPR (DVE) & 20.09 & 70.77 & 71.87 & 64.16 \\
243 & DMoN--DPR (DVE) & 19.56 & 70.27 & 70.03 & 60.73 \\
16 & DMoN--DPR (DVE) & 20.21 & 70.65 & 72.33 & 63.54 \\
716 & DMoN--DPR (DVE) & 19.26 & 71.08 & 71.81 & 62.97 \\
383 & DMoN--DPR (DVE) & 19.17 & 71.20 & 72.88 & 64.81 \\
277 & DMoN--DPR (DVE) & 18.96 & 72.10 & 73.54 & 65.43 \\
274 & DMoN--DPR (DVE) & 19.65 & 70.36 & 70.79 & 61.93 \\
188 & DMoN--DPR (DVE) & 20.98 & 69.20 & 67.10 & 57.05 \\
796 & DMoN--DPR (DVE) & 18.70 & 71.14 & 71.62 & 62.70 \\
\midrule
\multicolumn{2}{c}{\textbf{Mean $\pm$ Std}} & 19.69 $\pm$ 0.71 & 70.71 $\pm$ 0.76 & 71.28 $\pm$ 1.79 & 62.67 $\pm$ 2.40 \\
\bottomrule
\end{tabular}
\label{tab:dpr_cs_dve}
\end{table}

\subsection{Coauthor Physics}

The following tables list the performance of DiffPool, MinCut pooling, DMoN and different DMoN--DPR configurations across different seeds on the Coauthor Physics dataset.

\begin{table}[H]
\centering
\small
\caption{Results obtained by using DiffPool pooling on the Coauthor Physics dataset.}
\begin{tabular}{cccccc}
\toprule
\textbf{Seed} & \textbf{Method} & \textbf{Conductance} & \textbf{Modularity} & \textbf{NMI} & \textbf{F1} \\ \midrule
993 & DiffPool & 13.06 & 56.73 & 57.94 & 53.93 \\
550 & DiffPool & 15.25 & 59.42 & 59.33 & 53.71 \\
243 & DiffPool & 13.06 & 57.18 & 56.61 & 52.20 \\
16 & DiffPool & 13.26 & 56.55 & 57.69 & 53.75 \\
716 & DiffPool & 14.42 & 58.97 & 61.09 & 53.65 \\
383 & DiffPool & 14.03 & 57.15 & 58.03 & 54.06 \\
277 & DiffPool & 13.37 & 56.48 & 47.24 & 46.20 \\
274 & DiffPool & 13.26 & 55.70 & 58.06 & 56.79 \\
188 & DiffPool & 12.23 & 57.04 & 58.30 & 54.31 \\
796 & DiffPool & 17.18 & 59.19 & 47.89 & 38.47 \\
\midrule
\multicolumn{2}{c}{\textbf{Mean $\pm$ Std}} & 13.91 $\pm$ 1.42 & 57.44 $\pm$ 1.29 & 56.22 $\pm$ 4.71 & 51.71 $\pm$ 5.38 \\
\bottomrule
\end{tabular}
\label{tab:diffpool_physics}
\end{table}

\begin{table}[H]
\centering
\small
\caption{Results obtained by using MinCut pooling on the Coauthor Physics dataset.}
\begin{tabular}{cccccc}
\toprule
\textbf{Seed} & \textbf{Method} & \textbf{Conductance} & \textbf{Modularity} & \textbf{NMI} & \textbf{F1} \\ \midrule
993 & MinCut & 14.08 & 61.78 & 48.91 & 45.03 \\
550 & MinCut & 14.35 & 61.54 & 56.89 & 51.00 \\
243 & MinCut & 13.74 & 61.79 & 51.41 & 47.76 \\
16 & MinCut & 13.38 & 62.16 & 57.19 & 51.39 \\
716 & MinCut & 13.18 & 62.55 & 57.10 & 51.27 \\
383 & MinCut & 13.95 & 62.12 & 48.62 & 42.59 \\
277 & MinCut & 14.17 & 61.39 & 48.28 & 46.92 \\
274 & MinCut & 14.20 & 61.31 & 45.80 & 46.44 \\
188 & MinCut & 13.33 & 62.51 & 52.14 & 47.82 \\
796 & MinCut & 14.95 & 60.38 & 47.53 & 47.70 \\
\midrule
\multicolumn{2}{c}{\textbf{Mean $\pm$ Std}} & 13.93 $\pm$ 0.54 & 61.75 $\pm$ 0.65 & 51.39 $\pm$ 4.30 & 47.79 $\pm$ 2.84 \\
\bottomrule
\end{tabular}
\label{tab:mincut_physics}
\end{table}

\begin{table}[H]
\centering
\small
\caption{Results obtained by using DMoN pooling on the Coauthor Physics dataset.}
\begin{tabular}{cccccc}
\toprule
\textbf{Seed} & \textbf{Method} & \textbf{Conductance} & \textbf{Modularity} & \textbf{NMI} & \textbf{F1} \\ \midrule
993 & DMoN & 12.64 & 59.92 & 44.49 & 46.04 \\
550 & DMoN & 13.87 & 64.01 & 54.36 & 47.54 \\
243 & DMoN & 13.89 & 63.77 & 56.67 & 49.51 \\
16 & DMoN & 13.82 & 63.77 & 56.89 & 49.55 \\
716 & DMoN & 13.93 & 63.72 & 55.42 & 49.02 \\
383 & DMoN & 13.65 & 63.63 & 52.66 & 46.65 \\
277 & DMoN & 13.48 & 63.85 & 50.84 & 44.74 \\
274 & DMoN & 13.84 & 64.02 & 53.53 & 46.36 \\
188 & DMoN & 13.89 & 63.73 & 55.87 & 49.33 \\
796 & DMoN & 14.02 & 64.12 & 54.29 & 46.36 \\
\midrule
\multicolumn{2}{c}{\textbf{Mean $\pm$ Std}} & 13.70 $\pm$ 0.40 & 63.45 $\pm$ 1.25 & 53.50 $\pm$ 3.67 & 47.51 $\pm$ 1.73 \\
\bottomrule
\end{tabular}
\label{tab:dmon_phys}
\end{table}

\begin{table}[H]
\centering
\small
\caption{Results obtained by using DMoN--DPR pooling using the best epsilon value of 10, and distance weight of 1 on the Coauthor Physics dataset.}
\begin{tabular}{cccccc}
\toprule
\textbf{Seed} & \textbf{Method} & \textbf{Conductance} & \textbf{Modularity} & \textbf{NMI} & \textbf{F1} \\ \midrule
993 & DMoN--DPR (D) & 14.00 & 62.17 & 51.59 & 47.84 \\
550 & DMoN--DPR (D) & 19.30 & 56.61 & 50.82 & 46.32 \\
243 & DMoN--DPR (D) & 18.21 & 57.77 & 58.61 & 50.55 \\
16 & DMoN--DPR (D) & 16.19 & 54.39 & 53.19 & 55.67 \\
716 & DMoN--DPR (D) & 16.93 & 55.94 & 48.09 & 49.54 \\
383 & DMoN--DPR (D) & 19.58 & 55.58 & 52.71 & 46.44 \\
277 & DMoN--DPR (D) & 14.97 & 55.73 & 51.49 & 53.53 \\
274 & DMoN--DPR (D) & 15.55 & 56.74 & 59.80 & 55.95 \\
188 & DMoN--DPR (D) & 14.85 & 55.96 & 55.95 & 54.65 \\
796 & DMoN--DPR (D) & 14.78 & 58.04 & 52.68 & 49.37 \\
\midrule
\multicolumn{2}{c}{\textbf{Mean $\pm$ Std}} & 16.44 $\pm$ 1.99 & 56.89 $\pm$ 2.14 & 53.49 $\pm$ 3.61 & 50.99 $\pm$ 3.71 \\
\bottomrule
\end{tabular}
\label{tab:dpr_phys_d}
\end{table}

\begin{table}[H]
\centering
\small
\caption{Results obtained by using DMoN--DPR pooling using the best variance weight of 1 on the Coauthor Physics dataset.}
\begin{tabular}{cccccc}
\toprule
\textbf{Seed} & \textbf{Method} & \textbf{Conductance} & \textbf{Modularity} & \textbf{NMI} & \textbf{F1} \\ \midrule
993 & DMoN--DPR (V) & 12.28 & 58.05 & 45.73 & 45.50 \\
550 & DMoN--DPR (V) & 11.94 & 57.38 & 54.27 & 54.78 \\
243 & DMoN--DPR (V) & 13.49 & 60.85 & 59.11 & 53.04 \\
16 & DMoN--DPR (V) & 13.55 & 58.65 & 58.50 & 56.62 \\
716 & DMoN--DPR (V) & 11.83 & 58.84 & 56.84 & 53.74 \\
383 & DMoN--DPR (V) & 14.44 & 57.84 & 54.21 & 55.65 \\
277 & DMoN--DPR (V) & 12.03 & 58.71 & 57.09 & 53.73 \\
274 & DMoN--DPR (V) & 14.15 & 57.02 & 53.90 & 55.44 \\
188 & DMoN--DPR (V) & 13.11 & 57.40 & 54.62 & 53.97 \\
796 & DMoN--DPR (V) & 15.02 & 61.14 & 45.01 & 45.34 \\
\midrule
\multicolumn{2}{c}{\textbf{Mean $\pm$ Std}} & 13.18 $\pm$ 1.14 & 58.59 $\pm$ 1.41 & 53.93 $\pm$ 4.87 & 52.78 $\pm$ 4.02 \\
\bottomrule
\end{tabular}
\label{tab:dpr_phys_v}
\end{table}

\begin{table}[H]
\centering
\small
\caption{Results obtained by using DMoN--DPR pooling using the best entropy weight of 0.1 on the Coauthor Physics dataset.}
\begin{tabular}{cccccc}
\toprule
\textbf{Seed} & \textbf{Method} & \textbf{Conductance} & \textbf{Modularity} & \textbf{NMI} & \textbf{F1} \\ \midrule
993 & DMoN--DPR (E) & 13.59 & 59.64 & 48.13 & 45.60 \\
550 & DMoN--DPR (E) & 11.94 & 57.60 & 54.02 & 54.19 \\
243 & DMoN--DPR (E) & 13.72 & 61.90 & 58.72 & 51.62 \\
16 & DMoN--DPR (E) & 13.53 & 58.94 & 57.15 & 55.16 \\
716 & DMoN--DPR (E) & 14.17 & 57.80 & 47.71 & 51.12 \\
383 & DMoN--DPR (E) & 14.24 & 58.41 & 53.68 & 54.15 \\
277 & DMoN--DPR (E) & 11.57 & 58.08 & 53.16 & 52.07 \\
274 & DMoN--DPR (E) & 13.26 & 57.99 & 53.77 & 51.77 \\
188 & DMoN--DPR (E) & 14.08 & 60.75 & 55.54 & 50.03 \\
796 & DMoN--DPR (E) & 14.81 & 62.19 & 46.39 & 45.17 \\
\midrule
\multicolumn{2}{c}{\textbf{Mean $\pm$ Std}} & 13.49 $\pm$ 1.02 & 59.33 $\pm$ 1.72 & 52.83 $\pm$ 4.13 & 51.09 $\pm$ 3.39 \\
\bottomrule
\end{tabular}
\label{tab:dpr_phys_e}
\end{table}

\begin{table}[H]
\centering
\small
\caption{Results obtained by using DMoN--DPR pooling using the best epsilon value of 10, distance weight of 1, and variance weight 1 on the Coauthor Physics dataset.}
\begin{tabular}{cccccc}
\toprule
\textbf{Seed} & \textbf{Method} & \textbf{Conductance} & \textbf{Modularity} & \textbf{NMI} & \textbf{F1} \\ \midrule
993 & DMoN--DPR (DV) & 15.37 & 55.95 & 48.78 & 53.49 \\
550 & DMoN--DPR (DV) & 14.84 & 55.32 & 52.13 & 57.57 \\
243 & DMoN--DPR (DV) & 13.82 & 56.19 & 56.71 & 57.69 \\
16 & DMoN--DPR (DV) & 11.67 & 56.14 & 54.87 & 57.53 \\
716 & DMoN--DPR (DV) & 10.97 & 57.41 & 62.02 & 61.81 \\
383 & DMoN--DPR (DV) & 13.45 & 56.41 & 54.45 & 57.87 \\
277 & DMoN--DPR (DV) & 14.25 & 56.26 & 52.86 & 57.93 \\
274 & DMoN--DPR (DV) & 11.79 & 56.58 & 57.37 & 59.70 \\
188 & DMoN--DPR (DV) & 14.17 & 56.86 & 60.24 & 58.29 \\
796 & DMoN--DPR (DV) & 13.98 & 57.70 & 58.95 & 58.04 \\
\midrule
\multicolumn{2}{c}{\textbf{Mean $\pm$ Std}} & 13.43 $\pm$ 1.46 & 56.48 $\pm$ 0.70 & 55.84 $\pm$ 4.02 & 57.99 $\pm$ 2.06 \\
\bottomrule
\end{tabular}
\label{tab:dpr_phys_dv}
\end{table}

\begin{table}[H]
\centering
\small
\caption{Results obtained by using DMoN--DPR pooling using the best epsilon value of 10, distance weight of 1, and entropy weight of 0.1 on the Coauthor Physics dataset.}
\begin{tabular}{cccccc}
\toprule
\textbf{Seed} & \textbf{Method} & \textbf{Conductance} & \textbf{Modularity} & \textbf{NMI} & \textbf{F1} \\ \midrule
993 & DMoN--DPR (DE) & 13.94 & 61.00 & 47.60 & 46.37 \\
550 & DMoN--DPR (DE) & 14.11 & 56.74 & 54.67 & 55.98 \\
243 & DMoN--DPR (DE) & 13.89 & 57.00 & 60.05 & 58.60 \\
16 & DMoN--DPR (DE) & 13.90 & 54.44 & 53.68 & 57.90 \\
716 & DMoN--DPR (DE) & 11.35 & 57.47 & 63.18 & 61.39 \\
383 & DMoN--DPR (DE) & 15.41 & 54.93 & 51.58 & 53.92 \\
277 & DMoN--DPR (DE) & 15.84 & 54.43 & 51.48 & 57.57 \\
274 & DMoN--DPR (DE) & 14.14 & 57.42 & 53.66 & 56.28 \\
188 & DMoN--DPR (DE) & 15.33 & 56.20 & 56.80 & 55.88 \\
796 & DMoN--DPR (DE) & 13.83 & 59.24 & 47.47 & 46.26 \\
\midrule
\multicolumn{2}{c}{\textbf{Mean $\pm$ Std}} & 14.17 $\pm$ 1.24 & 56.89 $\pm$ 2.09 & 54.02 $\pm$ 5.00 & 55.02 $\pm$ 4.99 \\
\bottomrule
\multicolumn{2}{c}{\textbf{Mean}} & 14.22 & 56.10 & 54.71 & 56.94 \\
\multicolumn{2}{c}{\textbf{Variance}} & 0.01 & 0.01 & 0.16 & 0.10 \\ \hline
\end{tabular}
\label{tab:dpr_phys_de}
\end{table}

\begin{table}[H]
\centering
\small
\caption{Results obtained by using DMoN--DPR pooling using the best variance weight of 1, and entropy weight 0f 0.1 on the Coauthor Physics dataset.}
\begin{tabular}{cccccc}
\toprule
\textbf{Seed} & \textbf{Method} & \textbf{Conductance} & \textbf{Modularity} & \textbf{NMI} & \textbf{F1} \\ \midrule
993 & DMoN--DPR (VE) & 8.51 & 36.26 & 14.21 & 44.28 \\
550 & DMoN--DPR (VE) & 13.11 & 57.01 & 52.67 & 57.06 \\
243 & DMoN--DPR (VE) & 14.40 & 59.91 & 56.09 & 52.22 \\
16 & DMoN--DPR (VE) & 13.97 & 58.17 & 62.31 & 59.21 \\
716 & DMoN--DPR (VE) & 13.66 & 58.24 & 53.74 & 53.47 \\
383 & DMoN--DPR (VE) & 14.12 & 57.60 & 56.52 & 57.86 \\
277 & DMoN--DPR (VE) & 16.66 & 60.87 & 48.23 & 44.09 \\
274 & DMoN--DPR (VE) & 12.18 & 56.56 & 51.47 & 56.39 \\
188 & DMoN--DPR (VE) & 11.16 & 56.05 & 56.50 & 60.71 \\
796 & DMoN--DPR (VE) & 13.32 & 58.12 & 47.75 & 50.50 \\
\midrule
\multicolumn{2}{c}{\textbf{Mean $\pm$ Std}} & 13.11 $\pm$ 2.17 & 55.88 $\pm$ 7.04 & 49.95 $\pm$ 13.27 & 53.58 $\pm$ 5.86 \\
\bottomrule
\end{tabular}
\label{tab:dpr_phys_ve}
\end{table}

\begin{table}[H]
\centering
\small
\caption{Results obtained by using DMoN--DPR pooling using the best epsilon value of 10, distance weight of 1, variance weight of 1, and entropy weight of 0.1 on the Coauthor Physics dataset.}
\begin{tabular}{cccccc}
\toprule
\textbf{Seed} & \textbf{Method} & \textbf{Conductance} & \textbf{Modularity} & \textbf{NMI} & \textbf{F1} \\ \midrule
993 & DMoN--DPR (DVE) & 14.25 & 57.43 & 42.12 & 46.26 \\
550 & DMoN--DPR (DVE) & 14.53 & 55.66 & 53.54 & 57.87 \\
243 & DMoN--DPR (DVE) & 12.50 & 55.96 & 54.67 & 59.77 \\
16 & DMoN--DPR (DVE) & 13.27 & 54.52 & 51.00 & 57.72 \\
716 & DMoN--DPR (DVE) & 12.95 & 54.76 & 54.90 & 58.06 \\
383 & DMoN--DPR (DVE) & 12.76 & 55.00 & 56.93 & 60.32 \\
277 & DMoN--DPR (DVE) & 14.47 & 54.87 & 52.66 & 59.00 \\
274 & DMoN--DPR (DVE) & 12.47 & 55.39 & 58.52 & 61.38 \\
188 & DMoN--DPR (DVE) & 11.29 & 57.23 & 62.59 & 62.06 \\
796 & DMoN--DPR (DVE) & 9.90 & 53.89 & 48.04 & 57.18 \\
\midrule
\multicolumn{2}{c}{\textbf{Mean $\pm$ Std}} & 12.84 $\pm$ 1.45 & 55.47 $\pm$ 1.14 & 53.50 $\pm$ 5.67 & 57.96 $\pm$ 4.42 \\
\bottomrule
\end{tabular}
\label{tab:dpr_phys_dve}
\end{table}

\section{Implementation Details}

The code was implemented by extending the DMoN implementation in PyTorch Geometric \citep{Fey/Lenssen/2019}, and was trained and evaluated using the evaluation protocols found in the official DMoN repository \citep{tsitsulin2023graph}. The experiments were run on an A100 GPU with 40GB of memory offered by Google Colab Pro. Regarding the runtime analysis results and clustering visualization, the code was run on an Apple M2 Max CPU.

%% file: paper/results_appendix.tex
\section{Quantification of Diversity}
\label{sec:quantification_of_diversity}

\input{paper/diversity_tables}

The diversity diagnostics in Tables \ref{tab:diversity_1}–\ref{tab:diversity_2} echo the accuracy gains: augmenting DMoN with our diversity-preserving regularizers widens the geometric spread of clusters—captured by the average inter-centroid distance (AICD) and minimum inter-centroid distance (MICD)—while largely preserving internal cohesion, as indicated by the near-constant average intra-cluster variance (AICV) and Silhouette score (Sil). On the citation graphs, most DPR variants leave AICD and MICD within a few percent of the DMoN baseline (the largest jump is $\approx 10 \%$ on Cora for V/VE), while AICV and Silhouette score remain virtually unchanged, confirming that clusters are not overstretched when the feature signal is weak. The feature-rich Coauthor graphs display a stronger effect: the entropy-weighted VE model, for example, boosts AICD on Coauthor CS from 20.5 to 31.8 and MICD from 8.2 to 11.2—at the cost of higher AICV—while hybrids that include the distance term (DE, DVE) achieve a better trade-off, enlarging centroid spacing by $15-25 \%$ but capping the rise in AICV so that Silhouette score stays on par with, or slightly above, the DMoN baseline. The sole exception is Coauthor Physics, where Silhouette score dips because DPR variants both push centroids outward and allow somewhat broader clusters; this reduces compactness even as separation improves. Overall, DPR regularization fulfills its design goal: clusters become globally more dispersed yet remain locally cohesive, yielding modest benefits on low-feature graphs and pronounced gains when node attributes are rich and heterogeneous. Note that AICV measures embedding variance within clusters, whereas our V term maximizes assignment variance across nodes; any changes in AICV under V are therefore indirect.

\section{Visualization of Clusters}
\label{sec:visualization}

\input{paper/plots}

The clustering results visualized in Figure \ref{fig:clusters}, obtained using t--SNE with a fixed perplexity (30) and learning rate (200), demonstrate that DMoN–DPR generally outperforms DMoN across the benchmark datasets Cora, CiteSeer, PubMed, Coauthor CS, and Coauthor Physics. For Cora and CiteSeer datasets, DMoN–DPR produces tighter, more well-separated clusters with improved alignment to the ground truth labels, contrasting with the more scattered and less defined clusters formed by DMoN. On the other hand, PubMed remains challenging for both methods, resulting in notable overlap and scattered clusters; yet, DMoN–DPR still provides slightly tighter groupings. Turning to the Coauthor CS and Coauthor Physics datasets, a similar trend emerges. While DMoN adequately captures overarching modular structures, it can occasionally struggle to separate nuanced subgroups. In contrast, DMoN–DPR consistently yields more cohesive and better-separated clusters, offering clearer decision boundaries and higher alignment with the true labels. This improvement is particularly notable in Coauthor CS, where DMoN–DPR visibly reduces overlap and sharpens cluster delineations. These findings underscore the effectiveness of promoting diversity within hidden representations, as DMoN–DPR not only excels in forming compact and accurate clusters for datasets with greater feature variability but also remains competitive on more challenging datasets.

%% file: paper/diversity_tables.tex
\begin{table*}[ht]
\centering
\small
\setlength{\tabcolsep}{4pt} 
\renewcommand{\arraystretch}{1.2} 
\caption{Comparison of diversity resulted from the clustering methods on three datasets (Cora, CiteSeer, PubMed). AICD is average inter-centroid distance, MICD is minimum inter-centroid distance, AICV is average intra-cluster variance, and Sil is the Silhouette score.}
\begin{tabular}{@{}lcccccccccccc@{}}
\toprule
\multirow{2}{*}{\textbf{Method}} & \multicolumn{4}{c}{\textbf{Cora}} & \multicolumn{4}{c}{\textbf{CiteSeer}} & \multicolumn{4}{c}{\textbf{PubMed}} \\
\cmidrule(lr){2-5} \cmidrule(lr){6-9} \cmidrule(lr){10-13} 
 & AICD & MICD & AICV  & Sil  & AICD & MICD & AICV  & Sil & AICD & MICD & AICV  & Sil \\
\midrule
DiffPool & 3.69 & 2.62 & 0.02 & 0.16 & 5.03 & 4.13 & 0.02 & 0.25 & \multicolumn{4}{c}{--} \\
MinCut & 6.37 & 5.42 & 0.02 & 0.36 & 9.03 & 8.18 & 0.03 & 0.46 & 3.03 & 2.62 & 0.00 & 0.46 \\
DMoN & 8.10 & 5.14 & 0.04 & 0.27 & 10.11 & 7.34 & 0.04 & 0.37 & 2.77 & 2.59 & 0.00 & 0.48  \\
\midrule
DPR(D) & 8.01 & 4.99 & 0.04 & 0.26 & 10.15 & 7.33 & 0.04 & 0.37 & 2.77 & 2.59 & 0.00 & 0.48  \\
DPR(V) & 8.93 & 5.25 & 0.05 & 0.23 & 10.20 & 7.17 & 0.05 & 0.35 & 2.77 & 2.59 & 0.00 & 0.48  \\
DPR(E) & 8.12 & 5.14 & 0.04 & 0.2 & 10.31 & 7.14 & 0.05 & 0.35 & 2.77 & 2.59 & 0.00 & 0.48 \\
DPR(DV) & 8.12 & 4.80 & 0.04 & 0.26 & 10.26 & 7.26 & 0.05 & 0.35 & 2.77 & 2.59 & 0.00 & 0.48  \\
DPR(DE) & 8.03 & 4.88 & 0.04 & 0.26 & 10.35 & 7.23 & 0.05 & 0.35 & 2.78 & 2.60 & 0.00 & 0.48  \\
DPR(VE) & 8.93 & 5.25 & 0.05 & 0.23 & 10.39 & 7.12 & 0.05 & 0.34 & 2.77 & 2.59 & 0.00 & 0.48\\
DPR(DVE) & 8.13 & 4.81 & 0.04 & 0.26 & 10.43 & 7.07 & 0.05 & 0.33 & 2.78 & 2.60 & 0.00 & 0.48  \\
\bottomrule
\end{tabular}
\label{tab:diversity_1}
\end{table*}

\begin{table}[ht]
\centering
\small
\setlength{\tabcolsep}{4pt} 
\renewcommand{\arraystretch}{1.2} 
\caption{Comparison of diversity resulted from the clustering methods on two datasets (Coauthor CS, Coauthor Physics). AICD is average inter-centroid distance, MICD is minimum inter-centroid distance, AICV is average intra-cluster variance, and Sil is the Silhouette score.}
\begin{tabular}{@{}lcccccccc@{}}
\toprule
\multirow{2}{*}{\textbf{Method}} & \multicolumn{4}{c}{\textbf{Coauthor CS}} & \multicolumn{4}{c}{\textbf{Coauthor Physics}} \\
\cmidrule(lr){2-5} \cmidrule(lr){6-9}
 & AICD & MICD & AICV  & Sil & AICD & MICD & AICV  & Sil\\
\midrule
DiffPool & 6.05 & 2.93 & 0.03 & 0.12 & 4.35 & 2.14 & 0.01 & 0.23\\
MinCut & 10.46 & 6.89 & 0.06 & 0.27 & 10.88 & 8.56 & 0.04 & 0.42 \\
DMoN & 20.47 & 8.15 & 0.26 & 0.17 & 13.97 & 10.51 & 0.09 & 0.37  \\
\midrule
DPR(D) & 17.72 & 7.01 & 0.19 & 0.18 & 14.55 & 7.27 & 0.11 & 0.23\\
DPR(V) & 21.02 & 8.33 & 0.27 & 0.17 & 17.44 & 9.79 & 0.16 & 0.27 \\
DPR(E) & 31.67 & 11.14 & 0.58 & 0.15 & 14.72 & 7.46 & 0.11 & 0.25\\
DPR(DV) & 18.16 & 6.94 & 0.20 & 0.17 & 14.72 & 7.46 & 0.11 & 0.25 \\
DPR(DE) & 24.48 & 9.38 & 0.37 & 0.19 & 14.58 & 7.66 & 0.11 & 0.25   \\
DPR(VE) & 31.75 & 11.17 & 0.59 & 0.15  & 19.16 & 11.15 & 0.19 & 0.29 \\
DPR(DVE) & 24.66 & 9.79 & 0.36 & 0.18 & 14.31 & 7.73 & 0.11 & 0.26 \\
\bottomrule
\end{tabular}
\label{tab:diversity_2}
\end{table}

%% file: paper/plots.tex
\begin{figure*}[!t]
\centering
\begin{subfigure}[t]{0.32\textwidth} 
  \includegraphics[width=\linewidth]{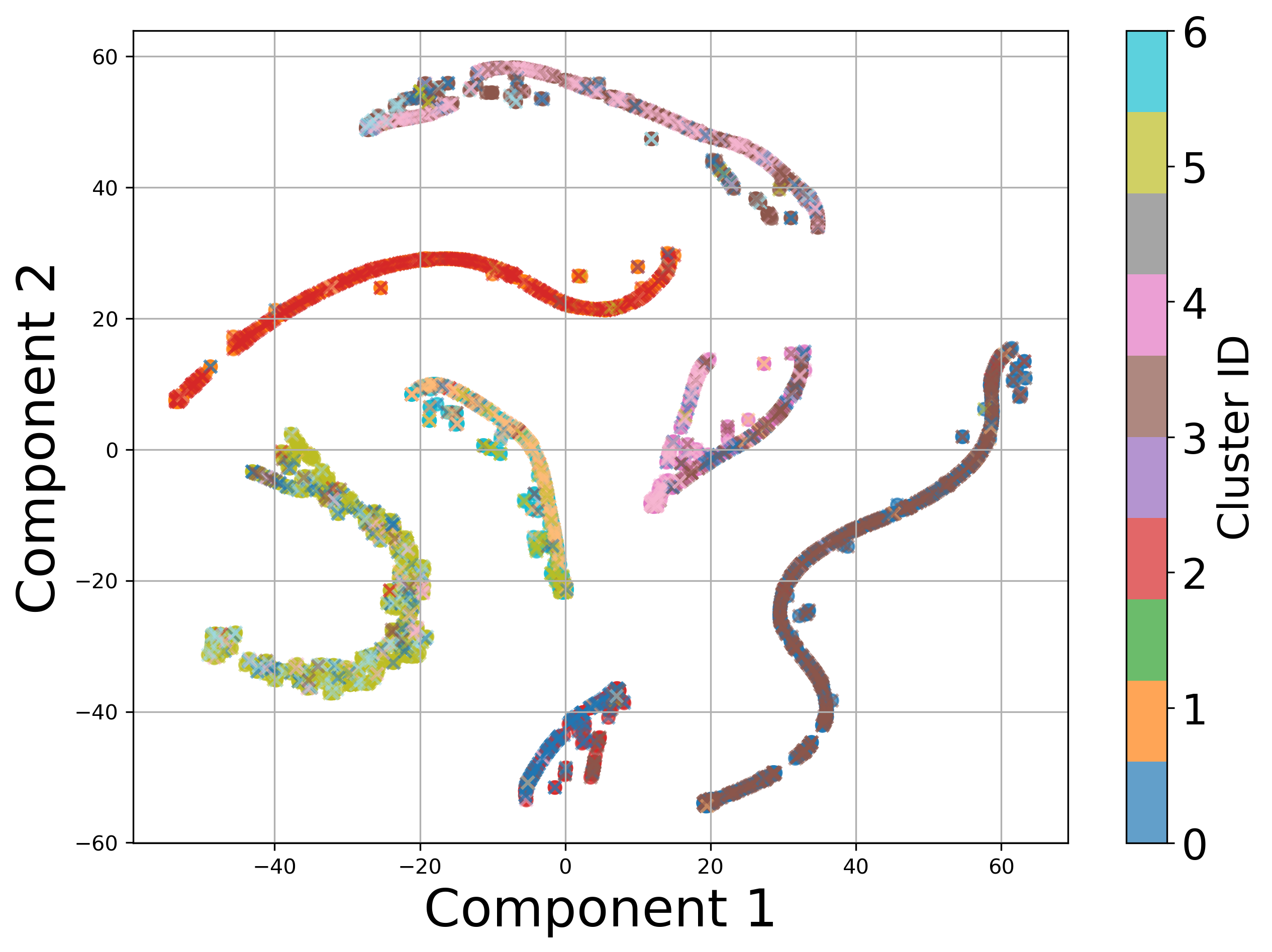}
  \captionsetup{justification=centering}
  \caption{}
  \label{fig:cora_dmon}
\end{subfigure}
\begin{subfigure}[t]{0.32\textwidth} 
  \includegraphics[width=\linewidth]{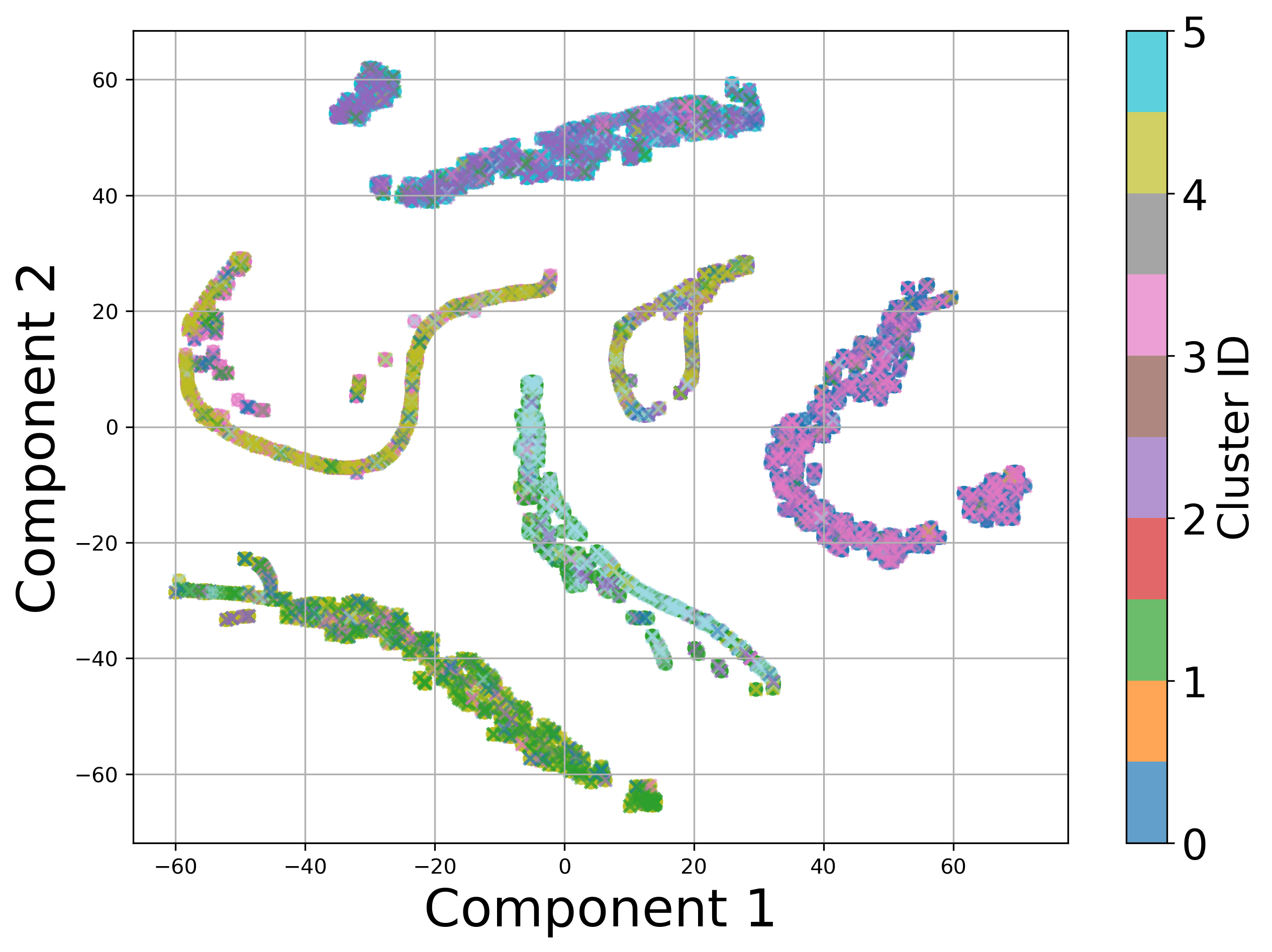}
  \captionsetup{justification=centering}
  \caption{}
  \label{fig:citeseer_dmon}
\end{subfigure}
\begin{subfigure}[t]{0.34\textwidth} 
  \includegraphics[width=\linewidth]{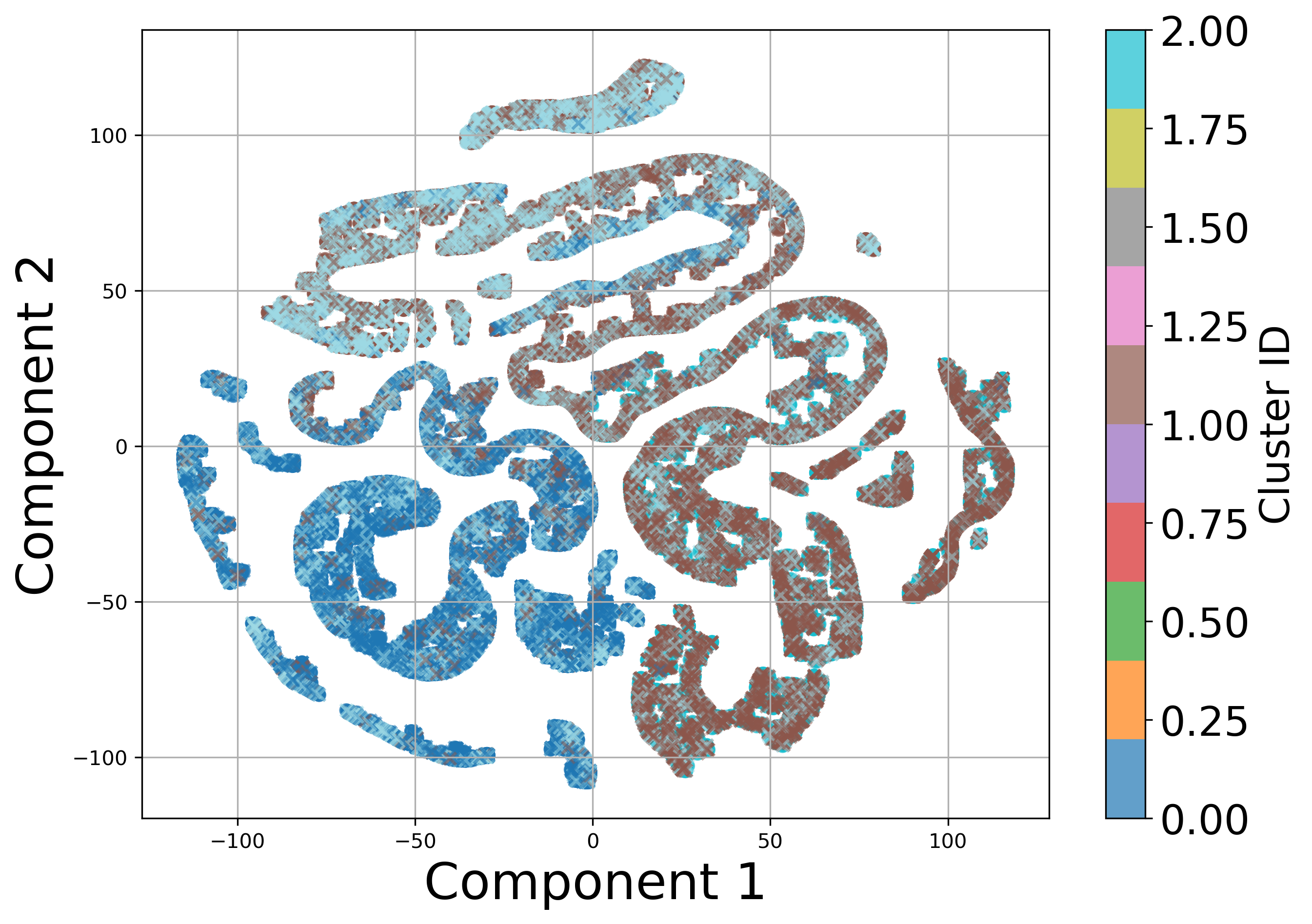}
  \captionsetup{justification=centering}
  \caption{}
  \label{fig:pubmed_dmon}
\end{subfigure}
\begin{subfigure}[t]{0.32\textwidth} 
  \includegraphics[width=\linewidth]{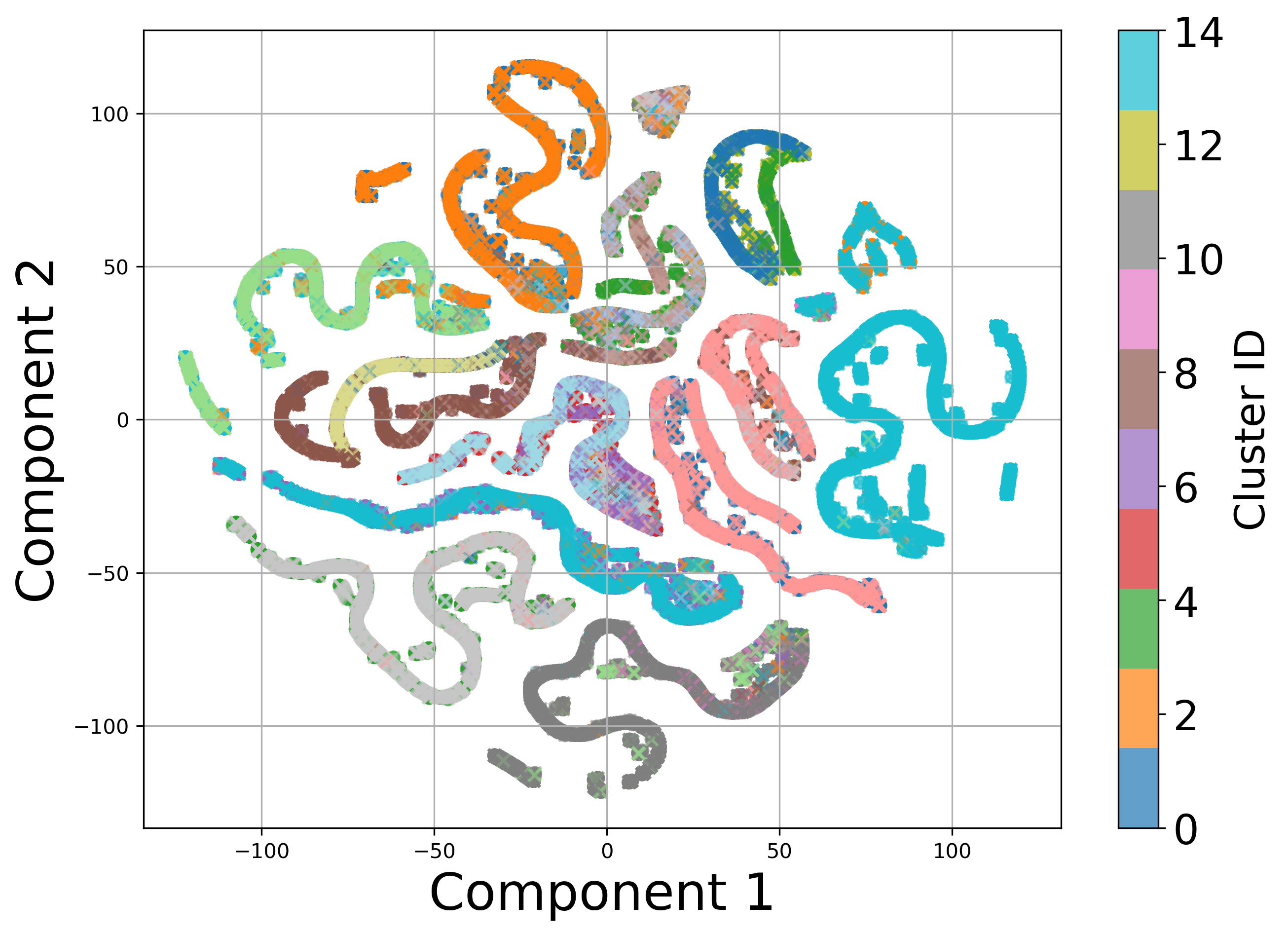}
  \captionsetup{justification=centering}
  \caption{}
  \label{fig:cs_dmon}
\end{subfigure}
\begin{subfigure}[t]{0.32\textwidth} 
  \includegraphics[width=\linewidth]{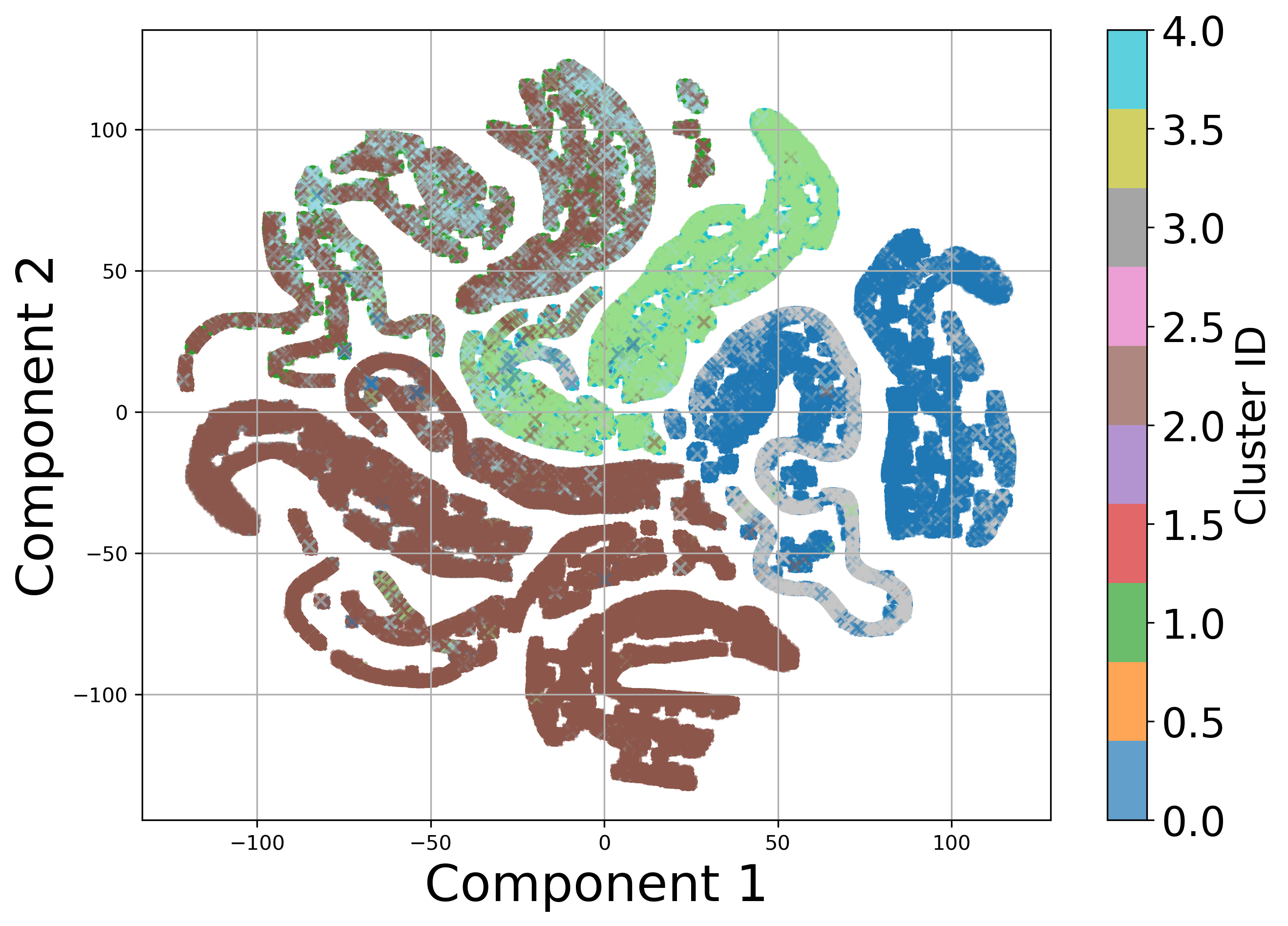}
  \captionsetup{justification=centering}
  \caption{}
  \label{fig:physics_dmon}
\end{subfigure}

\begin{subfigure}[t]{0.32\textwidth} 
  \includegraphics[width=\linewidth]{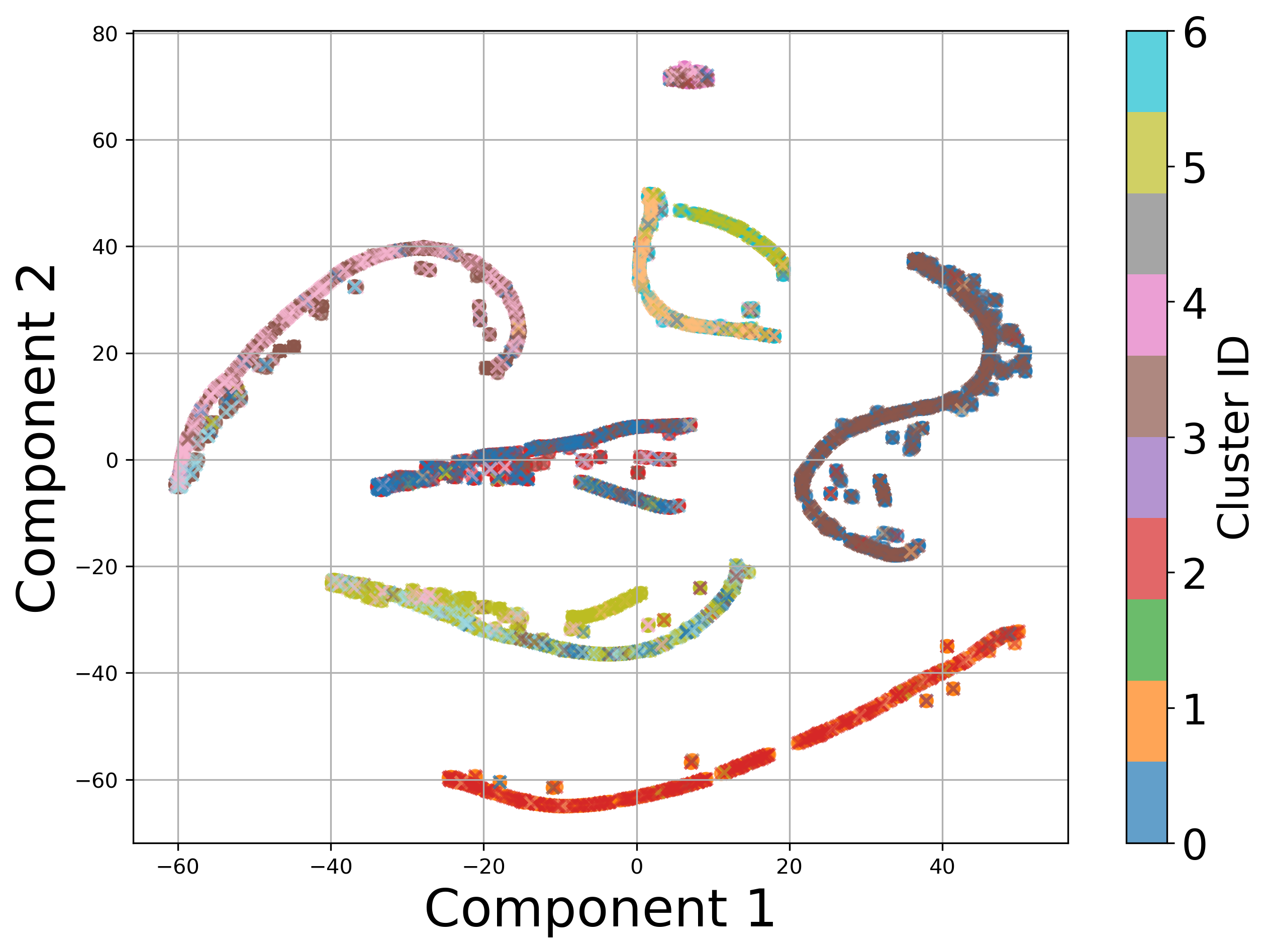}
  \captionsetup{justification=centering}
  \caption{}
  \label{fig:cora_dpr}
\end{subfigure}
\begin{subfigure}[t]{0.32\textwidth} 
  \includegraphics[width=\linewidth]{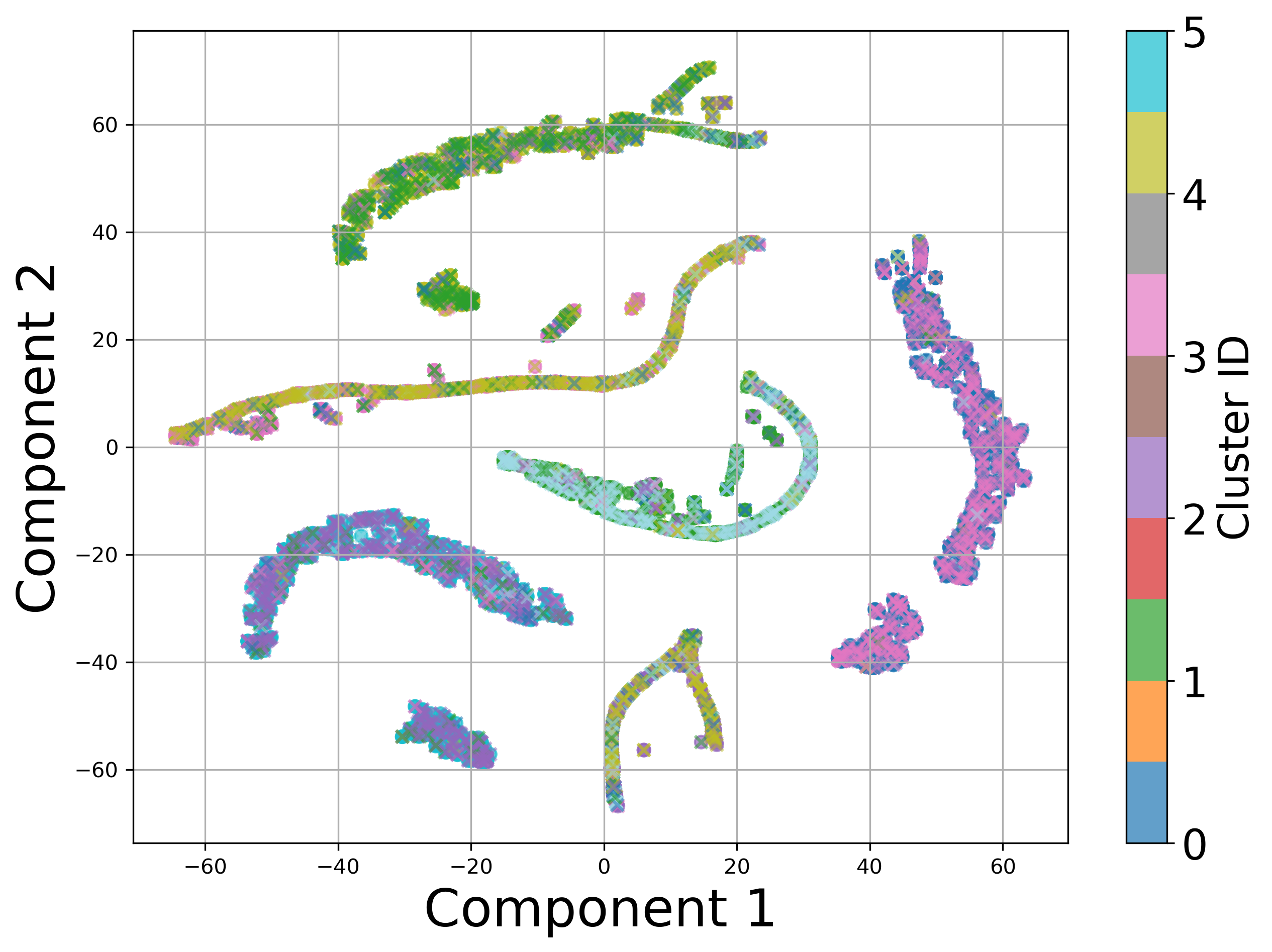}
  \captionsetup{justification=centering}
  \caption{}
  \label{fig:citeseer_dpr}
\end{subfigure}
\begin{subfigure}[t]{0.34\textwidth} 
  \includegraphics[width=\linewidth]{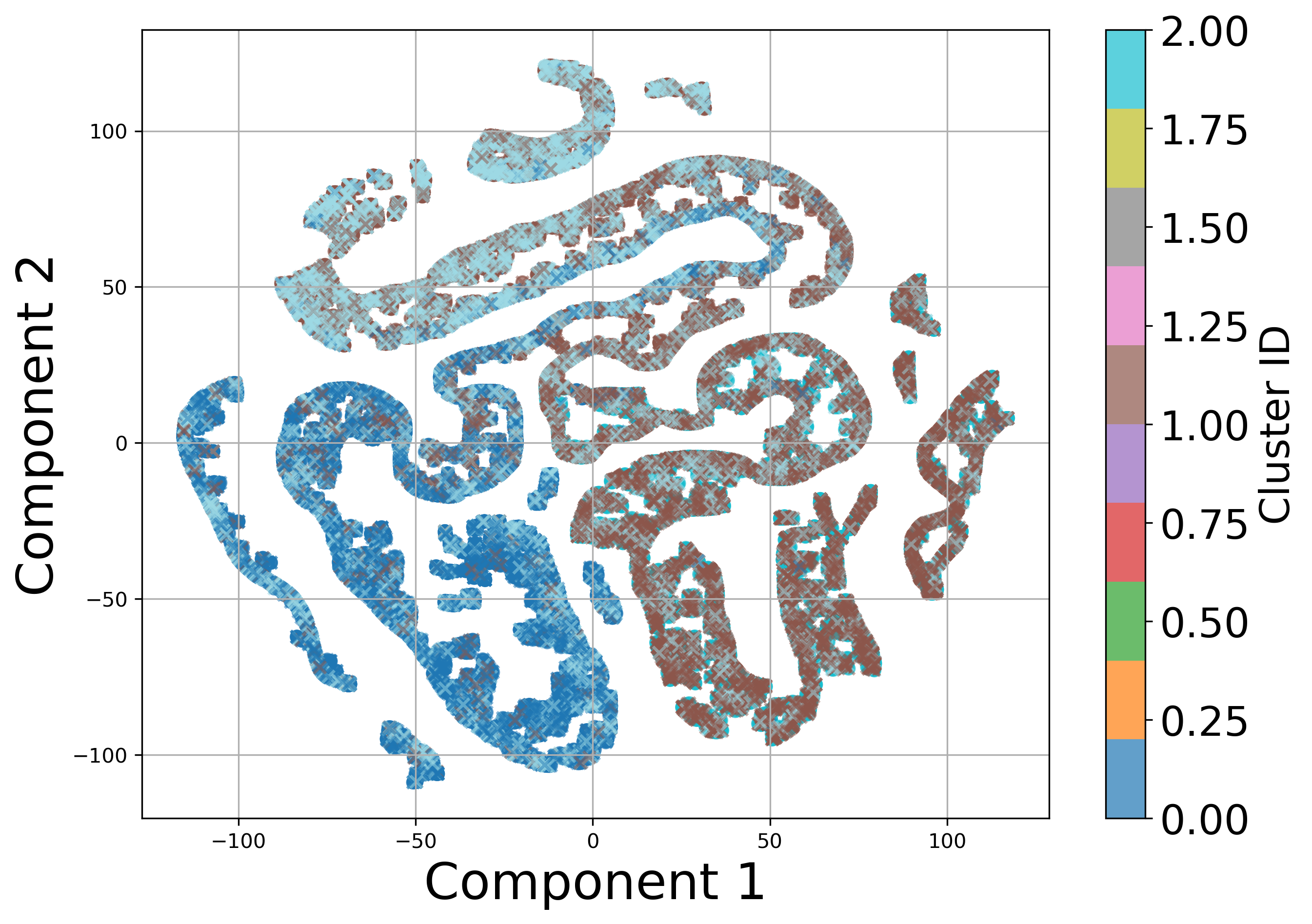}
  \captionsetup{justification=centering}
  \caption{}
  \label{fig:pubmed_dpr}
\end{subfigure}
\begin{subfigure}[t]{0.32\textwidth} 
  \includegraphics[width=\linewidth]{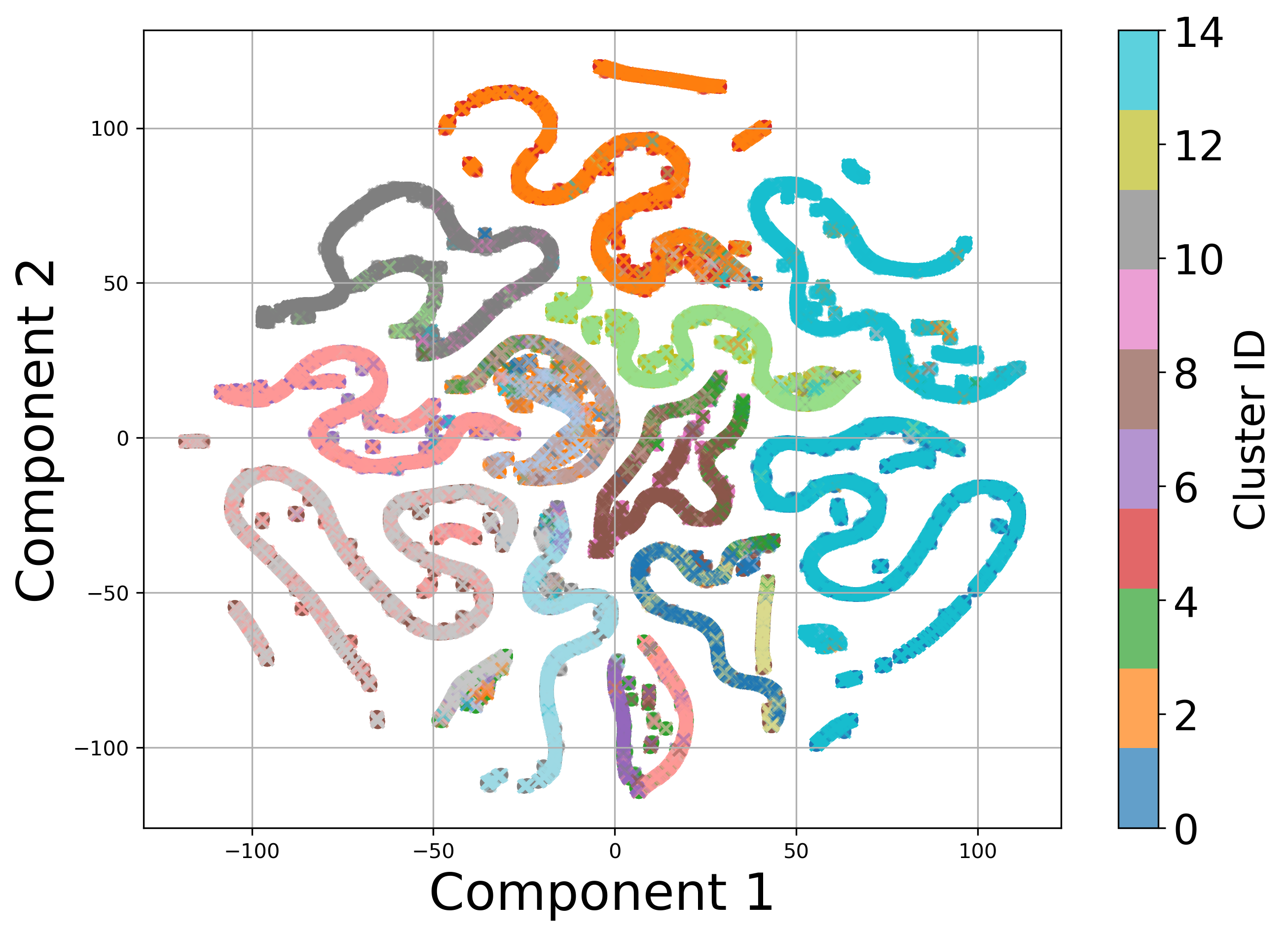}
  \captionsetup{justification=centering}
  \caption{}
  \label{fig:cs_dpr}
\end{subfigure}
\begin{subfigure}[t]{0.32\textwidth} 
  \includegraphics[width=\linewidth]{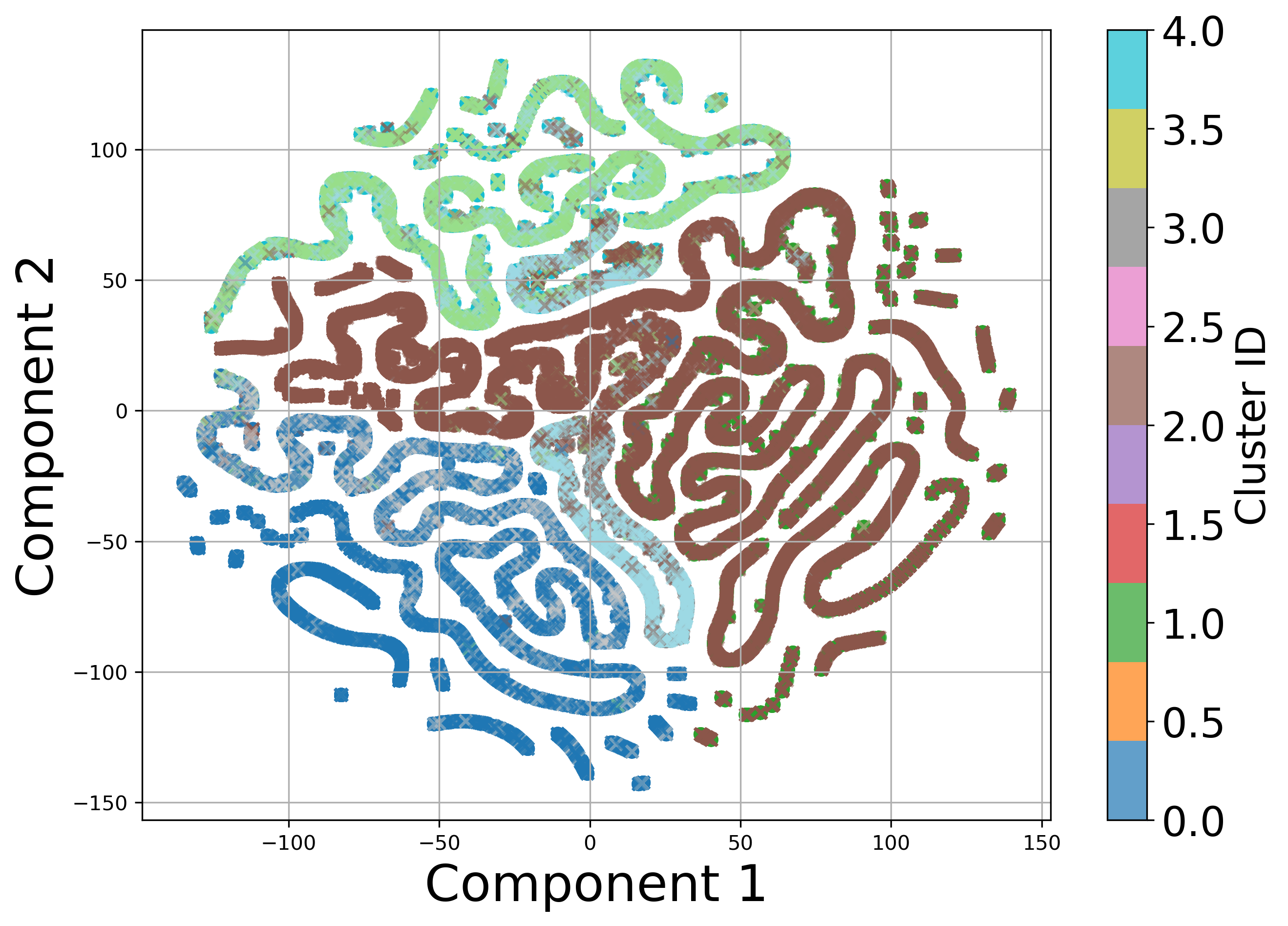}
  \captionsetup{justification=centering}
  \caption{}
  \label{fig:physics_dpr}
\end{subfigure}
\caption{Visualization of clusters (circles) vs. ground truth labels (crosses) formed by DMoN on (a) Cora, (b) CiteSeer, (c) PubMed, (d) Coauthor CS, and (e) Coauthor Physics, and by DMoN--DPR on (f) Cora, (g) CiteSeer, (h) PubMed, (i) Coauthor CS, and (j) Coauthor Physics datasets.}
\label{fig:clusters}
\end{figure*}

%% file: neurips_2025.bbl
\begin{thebibliography}{27}
\expandafter\ifx\csname natexlab\endcsname\relax\def\natexlab#1{#1}\fi

\bibitem[{Aytekin and Karakaya(2014)}]{aytekin2014clustering}
Tevfik Aytekin and Mahmut~{\"O}zge Karakaya. 2014.
\newblock Clustering-based diversity improvement in top-n recommendation.
\newblock \emph{Journal of Intelligent Information Systems}, 42(1):1--18.

\bibitem[{Bianchi et~al.(2019)Bianchi, Grattarola, and Alippi}]{bianchi2019mincut}
Filippo~Maria Bianchi, Daniele Grattarola, and Cesare Alippi. 2019.
\newblock Mincut pooling in graph neural networks.

\bibitem[{Bianchi et~al.(2020)Bianchi, Grattarola, and Alippi}]{bianchi2020spectral}
Filippo~Maria Bianchi, Daniele Grattarola, and Cesare Alippi. 2020.
\newblock Spectral clustering with graph neural networks for graph pooling.
\newblock In \emph{International conference on machine learning}, pages 874--883. PMLR.

\bibitem[{Cangea et~al.(2018)Cangea, Veli{\v{c}}kovi{\'c}, Jovanovi{\'c}, Kipf, and Li{\`o}}]{cangea2018towards}
C{\u{a}}t{\u{a}}lina Cangea, Petar Veli{\v{c}}kovi{\'c}, Nikola Jovanovi{\'c}, Thomas Kipf, and Pietro Li{\`o}. 2018.
\newblock Towards sparse hierarchical graph classifiers.
\newblock \emph{arXiv preprint arXiv:1811.01287}.

\bibitem[{Chen et~al.(2020)Chen, Kornblith, Norouzi, and Hinton}]{chen2020simple}
Ting Chen, Simon Kornblith, Mohammad Norouzi, and Geoffrey Hinton. 2020.
\newblock A simple framework for contrastive learning of visual representations.
\newblock In \emph{International conference on machine learning}, pages 1597--1607. PMLR.

\bibitem[{Defferrard et~al.(2016)Defferrard, Bresson, and Vandergheynst}]{defferrard2016convolutional}
Micha{\"e}l Defferrard, Xavier Bresson, and Pierre Vandergheynst. 2016.
\newblock Convolutional neural networks on graphs with fast localized spectral filtering.
\newblock \emph{Advances in neural information processing systems}, 29.

\bibitem[{Fey and Lenssen(2019)}]{Fey/Lenssen/2019}
Matthias Fey and Jan~Eric Lenssen. 2019.
\newblock Fast graph representation learning with {PyTorch Geometric}.
\newblock In \emph{ICLR Workshop on Representation Learning on Graphs and Manifolds}.

\bibitem[{Jin et~al.(2021)Jin, Zeng, Xia, Huang, and Liu}]{jin2021application}
Shuting Jin, Xiangxiang Zeng, Feng Xia, Wei Huang, and Xiangrong Liu. 2021.
\newblock Application of deep learning methods in biological networks.
\newblock \emph{Briefings in bioinformatics}, 22(2):1902--1917.

\bibitem[{Kim and Kang(2025)}]{kim2025sequentially}
Jongjin Kim and U~Kang. 2025.
\newblock Sequentially diversified and accurate recommendations in chronological order for a series of users.
\newblock In \emph{Proceedings of the Eighteenth ACM International Conference on Web Search and Data Mining}, pages 811--819.

\bibitem[{Kipf and Welling(2016)}]{kipf2016semi}
Thomas~N Kipf and Max Welling. 2016.
\newblock Semi-supervised classification with graph convolutional networks.
\newblock \emph{arXiv preprint arXiv:1609.02907}.

\bibitem[{Lee et~al.(2019)Lee, Lee, and Kang}]{lee2019self}
Junhyun Lee, Inyeop Lee, and Jaewoo Kang. 2019.
\newblock Self-attention graph pooling.
\newblock In \emph{International conference on machine learning}, pages 3734--3743. pmlr.

\bibitem[{Liu et~al.(2023)Liu, Zhan, Yu, Liu, Du, Hu, and Liu}]{liu2023exploring}
Chuang Liu, Yibing Zhan, Baosheng Yu, Liu Liu, Bo~Du, Wenbin Hu, and Tongliang Liu. 2023.
\newblock On exploring node-feature and graph-structure diversities for node drop graph pooling.
\newblock \emph{Neural Networks}, 167:559--571.

\bibitem[{Lloyd(1982)}]{lloyd1982least}
Stuart Lloyd. 1982.
\newblock Least squares quantization in pcm.
\newblock \emph{IEEE transactions on information theory}, 28(2):129--137.

\bibitem[{Newman(2006)}]{newman2006modularity}
Mark~EJ Newman. 2006.
\newblock Modularity and community structure in networks.
\newblock \emph{Proceedings of the national academy of sciences}, 103(23):8577--8582.

\bibitem[{Perozzi and Akoglu(2018)}]{perozzi2018discovering}
Bryan Perozzi and Leman Akoglu. 2018.
\newblock Discovering communities and anomalies in attributed graphs: Interactive visual exploration and summarization.
\newblock \emph{ACM Transactions on Knowledge Discovery from Data (TKDD)}, 12(2):1--40.

\bibitem[{Perozzi et~al.(2014)Perozzi, Al-Rfou, and Skiena}]{perozzi2014deepwalk}
Bryan Perozzi, Rami Al-Rfou, and Steven Skiena. 2014.
\newblock Deepwalk: Online learning of social representations.
\newblock In \emph{Proceedings of the 20th ACM SIGKDD international conference on Knowledge discovery and data mining}, pages 701--710.

\bibitem[{Scarselli et~al.(2008)Scarselli, Gori, Tsoi, Hagenbuchner, and Monfardini}]{scarselli2008graph}
Franco Scarselli, Marco Gori, Ah~Chung Tsoi, Markus Hagenbuchner, and Gabriele Monfardini. 2008.
\newblock The graph neural network model.
\newblock \emph{IEEE transactions on neural networks}, 20(1):61--80.

\bibitem[{Shchur and G{\"u}nnemann(2019)}]{shchur2019overlapping}
Oleksandr Shchur and Stephan G{\"u}nnemann. 2019.
\newblock Overlapping community detection with graph neural networks.
\newblock \emph{arXiv preprint arXiv:1909.12201}.

\bibitem[{Shchur et~al.(2018)Shchur, Mumme, Bojchevski, and G{\"u}nnemann}]{shchur2018pitfalls}
Oleksandr Shchur, Maximilian Mumme, Aleksandar Bojchevski, and Stephan G{\"u}nnemann. 2018.
\newblock Pitfalls of graph neural network evaluation.
\newblock \emph{arXiv preprint arXiv:1811.05868}.

\bibitem[{Spirin and Mirny(2003)}]{spirin2003protein}
Victor Spirin and Leonid~A Mirny. 2003.
\newblock Protein complexes and functional modules in molecular networks.
\newblock \emph{Proceedings of the national Academy of sciences}, 100(21):12123--12128.

\bibitem[{Sun et~al.(2019)Sun, Hoffmann, Verma, and Tang}]{sun2019infograph}
Fan-Yun Sun, Jordan Hoffmann, Vikas Verma, and Jian Tang. 2019.
\newblock Infograph: Unsupervised and semi-supervised graph-level representation learning via mutual information maximization.
\newblock \emph{arXiv preprint arXiv:1908.01000}.

\bibitem[{Tsitsulin et~al.(2023)Tsitsulin, Palowitch, Perozzi, and M{\"u}ller}]{tsitsulin2023graph}
Anton Tsitsulin, John Palowitch, Bryan Perozzi, and Emmanuel M{\"u}ller. 2023.
\newblock Graph clustering with graph neural networks.
\newblock \emph{Journal of Machine Learning Research}, 24(127):1--21.

\bibitem[{Veli{\v{c}}kovi{\'c} et~al.(2018)Veli{\v{c}}kovi{\'c}, Fedus, Hamilton, Li{\`o}, Bengio, and Hjelm}]{velivckovic2018deep}
Petar Veli{\v{c}}kovi{\'c}, William Fedus, William~L Hamilton, Pietro Li{\`o}, Yoshua Bengio, and R~Devon Hjelm. 2018.
\newblock Deep graph infomax.
\newblock \emph{arXiv preprint arXiv:1809.10341}.

\bibitem[{Velickovic et~al.(2019)Velickovic, Fedus, Hamilton, Li{\`o}, Bengio, and Hjelm}]{velickovic2019deep}
Petar Velickovic, William Fedus, William~L Hamilton, Pietro Li{\`o}, Yoshua Bengio, and R~Devon Hjelm. 2019.
\newblock Deep graph infomax.
\newblock \emph{ICLR (Poster)}, 2(3):4.

\bibitem[{Xiao et~al.(2015)Xiao, Freeman, and Hwa}]{xiao2015detecting}
Cao Xiao, David~Mandell Freeman, and Theodore Hwa. 2015.
\newblock Detecting clusters of fake accounts in online social networks.
\newblock In \emph{Proceedings of the 8th ACM Workshop on Artificial Intelligence and Security}, pages 91--101.

\bibitem[{Yang et~al.(2016)Yang, Cohen, and Salakhudinov}]{yang2016revisiting}
Zhilin Yang, William Cohen, and Ruslan Salakhudinov. 2016.
\newblock Revisiting semi-supervised learning with graph embeddings.
\newblock In \emph{International conference on machine learning}, pages 40--48. PMLR.

\bibitem[{Ying et~al.(2018)Ying, You, Morris, Ren, Hamilton, and Leskovec}]{ying2018hierarchical}
Zhitao Ying, Jiaxuan You, Christopher Morris, Xiang Ren, Will Hamilton, and Jure Leskovec. 2018.
\newblock Hierarchical graph representation learning with differentiable pooling.
\newblock \emph{Advances in neural information processing systems}, 31.

\end{thebibliography}
